%% file: 0_main_arxiv.tex
\documentclass[lettersize,journal,compsoc]{IEEEtran}

\input{0_config}

\def\sepappendix{0}

\begin{document}
	\title{CoVR-2: Automatic Data Construction for \\
    Composed Video Retrieval}
	
	\author{Lucas Ventura, \quad Antoine Yang, \quad Cordelia Schmid,~\IEEEmembership{Fellow,~IEEE,} \quad G\"ul Varol
		\IEEEcompsocitemizethanks{
			\IEEEcompsocthanksitem Lucas Ventura and G\"ul Varol are with LIGM, \'{E}cole des Ponts, Univ Gustave Eiffel, CNRS, France.\protect\\
			Email: lucas.ventura@enpc.fr
            \IEEEcompsocthanksitem Antoine Yang is with Google DeepMind in London.
			\IEEEcompsocthanksitem Cordelia Schmid is with WILLOW project-team, ENS/Inria/CNRS, France.
		}
	}

	\IEEEtitleabstractindextext{
		\input{0b_abstract}
	}
	
	\maketitle
	
	\input{1_intro}

	\input{2_relatedwork}

	\input{3_method}

	\input{4_experiments}
	\input{5_conclusions}

	\input{5b_acknowledgements}

	\bibliographystyle{IEEEtran}   
	\bibliography{6_references}

    \vspace{0.5cm}
	{\noindent \large \bf {APPENDIX}}\\
	\input{7_appendix}

\end{document}

%% file: 0_config.tex
\usepackage{amsmath,amsfonts}
\usepackage{algorithmic}
\usepackage{algorithm}
\usepackage{array}
\usepackage[caption=false,font=normalsize,labelfont=sf,textfont=sf]{subfig}
\usepackage{textcomp}
\usepackage{stfloats}
\usepackage{url}
\usepackage{verbatim}
\usepackage{graphicx}
\usepackage{cite}

\usepackage{titletoc} %
\usepackage[pagebackref,breaklinks,colorlinks,citecolor=cvprblue]{hyperref}

\usepackage[utf8]{inputenc} %
\usepackage[T1]{fontenc}    %
\usepackage{booktabs}       %
\usepackage{nicefrac}       %
\usepackage{microtype}      %
\usepackage{fontawesome5} %
\usepackage{multirow} 
\usepackage{xspace} %
\usepackage[table]{xcolor}
\usepackage{pifont} %

\newcommand{\cmark}{\ding{51}}
\newcommand{\xmark}{\ding{55}}

\newcommand{\new}[1]{\textcolor{black}{#1}} %
\newcommand{\newt}[1]{\textcolor{black}{#1}} %

\newcommand{\ourMone}{CoVR-BLIP\xspace}    %
\newcommand{\ourMtwo}{CoVR-BLIP-2\xspace}    %
\newcommand{\ourWV}{WebVid-CoVR\xspace}  %
\newcommand{\ourWVt}{WebVid-CoVR-Test\xspace} %
\newcommand{\ourCC}{CC-CoIR\xspace}

\newcommand{\ourWVandCC}{WV-CC-CoVIR\xspace}

\definecolor{ourcolor}{rgb}{0.807, 0.725, 0.847}
\definecolor{lavanda}{RGB}{144,101,202}
\definecolor{bluecolor}{RGB}{192, 237, 254}
\definecolor{yellowcolor}{RGB}{255, 239, 203}
\definecolor{backcolour}{rgb}{0.95,0.95,0.92}
\definecolor{cvprblue}{rgb}{0.21,0.49,0.74}

\usepackage{ifthen}

\newboolean{showconfidential}
\setboolean{showconfidential}{false}  %

\newcommand{\confidential}[1]{%
  \ifthenelse{\boolean{showconfidential}}{%
    #1%
  }{%
    Due to anonymity reasons, response is hidden.%
  }%
}

\makeatletter
\@ifpackageloaded{hyperref}{%
    \newcommand{\ourURL}{\href{https://imagine.enpc.fr/~ventural/covr}{\texttt{https://imagine.enpc.fr/{\textnormal \textasciitilde}ventural/covr}}\xspace}
}{%
    \newcommand{\ourURL}{{https://imagine.enpc.fr/{\textnormal \textasciitilde}ventural/covr}\xspace}
}
\makeatother

\newboolean{showcr}
\setboolean{showcr}{true}  %

\newcommand{\cameraready}[1]{%
  \ifthenelse{\boolean{showcr}}{%
    #1%
  }{
  }
}

\usepackage{listings}

\renewcommand{\footnoterule}{%
  \kern -3pt
  \hrule width \linewidth height 1pt
  \kern 2pt
}

%% file: 0b_abstract.tex
\begin{abstract}
Composed Image Retrieval (CoIR) has recently gained popularity
as a task that considers \emph{both} text and image queries together, to search for relevant images in a database.
Most CoIR approaches require manually annotated datasets, comprising image-text-image triplets,
where the text describes a modification from the query image to the target image. 
However, manual curation of CoIR \emph{triplets} is expensive and prevents scalability.
In this work, we instead propose
a scalable automatic dataset creation methodology
that generates triplets given video-caption \emph{pairs},
while also expanding the scope of the task to include 
Composed \emph{Video} Retrieval (CoVR).
To this end, we mine paired videos with a similar caption from a large database,
and leverage a large language model to generate the corresponding modification text.
Applying this methodology
to the extensive WebVid2M collection, we automatically construct our \ourWV dataset,
resulting in 1.6 million triplets. 
Moreover, we introduce a new benchmark for CoVR
with a manually annotated evaluation set, along with baseline results.
\new{We further validate that our methodology is equally applicable to image-caption pairs,
    by generating 3.3 million CoIR training triplets using the Conceptual Captions dataset.}
\new{Our model builds on BLIP-2 pretraining, adapting it to composed video (or image) retrieval, and incorporates an additional caption retrieval loss to exploit extra supervision beyond the triplet, which is possible since captions are readily available for our training data by design.}
\new{We provide extensive ablations to analyze the design choices on our new CoVR benchmark.}
Our experiments also demonstrate that training a CoVR model on our datasets effectively transfers to CoIR,
leading to improved state-of-the-art performance in the zero-shot setup on the CIRR, FashionIQ, \new{and CIRCO} benchmarks.
Our code, datasets, and models are publicly available at \ourURL.

\end{abstract}


%% file: 1_intro.tex
\section{Introduction}
\label{sec:intro}

\begin{figure} %
  \centering
  \includegraphics[width=.99\linewidth]{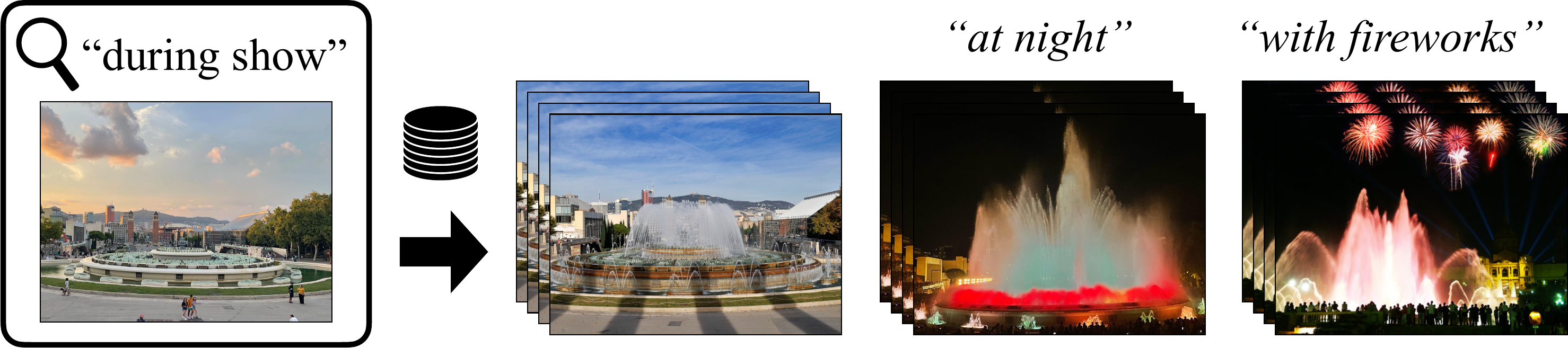}
  \caption{\textbf{Task:} Composed Video Retrieval (CoVR) seeks to retrieve \emph{videos} from a database by searching with both a query image and a query text. The text typically specifies the desired modification to the query image. In this example, a traveller might wonder how the photographed place looks like during a fountain show, by describing several modifications, such as ``during show at night, with fireworks''.
  }
  \vspace{-0.3cm}
  \label{fig:teaser}
\end{figure}

\IEEEPARstart{C}{onsider} the scenario where a traveller takes a picture of a landmark or scenic spot
and wants to discover videos that capture the essence of that location, by specifying
certain conditions via text. 
For example, the query image in Figure~\ref{fig:teaser} (of a fountain in Barcelona),
along with the text ``during show'' should bring the video showcasing the fountain show.
Further refining the text query such as ``during show at night'', would allow the traveller
to decide whether to wait for the show until the night time.
In this work, our goal is composed video retrieval (CoVR), where the user
performs such multi-modal search, by querying an image of a particular visual concept
and a modification text, to find videos that exhibit the similar visual characteristics
with the desired modification, in a dynamic context. 

CoVR has many use cases,
including but not limited to searching online videos for finding reviews of a specific
product, how-to videos of a tool for specific usages, live events in specific locations,
sports matches of specific players. 
Similar to composed image retrieval (CoIR),
CoVR is also particularly useful when conveying a concept with a visual is easier and/or more
accurate than only using words
(e.g., unknown location/object, a specific camera view, a specific color).

\begin{figure*}
  \centering
  \includegraphics[width=.75\linewidth]{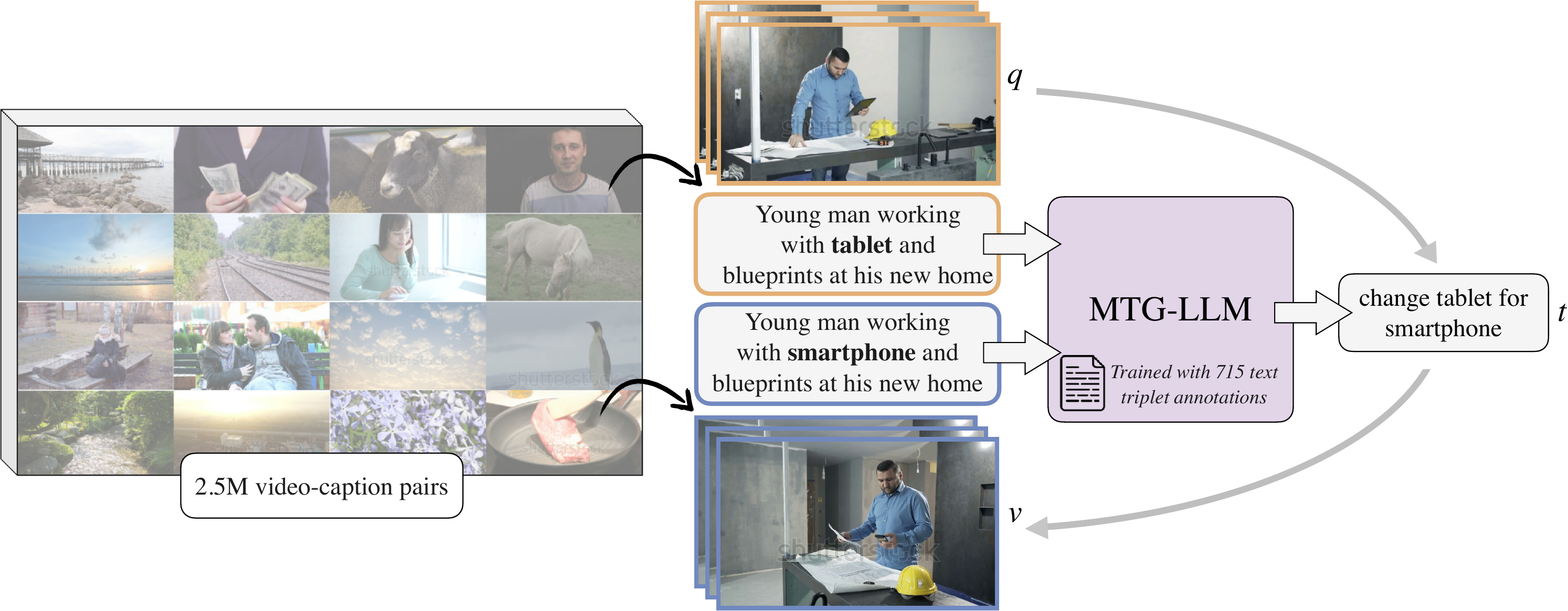}
  \caption{\textbf{Method overview:} We automatically mine similar caption pairs from a large video-caption database from the Web,
  and use our modification text generation language model (MTG-LLM) to describe
  the difference between the two captions.
  MTG-LLM is trained on a dataset of 715 triplet text annotations \cite{brooks2022instructpix2pix}.
  The resulting triplet with the two corresponding videos (query $q$ and target video $v$) and the modification text ($t$) is therefore obtained fully automatically, allowing a scalable CoVR training data generation. 
  }
  \label{fig:method}
\end{figure*}

Given the increased momentum in vision and language research in the recent years \cite{BLIP, clip2021},
CoIR has emerged as a new task \cite{tirg}, and since then witnessed improvements of both models and benchmarks
\cite{baldrati2023zero, cclip, gu2023compodiff, levy2023case, cirr, fashioniq}.
However, to the best of our knowledge, CoVR was not studied before.
A key challenge in building CoVR models is the difficulty of gathering suitable training data
of video-text-video triplets. 
We overcome this limitation by developing
an automatic approach to generate triplets from existing video-caption collections.
Specifically, we mine video pairs whose corresponding captions slightly differ
in text space. 
We automatically describe this difference with a language model,
which we train for a \emph{modification-text generation} task. 
In particular, we use manually annotated triplets, each containing:
(a) source caption, (b) target caption, (c) the modification text. 
We then finetune a large language model (LLM) \cite{touvron2023llama} 
by inputting (a-b), and outputting (c).
We assume the resulting modification to describe
the difference between the corresponding videos, thus obtaining video-text-video triplets
(see Figure~\ref{fig:method} for an overview).
When training our CoVR/CoIR models, we can flexibly select one or more frames from the videos,
enabling multiple settings (i.e., retrieving images or videos).

We apply our triplet generation approach to \new{two seed datasets:
(1) WebVid2M~\cite{bain21_frozen} with 2.5M video-caption pairs,
and (2) Conceptual Captions~\cite{sharma2018CC3M} with 3.3M image-caption pairs.
We call the resulting training sets as  \ourWV and \ourCC, which contain 1.6M CoVR and 3.3M CoIR triplets, respectively.}
By virtue of its automatic generation procedure, %
\new{our training data} is inherently noisy.
To efficiently train on such large-scale and noisy data, we use a contrastive loss \cite{InfoNCE}, 
adopting the HN-NCE variant from \cite{hnnce2023} to upsample the significance of hard negatives. 
\new{Moreover, we integrate an additional contrastive loss,
	by also retrieving the textual embeddings of the target image, aiming to enhance the attention on the modification text detail.} 
We design a CoVR model based on the cross-modal 
\new{BLIP-2~\cite{li2023blip2}}
and use query scoring \cite{max2022Hitchhiker} to exploit information from multiple video frames.
Training this model on \ourWV shows strong transferability to the CoIR task, in both zero-shot and finetuning settings, achieving state-of-the-art results
on the standard CIRR~\cite{cirr}, FashionIQ~\cite{fashioniq}, \new{and CIRCO~\cite{baldrati2023zero}} benchmarks in the zero-shot setup.
\new{We further improve the performance by jointly training on our \ourWV and \ourCC data together.}
Finally, to foster research in CoVR, we repeat our generation procedure on a distinct subset of the WebVid10M dataset~\cite{bain21_frozen} and manually select correctly generated samples to constitute \ourWVt, a test set of 2,435 CoVR triplets.
We find that our model achieves promising results on \ourWVt compared to standard baselines.

To summarize, our contributions are:
(i) We propose a scalable approach to automatically generate composed visual retrieval training data.
\new{With this methodology, we generate 1.6M \ourWV and 3.3M \ourCC triplets.}
(ii) We show that training \new{composed retrieval models} %
on \new{our generated datasets}
transfers well to the CoIR %
\new{benchmarks}, and achieves state-of-the-art results on CIRR, FashionIQ, \new{and CIRCO} \new{datasets} %
in the zero-shot setup.
(iii) We evaluate our model on \ourWVt, a new CoVR benchmark that we manually annotate. 
Our code, datasets, %
and models are publicly available at \ourURL.

%% file: 2_relatedwork.tex
\section{Related Work}
\label{sec:related}

\input{tables/datasets}

\noindent\textbf{Composed image retrieval (CoIR).}
CoIR~\cite{tirg} has been an active area of research in recent 
years~\cite{DialogBasedIIR,tirg,fashioniq,cclip,cirr,ARTEMIS,Kim_Yu_Kim_Kim_2021_dcnet,baldrati2023zero,saito2023pic2word,gu2023compodiff,Zhang2024MagicLens}.
Most methods designed for this problem use manually annotated image-text-image triplets
for training~\cite{fashioniq,cirr,cclip,ARTEMIS}.
Recent works, such as Pic2Word~\cite{saito2023pic2word}, SEARLE~\cite{baldrati2023zero}, \new{TFCIR~\cite{sun2023trainingfree}, and LinCIR~\cite{gu2023languageonly}}, explore zero-shot CoIR setups where no manually annotated CoIR triplet is used.
\new{More recently, Karthik et al.~\cite{karthik2023visionbylanguage}
propose
a training-free
method
for CoIR.}
The approaches~\cite{saito2023pic2word,baldrati2023zero,sun2023trainingfree,gu2023languageonly} build on CLIP~\cite{clip2021} and train 
a mapping network using
image-only data for text inversion
so that they can be flexibly composed with text descriptions. 
Our approach is similar in that it avoids collecting manual triplets; however, we instead 
perform supervised training on automatically
generated image-text-video triplets given only video-text pairs.
We also differ from above works by focusing on the composed video retrieval (CoVR) task, as opposed to \new{only} CoIR.

\noindent\textbf{Datasets for composed image retrieval.}
CIRR~\cite{cirr} and Fashion-IQ~\cite{fashioniq} are the two most widely used CoIR benchmarks.
\new{Very recently, Baldrati et al. proposed a new test CoIR dataset CIRCO~\cite{baldrati2023zero}, which has gained popularity for having multiple ground truths and many distractors.}
All three datasets are manually annotated, hence small scale (about 30k triplets, see Table~\ref{tab:datasets}) due to the high cost implied in collecting CoIR triplets.
To scale up, two recent works proposed larger, automatically generated CoIR datasets: LaSCo~\cite{levy2023case} and SynthTriplets18M~\cite{gu2023compodiff}.
The LaSCo dataset~\cite{levy2023case} is generated using the visual question answering annotations and the pairing between images and counterfactual images in the VQAv2 dataset~\cite{VQA}.
In detail, this dataset provides for each (image, question, answer) triplet a counterfactual triplet with the same question and different image and answer.
In contrast, we do not rely on such expensive annotation schemes.
SynthTriplets18M~\cite{gu2023compodiff} uses the text-conditioned image editing framework InstructPix2Pix~\cite{brooks2022instructpix2pix} to automatically generate CoIR data.
Their edit text generation process is similar to ours, but our generation process differs in that we automatically mine similar videos from a dataset of video-text pairs to construct CoVR triplets instead of generating visual data.
In experiments, we show the superiority of our triplet construction procedure as we achieve much higher CoIR results (e.g., 43.7\% vs 26.7\% zero-shot R@1 on CIRR while generating fewer data). 
\new{Similar conclusions hold when pretraining on our automatic CoIR triplets (\ourCC).}
\newt{We additionally provide a controlled experiment by training our model on their synthetic images~\cite{brooks2022instructpix2pix,gu2023compodiff} and demonstrate the advantages of using real data.}
Lastly, our \ourWV dataset is not limited to still images and considers videos,
while standing out \new{as larger than all previous composed retrieval datasets}
in the natural domain, as depicted in Table~\ref{tab:datasets}.

\noindent\textbf{Vision-language pretraining.}
Many strong multi-modal models have been pretrained on large datasets of image-caption pairs~\cite{alayrac2022flamingo, chen2019uniter, li2021align, BLIP, li2020oscar, clip2021, jia2021scaling, su2019vl, schuhmannlaion,li2023blip2}
or video-caption pairs~\cite{akbari2021vatt, li2021align2, li2020hero, miech19howto100m, miech20endtoend, xu2021videoclip, sun2022long, xue2022advancing, zhao2022learning}.
In contrast, we generate CoVR training data from video-caption pairs instead of directly training on them.
Our data generation approach is also related to other generation approaches used for other tasks, e.g., action recognition~\cite{nagrani2020speech2action}, visual question answering~\cite{yang2021just} and visual dialog~\cite{liu2023visual}.
However, unlike all these tasks, the CoVR task requires retrieving visual data.

\noindent\textbf{Video retrieval.}
Text-to-video retrieval has received great attention over the last few years~\cite{CLIP2Video, CLIP2TV, BridgeFormer, liu2022ts2net, clip4clip2021, Ma2022X-CLIP, rasheed2023ViFiCLIP, xue2022clipvip, Jianwei2021TACo, FILIP, xu2021videoclip}. %
We also make use of multiple video frames with query scoring similar to \cite{max2022Hitchhiker}.
However, different from these methods, we focus on \emph{composed} video retrieval, where the query consists of both text and visual data.

%% file: tables/datasets.tex
\begin{table} %
 \caption{\textbf{Existing datasets:} We compare our proposed \new{\ourCC and} \ourWV training dataset\new{s}, \new{along with} %
 	its manually annotated test set \ourWVt with existing composed visual retrieval datasets. \faIcon{camera}~denotes image, \faIcon{video}~denotes video datasets. 
	We contribute the largest training dataset\new{s} for the natural domain. 
	Note that, while SynthTriplets18M is larger, the transfer performance to real images is ineffective potentially due to a domain gap (see Table~\ref{tab:sota_cirr+fiq}).
}
    \centering
    \resizebox{0.99\linewidth}{!}{
    \begin{tabular}{lcrrrrl}
        \toprule
        Dataset     &  Type & \#Triplets  & \vtop{\hbox{\strut \#Unique}\hbox{\strut visuals}} & \vtop{\hbox{\strut \#Unique}\hbox{\strut words}}  & \vtop{\hbox{\strut Avg. text}\hbox{\strut length}} & Domain \\
        \midrule
        CIRR~\cite{cirr}        & \faIcon{camera} & 36,554  & 21,185 & 7,129 & 59.51 & Natural \\
        FashionIQ~\cite{fashioniq}   & \faIcon{camera} & 30,132 & 7,988 & 4,425 & 27.13 & Fashion    \\
        CIRCO-Test~\cite{baldrati2023zero} & \faIcon{camera} & 800 & - & 870 & 50.94 & Natural \\
        LaSCo~\cite{levy2023case}       & \faIcon{camera} & 389,305 & 121,479 & 13,488 & 30.70 & Natural\\
        SynthTriplets18M~\cite{gu2023compodiff} & \faIcon{camera}  & 18,000,000 & - & - & - & Synthetic \\
        \rowcolor{ourcolor!64}\ourCC  & \faIcon{camera} & 3,315,773 & 356,582 	& 28,183 & 24.65 & Natural \\
        \rowcolor{ourcolor!64}\ourWV  & \faIcon{video}  & 1,644,276 & 130,559 	& 25,654 & 23.36 & Natural\\
        \rowcolor{ourcolor!64}\ourWVt & \faIcon{video}  & 2,556 	& 4,886 	& 1,910 & 21.97 & Natural \\
        \bottomrule
    \end{tabular}
    }
    \label{tab:datasets}
\end{table}

%% file: 3_method.tex
\section{Automatic Triplet Generation and Training}
\label{sec:method}

The goal of our composed video retrieval (CoVR) task is, given an input image $q$ and a modification text $t$, to retrieve a modified video $v$ in a large database of videos\footnote{Note that $q$ could
also be a video query, but in our main experiments we focus on an image query, and provide
more results in \if\sepappendix1{(Section~D.1} \else{(Section~\ref{app:subsec:video-query} }\fi
of the Appendix) with video queries.}.
Our goal is to avoid the manual annotation of $(q, t, v$) triplets for training.
Hence we automatically generate such triplets from Web-scraped video-caption pairs, as explained in Section~\ref{subsec:gen} and illustrated in Figure~\ref{fig:method}. %
The resulting \ourWV dataset, together with its manually curated evaluation set, is presented in Section~\ref{subsec:data}.
Finally, we present how we train a CoVR model using \ourWV in Section~\ref{subsec:training}.

\subsection{Generating composed video retrieval triplets}
\label{subsec:gen}
Given a large (Web-scraped) dataset of video-caption pairs, %
we wish to automatically generate video-text-video CoVR triplets $(q, t, v$) where the text $t$ describes a modification to the visual query $q$.
However, the dataset of video-caption pairs neither contains annotations of paired videos, nor modification text that describes their difference.
Hence we propose a methodology to automatically mine paired videos and describe their difference, as described below.
Note that for illustration, we take as an example the WebVid2M dataset~\cite{bain21_frozen} with 2.5M video-caption pairs, but this methodology could be applied to other large datasets of video-text (or image-text) pairs.
\new{
To strengthen our conclusions, we employ the same methodology with the Conceptual Captions image-text dataset~\cite{sharma2018CC3M},
which is briefly described along with experiments in Section~\ref{subsec:cc-coir}. While the rest of this section
focuses on CoVR, the data generation pipeline is similar for CoIR.}

\noindent\textbf{Mining paired videos by pairing captions.}
In order to obtain video pairs that exhibit visual similarity while differing in certain aspects, we leverage their associated captions.
The core idea is that videos with similar captions are likely to have similar visual content.
Specifically, we consider captions that differ by a single word, excluding punctuation marks.
For instance, the caption \textit{``Young woman smiling''} is paired with \textit{``Old woman smiling''} and \textit{``Young couple smiling''}.
In the 2M distinct captions from WebVid2M, this process allows us to identify a vast pool of 1.2M distinct caption pairs with 177k distinct captions, resulting in 3.1M paired videos. 
In the following, we describe further steps to filter the data into a smaller set.

\noindent\textbf{Filtering caption pairs.}
We wish to automatically generate the modification text between paired videos using their (paired) captions.
However, caption pairs with the same meaning are likely to result in meaningless differences.
On the contrary, caption pairs that differ too much are likely to result in large visual differences that cannot be easily described.
To address these issues, we filter out caption pairs that are too similar and too dissimilar.
Specifically, we exclude caption pairs with CLIP text embedding similarity $\ge$ 0.96 (e.g., \textit{``Fit and happy young couple \underline{playing} in the park''} and \textit{``Fit and happy young couple \underline{play} in the park''}) and caption pairs with CLIP text embedding similarity $\le$ 0.6 (e.g., \textit{``\underline{Zebra} on a white background''} and \textit{``\underline{Coins} on a white background''}).
We also exclude pairs where the captions differ by a digit (which mostly consist of a date in practice), a word not part of the English dictionary, or by a rare word.
Rare words are detected based on the \texttt{zipf} frequency~\cite{robyn2018word2freq}. 
Finally, we remove templated captions such as \textit{``abstract of''}, \textit{``concept of''}, and \textit{``flag of''} which are over-represented in WebVid2M.
At the end of this filtering stage, we have 370k distinct caption pairs with 92k distinct captions, resulting in 1.2M paired videos that we will use to generate the modification text.
Note that we can use these paired videos in both directions to generate triplets, 
as the source and target videos can be swapped.

\noindent\textbf{Generating a modification text from paired captions.}
In order to generate a modification text between paired videos, we develop and apply a ``modification text generation large language model'' (MTG-LLM) to their corresponding paired captions.
We describe the MTG-LLM inference process below and then explain its training details.
The MTG-LLM takes as input two paired captions and generates a modification text that describes
the difference between the two captions (see Figure~\ref{fig:method}).
In detail, the generation is auto-regressive, i.e., we recursively sample from the token likelihood 
distribution conditioned on the previously generated tokens 
until an end-of-sentence token is reached.
Examples of the input-output, and details about the prompt format, 
which involves concatenating the two captions with a delimiter, 
can be found in \if\sepappendix1{Section~C.4} \else{Section~\ref{app:subsec:mtg-llm} }\fi
of the Appendix.
We use top-k sampling~\cite{fan2018hierarchical} for generating the tokens instead of 
maximum-likelihood methods such as beam search.
Note that we only generate a single modification text per caption pair for computational efficiency,
but the MTG-LLM could be used to generate multiple modification texts per caption pair which could 
serve as a data augmentation in future work.

We now describe the training details of the MTG-LLM.
We start from a LLM pretrained with a next token prediction objective on a Web-scale text dataset, namely LLaMA~\cite{touvron2023llama}.
We then finetune this LLM for the MTG task on a manually annotated text dataset.
In particular, we repurpose the editing dataset 
from InstructPix2Pix~\cite{brooks2022instructpix2pix}, which provides a modification text and a target caption for 700 input captions.
We augment this dataset with 15 annotations that cover additional cases.
More details about the additional examples can be found in \if\sepappendix1{Section~C.4} \else{Section~\ref{app:subsec:mtg-llm} }\fi
.

\noindent\textbf{Filtering video pairs.}
We wish to avoid some modification texts being over-represented in the dataset as it could harm training.
Hence, if there are more than 10 video pairs associated with the same pair of captions (therefore leading to the same modification text), we only select top 10 video pairs.
As the CoVR task typically involves similar query-target video pairs, we choose pairs of videos with the highest visual similarity, as measured by the CLIP visual embedding similarity computed at the middle frame of the videos.

\begin{figure}%
  \centering
    \includegraphics[width=0.48\textwidth]{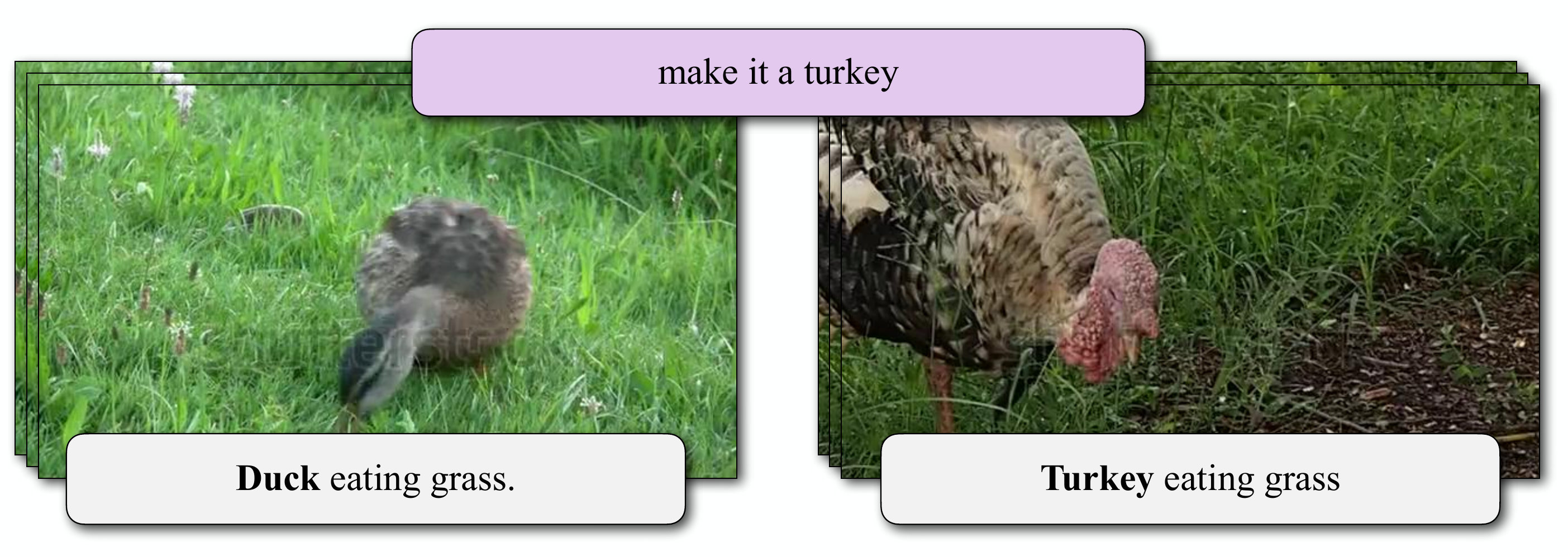}\\
    \vspace{0.25cm}
    \includegraphics[width=0.48\textwidth]{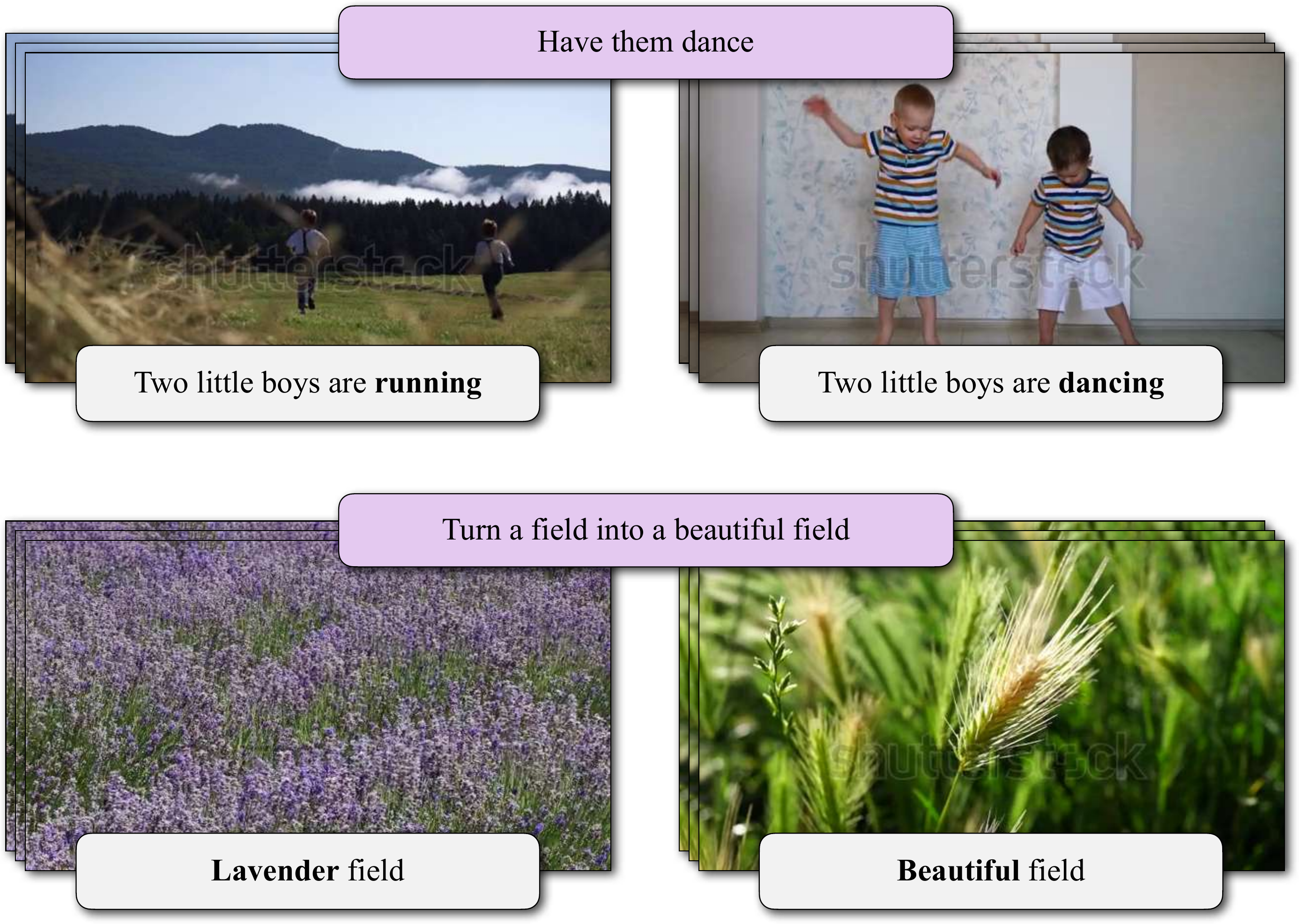}
  \caption{\textbf{Examples of generated CoVR triplets in \ourWV:}
  The middle frame of each video is shown with its corresponding caption,
  with the distinct word highlighted in bold. Additionally,
  the generated modification text is displayed on top of each pair of videos. 
  The bottom example illustrates a noisy generated modification text, as `beautiful' is subjective and
  both target and query videos can be considered as beautiful fields.
  } 
  \label{fig:examples}
\end{figure}

\subsection{Our resulting \ourWV dataset}
\label{subsec:data}
In the following, we describe the training and test partitions of our CoVR data.
While our training set is automatically generated, our test set is manually verified.

\noindent\textbf{\ourWV: a large-scale CoVR training dataset.}
By applying the previously described pipeline to the WebVid2M dataset~\cite{bain21_frozen}, we generate \ourWV, a dataset containing 1.6M CoVR triplets, which is significantly larger than prior datasets (see Table~\ref{tab:datasets}).
On average, a video lasts 16.8 seconds, a modification text contains 4.8 words, and one target video is associated with 12.7 triplets.
We study the effect of the modification text length in \if\sepappendix1{Section~D.5} \else{Section~\ref{app:sec:modification-text-length} }\fi
of the Appendix.
\ourWV is highly diverse with 131k distinct videos and 467k distinct modification texts.
Examples of CoVR triplets from the \ourWV dataset are illustrated in Figure~\ref{fig:examples}.
These examples (along with additional ones included in \if\sepappendix1{Section~E.3} \else{Section~\ref{app:subsec:qualitative_triplets} }\fi
)
demonstrate the diversity present in \ourWV, 
highlighting a wide range of content and variations in the modification texts. 
However, it is important to acknowledge that some noise naturally exists in the dataset, 
as shown in the bottom example of Figure~\ref{fig:examples},
where the text does not describe the difference between the two videos due to both videos describing beautiful fields.
We provide further analysis such as removal of inappropriate content,
and dataset statistics of \ourWV in \if\sepappendix1{Section~A} \else{Section~\ref{app:sec:WV-dataset-statistics} }\fi
of the Appendix.

\noindent\textbf{\ourWVt: a new CoVR evaluation benchmark.}
Due to the noise in \ourWV, we manually annotate a small test set, dubbed \ourWVt, for evaluation.
For this, we first repeat the data generation procedure described in Section~\ref{subsec:gen}, but on a different corpus of video-caption pairs.
Specifically, we consider video-caption pairs from the WebVid10M corpus~\cite{bain21_frozen} that are not included in the WebVid2M dataset, resulting in a pool of 8 million video-caption pairs. 
This ensures that other models using WebVid2M for pretraining have not been exposed to any of the test examples.
In the video pairs filtering stage, for each pair of captions, we here only keep one pair of videos (the one with the highest visual similarity).
This results in 163k candidate triplets that could be used for testing purposes.
We randomly sample 7k triplets that we use for validation and randomly sample 3.2k other triplets that we manually annotate as described below.

We augment the 3.2k triplets by generating two additional modification texts with the MTG-LLM.
The annotator reads the three generated modification texts, looks at three frames from the query and target videos, and either keeps the best modification text if at least one is valid or discards the sample.
Through this meticulous annotation process, we ensure that the test set comprises high-quality and meaningful CoVR triplets.
This results in a test set of 2.5k triplets, i.e., about 22\% of the examples are considered as noisy and are discarded.

\begin{figure*}%
  \centering
    \includegraphics[width=0.95\textwidth]{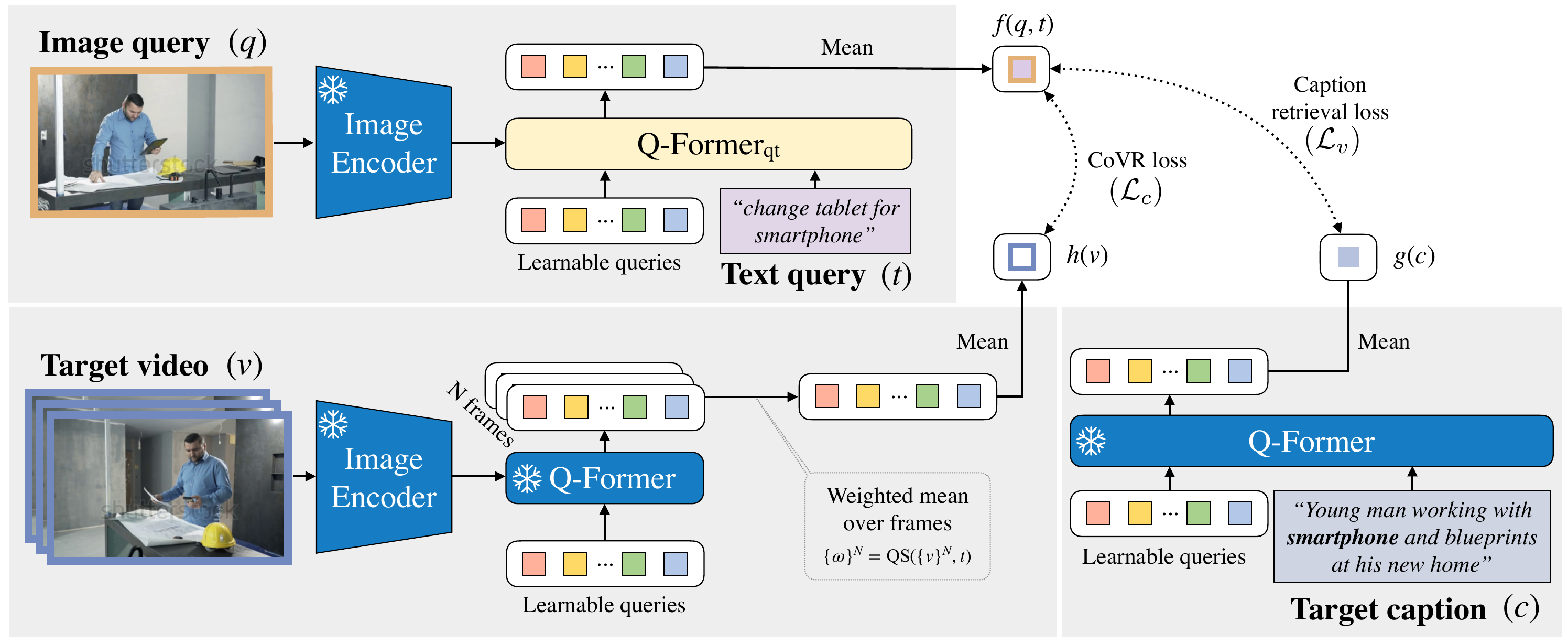}
      \caption{\new{\textbf{Model architecture of \ourMtwo:}
      		The BLIP-2 \cite{li2023blip2} image encoder extracts visual features from the image query $q$.
      These visual features are combined with the text query $t$ (modification text) through the BLIP-2 image-grounded text encoder
      to obtain a multi-modal query embedding $f(q, t)$. 
      To encode videos, $N$ frames are %
      individually encoded with the BLIP-2 image encoder and its Q-Former, 
      and aggregated via a weighted mean into a single video embedding $h(v)$. 
      The goal of CoVR (video retrieval) is to maximize similarity between the multi-modal query $f(q, t)$ and the target video $h(v)$. 
      During training, an additional caption retrieval loss $\mathcal{L_c}$ is defined between $f(q, t)$ and the target caption
      embedding $g(c)$. Note that for simplicity, we visualize one Q-Former block, but in practice there are 12 blocks as in \cite{li2023blip2}.
      To reduce the 32 tokens output by the Q-Former, we simply average them before computing cosine similarities.
      While each Q-Former is initialized from the BLIP-2 pretraining, they are finetuned on our CoVR/CoIR data. When training for CoIR,
      the target becomes a single image, removing the need for weighted mean.
       See Section~\ref{subsec:training} for more details.} 
  } 
  \label{fig:model}
\end{figure*}

\subsection{Training on \ourWV}\label{subsec:training}
Here, we describe our CoVR model architecture and how we train it on our \ourWV dataset.

\noindent\textbf{CoVR-BLIP-2 model architecture.} 
Our model, \new{illustrated in Figure~\ref{fig:model}}, builds on a pretrained image-text model, \mbox{BLIP-2}~\cite{li2023blip2},
\new{an enhancement over the original BLIP~\cite{BLIP}. %
	The main architectural difference with the CoVR-BLIP approach in our preliminary work~\cite{ventura23covr}
	is the use of BLIP-2~\cite{li2023blip2} instead of BLIP~\cite{BLIP}.
BLIP-2 introduces a lightweight Querying Transformer (\mbox{Q-Former}) with learnable queries
for more efficient visual feature extraction.
Like its predecessor,
\mbox{BLIP-2}} is pretrained on a large dataset of image-caption pairs with three vision-language objectives: image-text contrastive learning, image-text matching, and image-conditioned language modeling. 
However, BLIP-2 is not trained for composed visual retrieval with both visual and text inputs.
Therefore we adapt BLIP-2 to the CoIR/CoVR task as follows.

We use the BLIP-2 image encoder to encode the image query $q$
(which corresponds to the middle frame of the video in case of \ourWV).
The resulting visual features and the modification text ($t$) are then forwarded to the \mbox{BLIP\new{-2}} image-grounded text encoder together, which outputs a multi-modal embedding $f(q,t) \in \mathbb{R}^d$ where $d$ is the embedding dimension. 
\new{Note that we compute mean across the 32 output tokens corresponding to the learnable
query inputs of the Q-Former, after passing them through linear projection layer initialized from BLIP-2 text projection layer.
}

To retrieve a target video 
$v_k$ from a database of videos $V$, we compute embedding vectors for all gallery videos as follows.
We uniformly sample $N$ frames from the video and compute a weighted mean of the BLIP\new{-2} image embeddings to obtain the video embedding vector $h(v_k) \in \mathbb{R}^d$.
The weights $\{\omega\}^N$ are obtained by computing the similarity between 
the corresponding frame and the modification text 
using the pretrained BLIP\new{-2} image and text encoders, respectively
(\new{introduced as `query scoring'} in \cite{max2022Hitchhiker} in the context of text-to-video retrieval).
Using pretrained and frozen BLIP\new{-2} embeddings allows us to precompute and store all these weights, which we refer to as $\{\omega\}^N=QS(\{v\}^N, t)$
for query-scoring between each of the $N$ frames of the video $v$ and the text $t$.

\new{At test time,}
given a multi-modal embedding $f(q,t)$, the retrieved video is the one that maximizes the embedding similarity, i.e., arg\,max$_{{v}_k \in V}(h({{v}_k})\cdot f(q,t)^T)$.

\noindent\textbf{Training.} 
In order to train on \ourWV, we use a contrastive learning approach~\cite{InfoNCE,hnnce2023}, as it has been shown to be effective to learn strong multi-modal representations from large-scale noisy data~\cite{clip2021}.
We make the following design choices.
First, we create a training batch by sampling distinct target videos; and for each target video, 
we randomly sample an associated image-text query pair.
Iterating over videos ensures that the same target video appears only once in a batch and maximizes the number of different target videos that can be used as negatives in contrastive learning.
We show the benefit of this approach in Section~\ref{subsec:ablations} (Table~\ref{tab:training-strategies}).
Second, we employ HN-NCE~\cite{hnnce2023} which increases the weight of most similar samples 
and uses as negatives all target videos $v_{j \in \mathcal{B}}$ in the batch $\mathcal{B}$.
Formally, given a training batch $\mathcal{B}$ of triplets $(q_i, t_i, v_i)$, we minimize the following loss:
\begin{align} 
\mathcal{L}_v(\mathcal{B})=
    &-\sum_{i \in \mathcal{B}}\text{log}\left(\frac{e^{S_{i,i}/\tau}}{\alpha \cdot {e^{S_{i,i}/\tau}} + \sum_{j \neq i}{e^{S_{i,j}/\tau}w_{i,j}} }\right) \nonumber \\
    &-\sum_{i \in \mathcal{B}} \text{log}\left(\frac{e^{S_{i,i}/\tau}}{\alpha \cdot {e^{S_{i,i}/\tau}} + \sum_{j \neq i}{e^{S_{j,i}/\tau}w_{j,i}} }\right)
    \label{eq:Lv}
\end{align}
where $\alpha$ is set to 1, temperature $\tau$ is set to $0.07$, $S_{i,j}$ is the cosine similarity between the multi-modal embedding $f(q_i, t_i)$ and the target video embedding $h(v_j)$, and $w_{i,j}$ is set as in \cite{hnnce2023} with $\beta=0.5$.

\noindent \new{\textbf{Composed caption retrieval as an additional loss term.}
    In addition to using the video as a target, 
    our approach also leverages the supervision from the caption corresponding to each video. 
    This is possible in our training data because we in fact have 5 elements (video1, caption1, video2, caption2, modification text)
    for each data sample.
    This new loss term %
    involves aligning the multi-modal query embedding $f(q,t)$ 
    not only with the video embedding $h(v)$ but also with its descriptive caption $c$. 
    To this end, we encode the caption into an embedding $g(c)$, using frozen BLIP-2 text encoding
    and define an additional contrastive loss term ($\mathcal{L}_c$).
	Therefore, our final loss can be expressed as:
}
\begin{align} 
	\mathcal{L}(\mathcal{B})=
	&\lambda_v \cdot \mathcal{L}_v(\mathcal{B}) + \lambda_c \cdot \mathcal{L}_c(\mathcal{B}),
\end{align}
\new{
	where $\lambda_v$ and $\lambda_c$ are both set to 0.5. 
	Here, $\mathcal{L}_v$ refers to the initial loss in Eq.(\ref{eq:Lv}) utilizing cosine similarity
	with the target video embedding $h(v_j)$, 
	and $\mathcal{L}_c$ employs the same loss structure but applies it to the cosine similarity with the target video caption embedding $g(c_j)$.
}

%% file: 4_experiments.tex
\section{Experiments}
\label{sec:experiments}

We first describe the experimental protocol including the datasets, evaluation metrics, and implementation details (Section~\ref{subsec:setup}).
We then present the results of CoVR on our new video benchmark (Section~\ref{subsec:covr_results}). %
Additionally, we introduce \ourCC, 
a new large-scale CoIR training dataset derived with a similar methodology,
from the Conceptual Captions (CC) dataset  (Section~\ref{subsec:cc-coir}).
\newt{We show transfer results of CoIR on standard image benchmarks,
together with an extensive state-of-the-art comparison
(Section~\ref{subsec:coir_results}).}
\newt{We further provide a comparison by training on other automatic triplet datasets (Section~\ref{subsec:synth}).}
Finally, we provide ablations on our key components
\newt{such as the caption retrieval loss and data scale}.
(Section~\ref{subsec:ablations}) 
and we illustrate qualitative examples in Section~\ref{subsec:qualitative}).

\subsection{Experimental setup}
\label{subsec:setup}

\noindent\textbf{Datasets.} 
\textbf{\ourWV} is our proposed training CoVR dataset, and \textbf{\ourWVt} is our
new CoVR benchmark, both presented in Section~\ref{subsec:data}.
\new{For pre-training on CoIR datasets, we use \textbf{\ourWVandCC}, a combination of both
\ourWV and \textbf{\ourCC} which is a new large-scale CoIR training dataset derived from the Conceptual Captions dataset, that we generate using the same methodology as \ourWV.
See Section~\ref{subsec:cc-coir} for more details.}

\textbf{CIRR}~\cite{cirr} is a manually annotated CoIR dataset that contains open-domain natural images from NLVR2~\cite{nlvr2}, comprising
36.5k queries annotated on 19k images.
CIRR includes two evaluation protocols: a standard one with the entire validation set as the search gallery, and a fine-grained \textit{subset}, where the search space is a subgroup of six images similar to the query image (based on pretrained ResNet15 feature distance).
The dataset is divided into training, validation, and testing splits with
28225/16742, 4181/2265 and 4148/2178 queries/images,
respectively.

\textbf{FashionIQ}~\cite{fashioniq} is another CoIR dataset that contains images of fashion products,
divided into three categories of Shirts, Dresses, and Tops/Tees. 
The query and target images were automatically paired based on title similarities (crawled from the web), and modification texts were then manually annotated.
This dataset consists of 30k queries annotated on 40.5k different images. 
It is divided into training and validation splits with 18000/45429 and 6016/15415 queries/images, respectively.

\new{
\textbf{CIRCO}~\cite{baldrati2023zero}
is an open-domain dataset for CoIR, 
specifically designed for zero-shot CoIR tasks,
as there is no specific training split provided for this dataset.
It is unique in its inclusion of multiple ground truths for each query, 
with an average of 4.53 ground-truth images per query. 
This feature allows for a more reliable and robust evaluation 
using the mean Average Precision (mAP) metric. 
CIRCO is divided into a validation split (220 queries) 
and a test split (800 queries). 
The evaluation protocol uses all 120,000 images from the COCO dataset as its gallery set.
}

\noindent\textbf{Evaluation metrics.} 
Following standard evaluation protocols~\cite{cirr}, we report the video retrieval recall at rank 1, 5, 10, and 50.
Recall at rank k (R@k) quantifies the number of times the correct video is among the top k results.
MeanR denotes the average of R@1, R@5, R@10, and R@50.
Higher recall means better performance.

\noindent\textbf{Implementation details and environmental costs.}
For our MTG-LLM, we use LLaMA 7B model~\cite{touvron2023llama} that we finetune for one epoch with an initial learning rate of $3\mathrm{e}{-5}$.
For our CoVR model, we use the BLIP-2 with ViT-G/14~\cite{fang2022eva} at 364 pixels finetuned for text-image retrieval on COCO and freeze the ViT for computational efficiency.
We train our CoVR model on \ourWV for 5 epochs with a batch size of 2048 and an initial learning rate of $1\mathrm{e}{-5}$.
To finetune on CIRR/FashionIQ, we train for 6 epochs with a batch size of 2048/1024 and an initial learning rate of $1\mathrm{e}{-4}$.
We set hyperparameters based on the validation curve of \ourWV.
Experiments are conducted on 4 NVIDIA A100-SXM4-80GB GPUs.
The experiments conducted in this study incurred an environmental cost of approximately 49kg of $CO_2$ emissions.
More details are included in \if\sepappendix1{Section~C} \else{Section~\ref{app:sec:implementation-details} }\fi
of the Appendix.

\input{tables/video}

\subsection{Composed video retrieval results}
\label{subsec:covr_results}
We provide a number of baselines for our new benchmark on \ourWVt.
Table~\ref{tab:video} summarizes these CoVR results.
We first report the random chance performance in the first row.
The rest of the table is split into two. The top block
uses existing pretrained text and image encoders
from BLIP~\cite{BLIP}, \new{BLIP-2~\cite{li2023blip2}} or CLIP~\cite{clip2021} backbones
without any finetuning. Models in the bottom block are
finetuned on \ourWV. We report results
with the composed query, as well as with the
individual modalities. For combining modalities,
we experiment with the simple average fusion baseline (Avg) when
using frozen embeddings, and fusion with
a randomly-initialized MLP or BLIP-pretrained cross-attention (CA)
layers when finetuning.
Note that the MLP fusion baseline is similar to Combiner~\cite{cclip}
that concatenates the image and text embeddings from CLIP (or BLIP in CASE~\cite{levy2023case}),
and is referred to as late fusion by CASE.
For finetuning individual modalities, we train and test
either with text-only query using the modification text,
or with the visual-only image query.
Finally, we experiment with using the weighted
average of target video frame embeddings as explained in Section~\ref{subsec:training}
(with the exception that visual-only experiments use equal weights
due to not having access to the modification text for computing the scores).

We make several conclusions.
(i)~Combining both visual and text modalities yields better
performance than the models with individual modalities.
This result highlights that our new CoVR benchmark requires
paying attention to both modalities.
(ii)~Visual-only outperforms text-only suggesting that
the video pairs automatically mined through their caption similarity
indeed exhibits visual similarity, and that the image captures
the target video better than the modification text.
(iii)~Finetuning on \ourWV obtains substantial improvements over
using pretrained and frozen embeddings.
(iv)~When finetuning, fusion with BLIP\new{-2} cross-attention (CA)
performs better than the MLP fusion.
(v)~Results with the BLIP\new{-2} backbone are higher than
those with CLIP \new{or BLIP}.
\new{We analyze the effect of the number of frames in \if\sepappendix1{Section~D.6} \else{Section~\ref{app:sec:optimal-number-of-frames}}\fi
.}

\subsection{Effect of training with \ourCC}
\label{subsec:cc-coir}
\new{
We apply the same automatic triplet generation procedure 
to the Conceptual Captions dataset~\cite{sharma2018CC3M} (CC3M)
resulting in the formation of \ourCC. 
This new dataset comprises 3.3M CoIR triplets, 
utilizing the diverse and contextually rich content of CC3M~\cite{{sharma2018CC3M}}.
In contrast to \ourWV, which is video-based, \ourCC exclusively incorporates images,
adding a different modality to our training material.
The average length of modification texts in this dataset is 24.65, featuring 130k unique images and 28k distinct modification texts. 
}

\new{
Table~\ref{tab:cc-coir} presents the results when training with our newly introduced \ourCC dataset.
When used independently for pretraining (zero-shot) we observe the following:
(a) \ourCC outperforms \ourWV on the CIRR dataset,
(b) \ourCC has similar performance than \ourWV in FashionIQ,
(c) \ourCC has lower performance for the CIRCO benchamark,
and (d) \ourCC performance has good zero-shot results on our video retrieval test set \ourWVt.
Note that \ourWVt constitutes a zero-shot setting only when trained using the \ourCC pretraining, 
whereas pretraining on \ourWV means the model has seen samples from the same data distribution during training.
By combining both \ourWV and \ourCC into a unified pretraining dataset \ourWVandCC, 
we observe a slight improvement on the image-based datasets while experiencing 
a minor drop on \ourWVt compared to using only \ourWV.
We hypothesize that this minor drop is due to a domain gap with \ourCC, 
which is included in \ourWVandCC.
We therefore opt to use the jointly pretrained model going forward, 
as it leverages the strengths of both data sources to perform well across different datasets (i.e., best average recall across all datasets).
}

\input{tables/cc-coir}

\subsection{State-of-the-art comparison on CoIR benchmarks}
\label{subsec:coir_results}
While our focus is video retrieval, we also experiment with transferring
our CoVR models to image retrieval tasks on standard CoIR benchmarks.
We define zero-shot CoIR as not using any manually annotated CoIR triplet for training.
To perform zero-shot CoIR, we directly apply our model, 
which has been trained on our automatically generated \ourWVandCC dataset, to CoIR tasks.
\new{In addition to zero-shot CoIR, for the datasets that have training split (CIRR and FashionIQ),
we investigate how our model performs when finetuned on the training set of the downstream benchmark.}

\input{tables/sota_cirr+fiq}

\input{tables/circo}
\input{tables/caption-retrieval-loss}

We report results in Table~\ref{tab:sota_cirr+fiq} on CIRR and Fashion-IQ datasets,
\new{
and in Table~\ref{tab:circo_test} for the newer CIRCO dataset.
In Table~\ref{tab:sota_cirr+fiq}, we report both finetuning (top block), and zero-shot (bottom block) settings.
In Table~\ref{tab:circo_test}, we only report the zero-shot setting since CIRCO does not have a training set to finetune on.
}
These results show that our model highly benefits from training on \new{\ourWVandCC}, especially in the zero-shot setting.
\new{Remarkably, our model attains state-of-the-art performance on CIRR and \new{CIRCO}, exhibiting a 5\% improvement in CIRR (43.74 vs 38.48 R@1) and a 2\% in CIRCO (28.29 vs 26.77 mAP@5) compared to the nearest competitor's results.
As for the FashionIQ dataset, we outperform all other methods but CompoDiff~\cite{gu2023compodiff} \newt{and LinCIR~\cite{gu2023languageonly} with the ViT-G backbone.}
With CompoDiff, the results are mixed depending on the metric (38.15 vs 39.02 R@1 and 58.44 vs 51.71 R@10).
\newt{For LinCIR, if we compare with the ViT-L backbone, more similar to ours,
our methods obtains better results (38.15 vs 26.28). }
However, CompoDiff/LinCIR perform poorly on the other two datasets CIRR (43.74 vs 26.71/35.25) and CIRCO (28.29 vs 15.33/19.71).
Our \ourMtwo remains therefore the overall best zero-shot model when evaluating across three datasets.
}
In addition to this strong zero-shot performance, 
our model reaches state-of-the-art performance when finetuned on both 
CIRR and FashionIQ benchmarks (top block of Table~\ref{tab:sota_cirr+fiq}).
\newt{For each setting, we also provide results with pretraining only on the \ourCC subset, as opposed to the full \ourWVandCC,
and observe similar performance on CIRR and FashionIQ, but lower on the challenging CIRCO dataset.}

\subsection{\new{Comparison with synthetic training images}} 
\label{subsec:synth}
\newt{
    To further compare against other approaches that propose automatic triplets for training
    (i.e., by generating synthetic target images \cite{brooks2022instructpix2pix,gu2023compodiff}),
    we train \textit{our} \ourMtwo model on \textit{their} data in a controlled experiment.
    We summarize the results in Table~\ref{tab:different-datasets}.
    In the top block, we show numbers directly taken from the CompoDiff~\cite{gu2023compodiff} work,
    comparing their method trained on various synthetic datasets: InstructPix2Pix dataset of 1-million triplets~\cite{brooks2022instructpix2pix}, a 1M subset of their SynthTriplets~\cite{gu2023compodiff},
    and the full 18M version. Note that we use the same metrics as in their paper (e.g., average of R@1, R$_s$@1 for CIRR)
    and show the results with their ViT-L model.
    In the bottom block, we train our model on the 1M synthetic dataset versions, as well as a 1M subset of our real visual data from \ourWVandCC.
    When comparing the two models on the same training data, we observe similar performances
    (e.g., 28.32 vs 28.44 on CIRR), suggesting that the main difference comes from training data,
    rather than the model. Training on our \ourWVandCC with 1M triplets outperform 
    SynthTriplets with both 1M and 18M versions, highlighting the importance of real training images.
}

\input{tables/different-datasets}

\input{tables/mtg-llm}

\subsection{Ablation studies} 
\label{subsec:ablations}

We now ablate the importance of several key aspects of our method by
focusing primarily on experiments trained in \ourWV, 
and also examine the impact of different datasets on model performance.

\noindent\new{\textbf{The additional composed caption retrieval loss.}
	As explained in Section~\ref{subsec:training},
	we integrate a caption retrieval loss term as additional supervision.
	For both methods (\ourMone~\cite{ventura23covr} and \ourMtwo), this led to a significant improvement in the CoIR performance and a slight decrease on \ourWV on which it was originally trained on, see Table~\ref{tab:caption-retrieval-loss} (34 vs 41 R@1 on CIRR and 27 vs 36 R@1 on FashionIQ for instance).
}

\noindent\textbf{Importance of data scale.}
In Table~\ref{tab:size}, we evaluate the effect of the number of video-caption pairs used as a seed for our triplet generation pipeline.
We construct subsets of videos such that larger ones include smaller ones, and only keep triplets that contain the sampled videos for training.
We find that results steadily increase when using more videos, demonstrating that our method largely benefits from scaling the size of the seed dataset of video-captions.
We also observe the importance of the filtering techniques described in Section~\ref{subsec:gen}, as the model trained on unfiltered data underperforms.

\input{tables/size}

\noindent\textbf{Modification text generation.} 
We use a large language model finetuned for modification text generation (MTG-LLM) as explained in Section~\ref{subsec:gen}.
We here compare this solution 
\new{to prompting it without any training}
and to a simple rule-based baseline that uses several templates to generate the modification text given the two captions that differ by one word.
\new{For prompting, we prepend few-shot examples of pairs of captions and desired generated texts, before 
adding the two captions in question. 
Please check \if\sepappendix1{Section~C.5} \else{Section~\ref{app:subsec:prompting-llm} }\fi
of the Appendix for the full prompt.
Table~\ref{tab:mtg-llm} shows that finetuning the MTG-LLM for generating the training data is much more effective than prompting it without finetuning, as measured by CoVR performance on \ourWVt and CoIR performance on CIRR.
}
For the rule-based experiment, the modification text is based on the
two different words from the captions.
We generate templates that use these words and choose one at random during training.
These templates include variations such as \textit``{Remove $\mathtt{txt\_diff_1}$''} and \textit{ "Change $\mathtt{txt\_diff_1}$ for $\mathtt{txt\_diff_2}$''}.
A full list of all the templates can be seen in \if\sepappendix1{Section~C.3} \else{Section~\ref{app:subsec:rule-based} }\fi
of the Appendix.
Additionally, we investigate the possibility of paraphrasing the rule-based modification texts using \texttt{GPT-3.5-turbo} from OpenAI~\cite{GPT_3} as a source of augmentation,
by prompting \textit{``Paraphrase the following sentence: \texttt{\{Rule-base modification text\}}''}.
In preliminary analysis, we qualitatively observed that LLaMA~\cite{touvron2023llama} and LLaMA 2~\cite{touvron2023llama2}
alternatives were overly verbose when used for paraphrasing; however, GPT-3.5 outputs were satisfactory. 

In Table~\ref{tab:mtg-llm}, we show that our MTG-LLM generates better modification texts than the rule-based baseline, by evaluating the results of the model trained on the generated data.
Paraphrasing the rule-based examples significantly boosts the performance 
(from 39 to 57.9 R@1), while still being worse than our MTG-LLM, especially on the CIRR benchmark.
Note that the paraphrasing comes with the cost of running an expensive LLM (\$43 cost for this experiment for 1 paraphrasing per modification text on the entire dataset). On the other hand,
our MTG-LLM finetuning only requires 715 text examples.
Qualitative examples comparing MTG-LLM and rule-based are provided in \if\sepappendix1{Table~A.11} \else{Table~\ref{tab:rule-based-comparision} }\fi
 of the Appendix.

\input{tables/training-strategies}

\noindent\textbf{Training strategies.} 
In Table~\ref{tab:training-strategies}, we first show the benefit on \ourWV of training by iterating on target videos instead of CoVR triplets.
This is to avoid having the same target video appearing multiple times in a training batch, hence increasing the number of correct negatives that are used in the contrastive loss.
Furthermore, up-sampling hard negatives adopting the HN-NCE loss formulation from \cite{hnnce2023}
also slightly benefits the 
performance.

\subsection{\new{Qualitative analysis}} 
\label{subsec:qualitative}
\begin{figure*}%
  \centering
  \includegraphics[width=.95\linewidth]{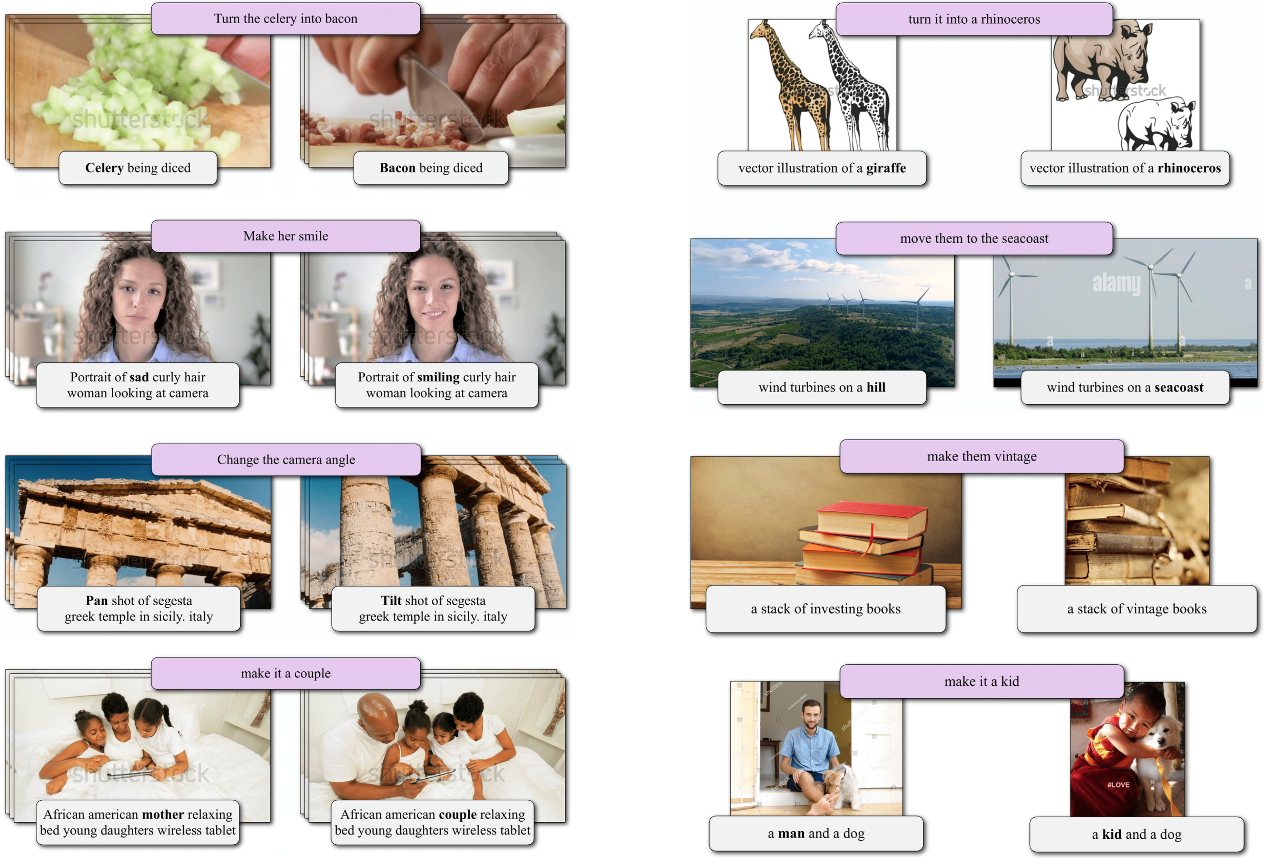}
  \caption{\textbf{Examples of generated triplets:} 
  \new{We illustrate triplet samples generated using our automatic dataset creation methodology 
  (left: \ourWV, right: \ourCC). 
  Each sample consists of two videos/images with their corresponding captions (at the bottom of each video/image)
  and the generated modification text using our MTG-LLM (in purple).}}
  \label{app:fig:qualitative}
\end{figure*}

\new{In this section, we provide 
qualitative examples of our \ourWV and \ourCC triplets. 
Figure~\ref{app:fig:qualitative}, shows examples of triplets generated using our automatic dataset creation.
These examples demonstrate the effectiveness of our approach 
in generating coherent modification texts for paired videos.
For more examples, we refer to \if\sepappendix1{Section~E.5}\else{Section~\ref{app:subsec:recall-webvid}}\fi} of the Appendix.

%% file: tables/video.tex
\begin{table*} %
\centering
 \caption{\textbf{Benchmarking on the \ourWVt set:} 
	We observe that using both the visual and text input modalities performs
	better than individual modalities alone, both with/without
	finetuning on \ourWV (shown at the top/bottom of the table, respectively).
	When using pretraining models without finetuning, we apply
	average fusion (Avg) for the embeddings. BLIP performs slightly better
	than CLIP on this benchmark.
	Finetuning on \ourWV brings significant benefits.
	In this case, fusing with the pretrained cross-attention (CA) from BLIP is more effective
	than training a randomly-initialized MLP fusion as done in \cite{cclip}.
	Moreover, using multiple frames to embed the target video brings
	further improvements over using the middle frame.
	The first row represents the random performance.
}
\setlength\tabcolsep{5pt}
    \resizebox{0.70\linewidth}{!}{
    \begin{tabular}{clll|cccc}
        \toprule
        & Input &  &  & \multicolumn{4}{c}{\ourWVt} \\ 
        & modalities & Fusion & Backbone & R@1 & R@5 & R@10 & R@50 \\ 
        \midrule
        Random & - & - & - &\textcolor{white}{0}0.08 & \textcolor{white}{0}0.23 & \textcolor{white}{0}0.35 & \textcolor{white}{0}1.76 \\
        \midrule
        & Text & - & BLIP-2 & 18.74 & 38.11 & 47.97 & 67.21 \\
        Not finetuned on & Visual & - &  BLIP-2 & 33.10 & 59.55 & 69.80 & 88.85 \\
        \ourWV & Visual + Text & Avg & CLIP & 44.37 & 69.13 & 77.62 & 93.00  \\
        & Visual + Text & Avg & BLIP & 45.46 & 70.46 & 79.54 & 93.27 \\
        & Visual + Text & Avg & BLIP-2 & \textbf{45.66} & \textbf{71.71} & \textbf{81.30} & \textbf{94.80} \\
        \midrule
        & Text & - & BLIP-2 & 23.32 & 46.21 & 56.22 & 78.36 \\
        & Visual & - & BLIP-2 & 36.03 & 64.24 & 74.77 & 92.64 \\ %
        & Visual + Text & MLP & CLIP &  50.86 & 77.46 & 85.32 & 96.75 \\
        Finetuned on & Visual + Text & MLP & BLIP & 50.59 & 74.65 & 83.57 & 95.46 \\
        \ourWV & Visual + Text & MLP & BLIP-2 & 51.88 & 79.38 & 86.46 & 97.42 \\
        & Visual + Text & CA & BLIP & 55.95 & 81.22 & 89.05 & 98.08 \\
        & Visual + Text & CA & BLIP-2 &  \textbf{59.82} & \textbf{83.84} & \textbf{91.28} & \textbf{98.24} \\
    \bottomrule
    \end{tabular}
}
\label{tab:video}
\end{table*}

%% file: tables/cc-coir.tex
\begin{table} %
\centering
 \caption{\new{\textbf{Training with \ourCC:} 
        We compare the performance metrics when pretraining with various combinations of 
        \ourWV and \ourCC datasets. 
        We observe that in the zero-shot setting, combining both datasets yields the best overall results.
        }
	}
	\setlength\tabcolsep{5pt}
	\resizebox{0.99\linewidth}{!}{
		\begin{tabular}{cc|c|c|c|c}
        \toprule 
         WebVid- & CC- & WebVid-CoVR-T. & CIRR & FashionIQ & CIRCO \\
         CoVR & CoIR & R@1 & R@1 & R@10 & mAP@5 \\
            \toprule 
            \xmark & \cmark & 53.83 & 43.35 &  36.49 & 23.50 \\
            \cmark & \xmark & \textbf{59.82} & 41.42 & 36.81 & \textbf{28.88} \\
            \cmark & \cmark & 57.71 & \textbf{43.74} & \textbf{38.15} & 28.29\\
            \bottomrule
		\end{tabular}
	}
	\label{tab:cc-coir}
\end{table}

%% file: tables/sota_cirr+fiq.tex
\begin{table*} [t]
\caption{\textbf{State-of-the-art comparison on the CIRR and FashionIQ test sets:} Our model benefits from training on our datasets in the zero-shot setting, and in the finetuning setting where it performs competitively.
	$\dagger$ denotes results reported by~\cite{cirr}.
	For methods that pretrain specifically for composed retrieval, we indicate their pretraining data.
	CC3M denotes Conceptual Captions 3M~\cite{sharma2018CC3M}, ST18M denotes ~\cite{gu2023compodiff}, and COCO denotes~\cite{coco}.
}
\setlength\tabcolsep{5pt}
    \resizebox{\linewidth}{!}{
    \begin{tabular}{clll|cccc|ccc||cc|cc|cc|cc}
        \toprule
        \multirow{3}{*}{\begin{tabular}[c]{@{}c@{}}\rotatebox{90}{Mode}\end{tabular}} &&&& \multicolumn{7}{c||}{\textbf{CIRR}} & \multicolumn{8}{c}{\textbf{FashionIQ}}\\
        & &  & Pretraining  & \multicolumn{4}{c|}{Recall@K} & \multicolumn{3}{c||}{$\text{R}_{\text{subset}}$@K} & \multicolumn{2}{c}{Dress} & \multicolumn{2}{c}{Shirt} &  \multicolumn{2}{c|}{Toptee} & \multicolumn{2}{c}{Average}  \\
        & Method & Year & Data & K=1 & K=5 & K=10 & K=50 & K=1 & K=2 & K=3 & R@10 & R@50 & R@10 & R@50 & R@10 & R@50 & R@10 & R@50 \\ 
        \midrule
        \multirow{28.2}{*}{
        \begin{tabular}[c]{@{}c@{}}
        \rotatebox{90}{Train}
        \end{tabular}}
        & TIRG~\cite{tirg}$\dagger$ & 2019 & - & 14.61 & 48.37 & 64.08 & 90.03 & 22.67 & 44.97 & 65.14 & 14.87 & 34.66 & 18.26 & 37.89 & 19.08 & 39.62 & 17.40 & 37.39 \\ %
        & TIRG+LastConv~\cite{tirg} $\dagger$ & 2019 & - & 11.04 & 35.68 & 51.27 & 83.29 & 23.82 & 45.65 & 64.55 & - & - & - & - & - & - & - & -  \\ %
        & CurlingNet~\cite{CurlingNet}$\ddagger$ & 2019 & - & - & - & - & - & - & - & - & 26.15 & 53.24 & 21.45 & 44.56 & 30.12 & 55.23 & 25.90 & 51.01 \\
        & JVSM~\cite{JVSM} & 2020 & - & - & - & - & - & - & - & - & 10.70 & 25.90 & 12.00 & 27.10 & 13.00 & 26.90 & 11.90 & 26.60 \\
        & TRACE w/BER~\cite{TRACE} & 2020 & - & - & - & - & - & - & - & - & 22.70 & 44.91 & 20.80 & 40.80 & 24.22 & 49.80 & 22.57 & 46.19 \\
        & VAL w/GloVe~\cite{VAL_IR} & 2020 & - & - & - & - & - & - & - & - & 22.53 & 44.00 & 22.38 & 44.15 & 27.53 & 51.68 & 24.15 & 46.61 \\
        & MAAF~\cite{MAAF}$\dagger$ & 2020 & - & 10.31 & 33.03 & 48.30 & 80.06 & 21.05 & 41.81 & 61.60 & 23.80 & 48.60 & 21.30 & 44.20 & 27.90 & 53.60 & 24.30 & 48.80 \\
        & MAAF-BERT~\cite{MAAF}$\dagger$ & 2020 & - & 10.12 & 33.10 & 48.01 & 80.57 & 22.04 & 42.41 & 62.14 & - & - & - & - & - & - & - & - \\
        & MAAF-IT~\cite{MAAF}$\dagger$ & 2020 & - & \textcolor{white}{0}9.90 & 32.86 & 48.83 & 80.27 & 21.17 & 42.04 & 60.91 & - & - & - & - & - & - & - & - \\
        & MAAF-RP~\cite{MAAF}$\dagger$ & 2020 & - & 10.22 & 33.32 & 48.68 & 81.84 & 21.41 & 42.17 & 61.60 & - & - & - & - & - & - & - & - \\
        & RTIC-GCN~\cite{RTIC}$\ddagger$ & 2021 & - & - & - & - & - & - & - & - & 29.15 & 54.04 & 23.79 & 47.25 & 31.61 & 57.98 & 28.18 & 53.09 \\
        & CoSMo~\cite{Lee_2021_CVPR_cosmo} & 2021 & - & - & - & - & - & - & - & - & 25.64 & 50.30 & 24.90 & 49.18 & 29.21 & 57.46 & 26.58 & 52.31 \\
        & DCNet~\cite{Kim_Yu_Kim_Kim_2021_dcnet} & 2021 & - & - & - & - & - & - & - & - & 28.95 & 56.07 & 23.95 & 47.30 & 30.44 & 58.29 & 27.78 & 53.89 \\
        & CIRPLANT~\cite{cirr}$\dagger$ & 2021 & - & 19.55 & 52.55 & 68.39 & 92.38 & 39.20 & 63.03 & 79.49 & 17.45 & 40.41 & 17.53 & 38.81 & 61.64 & 45.38 & 18.87 & 41.53\\
        & SAC w/BERT~\cite{SAC} & 2022 & - & - & - & - & - & - & - & - & 26.52 & 51.01 & 28.02 & 51.86 & 32.70 & 61.23 & 29.08 & 54.70 \\
        & FashionVLP~\cite{FashionVLP} & 2022 & - & - & - & - & - & - & - & - & 32.42 & 60.29 & 31.89 & 58.44 & 38.51 & 68.79 & 34.27 & 62.51 \\
        & ARTEMIS~\cite{ARTEMIS} & 2022 & - & 16.96 & 46.10 & 61.31 & 87.73 & 39.99 & 62.20 & 75.67 & 27.16 & 52.40 & 21.78 & 43.64 & 29.20 & 53.83 & 26.05 & 50.29  \\
        & LF-CLIP~\cite{cclip} $\dagger$ & 2022 & - & 33.59 & 65.35 & 77.35 & 95.21 & 62.39 & 81.81 & 92.02 & 31.63 & 56.67 & 36.36 & 58.00 & 38.19 & 62.42 & 35.39 & 59.03\\
        & LF-BLIP~\cite{cclip,levy2023case} $\dagger$ & 2022 & - & 20.89 & 48.07 & 61.16 & 83.71 & 50.22 & 73.16 & 86.82 & 25.31 & 44.05 & 25.39 & 43.57 & 26.54 & 44.48 & 25.75 & 43.98\\
        & Combiner~\cite{cclip} & 2022 & - & 33.59 & 65.35 & 77.35 & 95.21 & 62.39 & 81.81 & 92.02 & 31.63 & 56.67 & 36.36 & 58.00 & 38.19 & 62.42 & 35.39 & 59.03 \\
        & CompoDiff~\cite{gu2023compodiff} & 2023 & ST18M+LAION2B & 32.39 & 57.61 & 77.25 & 94.61 & 67.88 & 85.29 & 94.07 & 38.39 & 51.03 & 41.68 & 56.02 & 45.70 & 57.32 & 39.81 & 51.90 \\ %
        & CASE~\cite{levy2023case} & 2024 & LaSCo~\cite{levy2023case} & 48.68 & 79.98 & 88.51 & {97.49} & 76.39 & 90.12 & {95.86} & \textbf{47.44} & 69.36 & 48.48 & 70.23 & 50.18 & 72.24 & 48.79 & 70.68\\
        & CASE~\cite{levy2023case} & 2024 & LaSCo+COCO & {49.35} & {80.02} & {88.75} & 97.47 & {76.48} & {90.37} & 95.71 & - & - & - & - & - & - & - & - \\
        & \ourMone~\cite{ventura23covr} & 2024 & \ourWV & {49.69} & 78.60 & 86.77 & 94.31 & 75.01 & 88.12 & 93.16 & 44.55 & 69.03 & 48.43 & 67.42 & \textbf{52.60} & \textbf{74.31} & 48.53 & 70.25 \\
        \cmidrule(lr){2-19}
        & \ourMtwo & 2024 & - & \textbf{50.87} & 80.80 & 88.84 & 98.00 & 76.70 & 90.31 & 95.45 & 
        45.25 & 68.86 & 49.95 & 69.95 & 51.37 & 72.56 & 48.86 & 70.46 \\
        & \newt{\ourMtwo }& \newt{2024} & \newt{\ourCC} & \newt{50.63} & \newt{81.04} & \newt{\textbf{89.35}} & \newt{\textbf{98.15}} &  \newt{76.53} & \newt{\textbf{90.43}} & \newt{\textbf{96.00}} & \newt{46.49} & \newt{69.31} & \newt{\textbf{51.67}} & \newt{70.49} & \newt{51.53} & \newt{73.07} & \newt{49.90} & \newt{70.96} \\
        & \ourMtwo & 2024 & \ourWVandCC & {50.43} & {\textbf{81.08}} & {88.89} & {98.05} & {\textbf{76.75}} & {90.34} & {95.78} & 46.53 & \textbf{69.60} & 51.23 & \textbf{70.64} & {52.14} & 73.27 & \textbf{49.96} & \textbf{71.17} \\
        \midrule \\ \midrule
        \multirow{16.2}{*}{\begin{tabular}[c]{@{}c@{}} 
        \rotatebox{90}{Zero Shot}
        \end{tabular}} 
        & Random$\dagger$ &  & - & \textcolor{white}{0}0.04 & \textcolor{white}{0}0.22 & \textcolor{white}{0}0.44 & \textcolor{white}{0}2.18 & 16.67 & 33.33 & 50.00 & \textcolor{white}{0}0.26 & \textcolor{white}{0}1.31 & \textcolor{white}{0}0.16 & \textcolor{white}{0}0.79 & \textcolor{white}{0}0.19 & \textcolor{white}{0}0.95 & \textcolor{white}{0}0.06 & \textcolor{white}{0}0.32 \\
        & CompoDiff~\cite{gu2023compodiff} & 2023 & ST18M+LAION2B & 26.71 & 55.14 & 74.52 & 92.01 & 64.54 & 82.39 & 91.81 & \textbf{37.78} & 49.10 & 41.31 & 55.17 & \textbf{44.26} & 56.41 & \textbf{39.02} & 51.71 \\ %
        & Pic2Word~\cite{saito2023pic2word} & 2023 & CC3M & 23.90 & 51.70 & 65.30 & 87.80 & - & - & - & 20.00 & {40.20} & 26.20 & 43.60 & 27.90 & 47.40 & 24.70 & 43.70 \\
        & SEARLE-XL-OTI~\cite{baldrati2023zero} & 2023 & CC3M & 24.87 & 52.31 & 66.29 & 88.58 & 53.80 & 74.31 & 86.94 & 21.57 & 44.47 & 30.37 & 47.49 & 30.90 & 51.76 & 27.61 & 47.90 \\
        & SEARLE-XL~\cite{baldrati2023zero} & 2023 & CC3M & 24.24 & 52.48 & 66.29 & 88.84 & 53.76 & 75.01 & 88.19 & 26.89 & 45.58 & 20.48 & 43.13 & 29.32 & 49.97 & 25.56 & 46.23 \\
        & GRB~\cite{sun2023trainingfree} & 2023 & - & 24.19 & 52.07 & 65.48 & 85.28 & 60.17 & 79.13 & 89.34 & 24.14 & 45.56 & 34.54 & 55.15 & 33.55 & 53.60 & 30.74 & 51.44 \\
        & GRB+LCR~\cite{sun2023trainingfree} & 2023 & - & 30.92 & 56.99 & 68.58 & 85.28 & 66.67 & 78.68 & 82.60 & - & - & - & - & - & - & - & - \\
        & \newt{TFCIR}~\cite{sun2023trainingfree} & \newt{2023} & \newt{-} & \newt{32.82} & \newt{61.13} & \newt{71.76} & \newt{85.28} & \newt{66.63} & \newt{78.58} & \newt{82.68} & \newt{24.84} & \newt{45.56} & \newt{35.38} & \newt{55.15} & \newt{33.10} & \newt{53.60} & \newt{31.11} & \newt{51.44}  \\
        & CASE~\cite{levy2023case} & 2024 & LaSCo & 30.89 & 60.75 & 73.88 & 92.84 & 60.17 & 80.17 & 90.41 & - & - & - & - & - & - & - & - \\
        & CASE~\cite{levy2023case} & 2024 & LaSCo+COCO & 35.40 & 65.78 & {78.53} & {94.63} & 64.29 & 82.66 & {91.61} & - & - & - & - & - & - & - & - \\
        & \newt{CIReVL}~\cite{karthik2023visionbylanguage} & \newt{2024} & \newt{-} & \newt{34.65} & \newt{64.29} & \newt{75.06} & \newt{91.66} & \newt{67.95} & \newt{84.87} & \newt{93.21} & \newt{27.07} & \newt{49.53} & \newt{33.71} & \newt{51.42} & \newt{35.80} & \newt{56.14} & \newt{32.19} & \newt{52.36} \\
        & \newt{LinCIR (ViT-L)}~\cite{gu2023languageonly} & \newt{2024} & \newt{-} & \newt{25.04} & \newt{53.25} & \newt{66.68} & \newt{-} & \newt{57.11} & \newt{77.37} & \newt{88.89} & \newt{20.92} & \newt{42.44} & \newt{29.10} & \newt{46.81} & \newt{28.81} & \newt{50.18} & \newt{26.28} & \newt{46.49} \\
        & \newt{LinCIR (ViT-G)}~\cite{gu2023languageonly} & \newt{2024} & \newt{-} & \newt{35.25} & \newt{64.72} & \newt{76.05} & \newt{-} & \newt{63.35} & \newt{82.22} & \newt{91.98} & \newt{\textbf{38.08}} & \newt{\textbf{60.88}} & \newt{\textbf{46.76}} & \newt{\textbf{65.11}} & \newt{\textbf{50.48}} & \newt{\textbf{71.09}} & \newt{\textbf{45.11}} & \newt{\textbf{65.69}} \\
        & \ourMone~\cite{ventura23covr} & 2024 & \ourWV & {38.48} & {66.70} & 77.25 & 91.47 & {69.28} & {83.76} & 91.11 & {21.95} & {39.05} & {30.37} & {46.12} & {30.78} & {48.73} & {27.70} & {44.63} \\
        \cmidrule(lr){2-19} 
        & \newt{\ourMtwo} & \newt{2024} & \newt{\ourCC} & \newt{43.35} & \newt{\textbf{73.78}} & \newt{83.66} & \newt{96.07} & \newt{\textbf{75.25}} & \newt{\textbf{88.89}} & \newt{\textbf{95.23}} & \newt{33.17} & \newt{55.13} & \newt{38.81} & \newt{58.24} & \newt{37.48} & \newt{58.49} & \newt{36.49} & \newt{57.29} \\
        \rowcolor{ourcolor!64} \cellcolor{white} 
        & \ourMtwo & 2024 & \ourWVandCC & \textbf{43.74} & 73.61 & \textbf{83.95} & \textbf{96.10} & 72.84 & 87.52 & 94.39 & 34.26 & 56.22 & 41.22 & 59.32 & 38.96 & 59.77 & 38.15 & 58.44 \\
    \bottomrule
\end{tabular}
}
\label{tab:sota_cirr+fiq}
\end{table*}

%% file: tables/circo.tex
\begin{table} [t]
    \caption{\new{\textbf{State-of-the-art comparison on the CIRCO test set:} Our model benefits from training on \ourWVandCC in the zero-shot setting.
    For methods that pretrain specifically for composed retrieval, we indicate their pretraining data.
    CC3M denotes Conceptual Captions 3M~\cite{sharma2018CC3M} and ST18M denotes~\cite{gu2023compodiff}.
    }
    }
\setlength\tabcolsep{5pt}
    \resizebox{\linewidth}{!}{
    \begin{tabular}{lll|cccc}
        \toprule
        && Pretraining  & \multicolumn{4}{c}{mAP@K} \\
        Method & Year & Data & K=5 & K=10 & K=25 & K=50 \\ 
        \midrule
        Random &  & - & \textcolor{white}{0}0.00 & \textcolor{white}{0}0.00 & \textcolor{white}{0}0.00 & \textcolor{white}{0}0.00 \\
        SEARLE-XL~\cite{baldrati2023zero} & 2023 & CC3M & 11.68 & 12.73 & 14.33 & 15.12 \\
        CompoDiff (ViT-G)~\cite{gu2023compodiff} & 2023 & ST18M & 15.33 & 17.71 & 19.45 & 21.01 \\
        GRB+LCR~\cite{sun2023trainingfree} & 2023 & - & 25.38 & 26.93 & 29.82 & 30.74 \\
        \newt{TFCIR}~\cite{sun2023trainingfree} & \newt{2023} & \newt{-} & \newt{26.52} & \newt{28.25} & \newt{31.23} & \newt{31.99} \\
        CIReVL~\cite{karthik2023visionbylanguage} & 2024 & - & 26.77 & 27.59 & 29.96 & 31.03 \\
        \newt{LinCIR (ViT-G)}~\cite{gu2023languageonly} & \newt{2024} & \newt{-} & \newt{19.71} & \newt{21.01} & \newt{23.13} & \newt{24.18} \\
        \ourMone~\cite{ventura23covr} & 2024 & \ourWV & 21.43 & 22.33 & 24.47 & 25.48 \\
        \cmidrule(lr){0-6}
        \newt{\ourMtwo} & \newt{2024} & \newt{\ourCC} & \newt{23.50} & \newt{24.66} & \newt{27.05} & \newt{28.04}\\
        \rowcolor{ourcolor!64}
        \ourMtwo & 2024 & \ourWVandCC & \textbf{28.29} & \textbf{29.55} & \textbf{32.18} & \textbf{33.26} \\
    \bottomrule
\end{tabular}
}
\label{tab:circo_test}
\end{table}

%% file: tables/caption-retrieval-loss.tex
\begin{table} %
	\caption{
		\new{\textbf{Adding the composed caption retrieval loss:}
        We compare models with and without using our additional caption retrieval loss,
        on both \ourMone~\cite{ventura23covr} and \ourMtwo. 
        We observe a significant improvement %
        on CoIR datasets when training on the \ourWV with this additional loss. 
        }
	 }
    \centering
    \resizebox{0.99\linewidth}{!}{
    \begin{tabular}{lc|c|c|c|c}
        \toprule 
        & Caption & WebVid-CoVR-T. & CIRR  & FashionIQ & CIRCO \\
        Method & ret.~loss & R@1 & R@1 & R@10 & mAP@5 \\ 
        \midrule
        \ourMone & \xmark & 56.81 & 37.18 & 23.20 &  22.86 \\
        \ourMone & \cmark & 55.95 & 40.46 & 33.90 & 25.71 \\
        \midrule
        \ourMtwo & \xmark & \textbf{60.09} & 34.43 & 27.78 & 25.06 \\
        \ourMtwo & \cmark & 59.82 & \textbf{41.42} & \textbf{36.81} & \textbf{27.84} \\
        \bottomrule
    \end{tabular}
    }
    \label{tab:caption-retrieval-loss}
\end{table}

%% file: tables/different-datasets.tex
\begin{table} %
    \centering
    \caption{
        \textbf{
            \newt{Synthetic vs real training images:}
        }
        \newt{
            We compare the CoIR performance of our proposed method (\ourMtwo) and dataset (\ourWVandCC) against the CompoDiff~\cite{gu2023compodiff} method and their proposed dataset (SynthTriplets).
            The results demonstrate that our training data achieves better performance with 1M triplets compared to 1M or even 18M triplets of the SynthTriplets dataset, containing synthetically generated target images.
            IP2P denotes InstructPix2Pix, 1M public synthetic triplets by \cite{brooks2022instructpix2pix}.
        }
    }
	\setlength\tabcolsep{5pt}
	\resizebox{0.99\linewidth}{!}{
		\begin{tabular}{lll|c|c|c}
        \toprule 
         & Pretraining & Data & WebVid-CoVR-T & CIRR & FashionIQ \\
         Model & Data & Size & {\tiny R@1} & {\tiny Avg(R@1, R$_s$@1)} & {\tiny Avg(R@10, R@50)} \\
            \toprule 
            \multirow{3}{*}{CompoDiff~\cite{gu2023compodiff}} & IP2P~\cite{brooks2022instructpix2pix} & 1M & - & 27.42 & 27.24 \\
            & SynthTriplets~\cite{gu2023compodiff} & 1M & - & 28.32 & 31.91 \\
            & SynthTriplets~\cite{gu2023compodiff} & 18M & - & 37.83 & 42.33 \\
            \midrule
            \multirow{3}{*}{\ourMtwo} & IP2P~\cite{brooks2022instructpix2pix} & 1M & 16.55 & 34.72 & 15.42 \\ %
            & SynthTriplets~\cite{gu2023compodiff} & 1M & 45.42 & 28.44 & 33.23 \\ %
            & WV-CC-CoVIR & 1M & \textbf{55.87} & \textbf{58.79} & \textbf{48.14} \\ %
            \bottomrule
		\end{tabular}
	}
	\label{tab:different-datasets}
\end{table}

%% file: tables/mtg-llm.tex
\begin{table*} %
	\caption{\textbf{Modification text generation:} 
		We compare our finetuned model MTG-LLM 
		(LLaMA 7B parameters)
		against (a) a rule-based MTG baseline, (b) a paraphrased rule-based MTG baseline
		(using GPT-3.5-turbo from OpenAI), and \new{(c) simply prompting the frozen LLaMA LLM}.
		We observe important gains in the downstream performance of the model trained on the generated data.
	}
    \centering
    \begin{tabular}{l|cccc|cccc}
        \toprule
        & \multicolumn{4}{c|}{\ourWVt}  & \multicolumn{4}{c}{CIRR}\\
        Model & R@1 & R@5 & R@10 & R@50 & R@1 & R@5 & R@10 & R@50 \\ 
        \midrule
        Rule-based &  39.08 & 67.33 & 78.13 & 93.82 & 12.02 & 34.75 & 47.06 & 75.35 \\
        Rule-based \new{\& GPT-}paraphrased & 57.94 & 81.85 & 89.79 & 97.93 & 32.89 & 61.98 & 73.04 & 90.34 \\
        \new{Prompting LLaMA} & 56.46 & 82.08 & 89.32 & 97.85 & 34.27 & 64.29 & 75.76 & 91.61 \\
        \rowcolor{ourcolor!64}
        MTG-LLM & \textbf{59.82} & \textbf{83.84} & \textbf{91.28} & \textbf{98.24} & \textbf{41.42} & \textbf{72.58} & \textbf{82.55} & \textbf{96.31} \\
        \bottomrule
    \end{tabular}
    \label{tab:mtg-llm}
    \end{table*}

%% file: tables/size.tex
\begin{table*} %
	\caption{\textbf{Data size:} We %
		measure the importance of the number of videos used for data generation and of filtering the generated data, by evaluating
		on \ourWVt, CIRR, and FashionIQ. 
		All models are trained for the same number of iterations on the generated data.
	}
    \centering
    \begin{tabular}{rrr|r|c|cc|cc|c}
        \toprule
         \textit{Initial} & \textit{Generated} & & \multicolumn{2}{c|}{\ourWVt}  & 
         \multicolumn{2}{c|}{CIRR} & 
         \multicolumn{2}{c|}{FashionIQ} & CIRCO \\
        \#videos & \#triplets & Filtering & R@1 & MeanR & R@1 & MeanR  &  R@10 & R@50 & mAP@5 \\
        \midrule
        0 & - & - & 16.55 & 36.15 & 18.60 & 47.82 & \textcolor{white}{0}9.75 & 21.09 & \textcolor{white}{0}4.83 \\
        \midrule
        200k & 11k & \cmark & 51.53 & 77.92 & 40.72 & 70.82 & 34.51 & 55.82 & 24.00  \\
        500k & 66k & \cmark & 55.13 & 80.59 & 40.22 & 70.67 & 36.04 & 57.10 & 25.86 \\
        1M & 269k & \cmark &  57.32 & 82.13 & 40.55 & 71.08 & 36.76 & 57.13 &  26.92 \\
        \rowcolor{ourcolor!64}
        2.5M & 1.6M & \cmark & \textbf{59.82} & \textbf{83.30} & \textbf{41.42} & \textbf{73.22} & 36.81 & 56.70 & 27.84 \\
        \midrule
        2.5M & 3.6M & \xmark & 58.65 & 82.57 & 40.84 & 71.11 & \textbf{37.19} & \textbf{57.19} & \textbf{29.11} \\
        \bottomrule
    \end{tabular}
    \label{tab:size}
\end{table*}

%% file: tables/training-strategies.tex
\begin{table*} %
	\caption{\textbf{Training strategies:} 
		Iterating on batches of distinct target videos (instead of triplets) and up-sampling hard negatives both benefit the %
		CoVR/CoIR performance. 
	}
    \centering
    \begin{tabular}{cc|cccc|cccc}
        \toprule 
        & HN-NCE & \multicolumn{4}{c|}{\ourWVt}  & \multicolumn{4}{c}{CIRR}\\
        Iteration & \cite{hnnce2023} & R@1 & R@5 & R@10 & R@50 & R@1 & R@5 & R@10 & R@50 \\ 
        \midrule
        Triplets & \cmark & 53.79 & 81.69 & 88.97 & 97.89 & 37.49 & 66.63 & 77.11 & 91.28 \\
        Videos  & \xmark & 55.24 & 81.34 & 89.48 & 98.24 & 38.70 & 68.84 & 
        78.41 & 92.72 \\
        \rowcolor{ourcolor!64} Videos & \cmark &  \textbf{59.82} & \textbf{83.84} & \textbf{91.28} & \textbf{98.24} & \textbf{41.42} & \textbf{72.58} & \textbf{82.55} & \textbf{96.31}  \\
        \bottomrule
    \end{tabular}
    \label{tab:training-strategies}
\end{table*}

%% file: 5_conclusions.tex
\section{Conclusions and Limitations}
\label{sec:conclusions}

In this work, we studied the new task of CoVR
by proposing a simple yet effective methodology to create
automatic training data. Our results on several benchmarks
(including our manually curated video benchmark, as well
as existing image benchmarks)
suggest that, while noisy, such an automated and scalable approach
can provide effective CoVR model training.
One potential limitation of our method is that
the generated modification text may not depict
some visible changes due to not considering the image pair, but only their captions.
Moreover, our modification text %
is suboptimal
due to only inputting
one-word difference caption pairs
(i.e., focusing only on one change,
and not considering multi-word differences). 
For example, the following modification with multiple changes
from the CIRR dataset would not
exist in our data
``close up of a similar dog, but it is swimming on its own with a tennis ball in its mouth''.
Future work can incorporate
visually-grounded modification generation
and multiple modifications between query and target video pairs.

\section*{Ethics statement}
Our model constitutes a generic multi-modal search tool,
but is not intended for a specific application.
While there are helpful use cases such as
online shopping, traveling, and personal development (i.e., how-to),
there may be potential privacy and harmful risks 
when training the model on %
datasets with inappropriate content.
The risks include %
surveillance applications such as searching for a specific person, %
and looking up violent and graphic videos.
For our WebVid-CoVR data release, we refer to \if\sepappendix1{Section~A} \else{Section~\ref{app:sec:WV-dataset-statistics} }\fi
for further analysis about removal of inappropriate content.
We note that our dataset users must also adhere to the terms of use stipulated by WebVid~\cite{bain21_frozen}.

%% file: 5b_acknowledgements.tex
\section*{Acknowledgements}
This work was granted access to the HPC resources of IDRIS under the 
allocation {2023-AD011014223} made by GENCI.
The authors would like to acknowledge the research gift from Google,
the ANR project CorVis ANR-21-CE23-0003-01, 
Antoine Yang\textquotesingle s Google PhD fellowship, and 
thank Mathis Petrovich, Nicolas Dufour, Charles Raude, and Andrea Blazquez for their helpful feedback.

%% file: 7_appendix.tex
\renewcommand{\thefigure}{A.\arabic{figure}} %
\setcounter{figure}{0} 
\renewcommand{\thetable}{A.\arabic{table}}
\setcounter{table}{0} 
\renewcommand{\thesection}{\Alph{section}} %
\setcounter{section}{0}

\startcontents[sections]
{
	\hypersetup{linkcolor=black}
	\printcontents[sections]{l}{1}{}
}

\vspace{.3cm}

\noindent
This document provides 
\ourWV dataset statistics (Section~\ref{app:sec:WV-dataset-statistics}),
\new{\ourCC dataset statistics (Section~\ref{app:sec:CC-dataset-statistics}),}
implementation details (Section~\ref{app:sec:implementation-details}),
additional experiments (Section~\ref{app:sec:experiments}),
and additional qualitative examples (Section~\ref{app:sec:qualitative-examples}).
We also provide the code and %
dataset %
together with a datasheet,
and an illustrative video %
on our project page at \href{https://imagine.enpc.fr/~ventural/covr}{imagine.enpc.fr/\textasciitilde ventural/covr}.

\section{\ourWV dataset statistics and analysis}
\label{app:sec:WV-dataset-statistics}

In this section, we provide analysis on our \ourWV.
A detailed datasheet can be found as a separate file.\\

\noindent\textbf{Filtering inappropriate content and vulgar language.}
We take several measures to detect semi-automatically any inappropriate
content, and remove such instances from our dataset. To achieve this, we 
use a combination of tools (such as negative sentiment and profanity detectors)
and apply them on modification texts and video captions.

We conduct a sentiment analysis on the modification texts using the \texttt{TextBlob} library~\cite{textblob}
to identify instances of negative sentiment. %
We find that less than 0.5\% of the dataset (about 2k instances) exhibits negative sentiment. 
Upon manual review, we identify false positives in this categorization,
including examples such as ``make it an evil pumpkin'' or
``Change him into a frustrated businessman''.
The instances detected as negative sentiment are reviewed and
260 of them are removed from the dataset. 
We ensure that the dataset does not
include any videos marked for mature content, by checking the metadata of
WebVid~\cite{bain21_frozen} provided by \cite{cleanvid}.
Finally, using the \texttt{better-profanity} library~\cite{better_profanity},
we identify approximately 2k video captions that are marked for profanity.
Upon manual inspection, we find that
there were a large number of videos displaying computer-generated visuals with those words.
We also notice false positives (e.g., misinterpretation due to context),
such as the animal cock being incorrectly identified as profanity.
The videos detected to contain profanity in their captions
are reviewed and excluded from the dataset.\\ %

\begin{figure}
  \centering
  \includegraphics[width=0.48\textwidth]{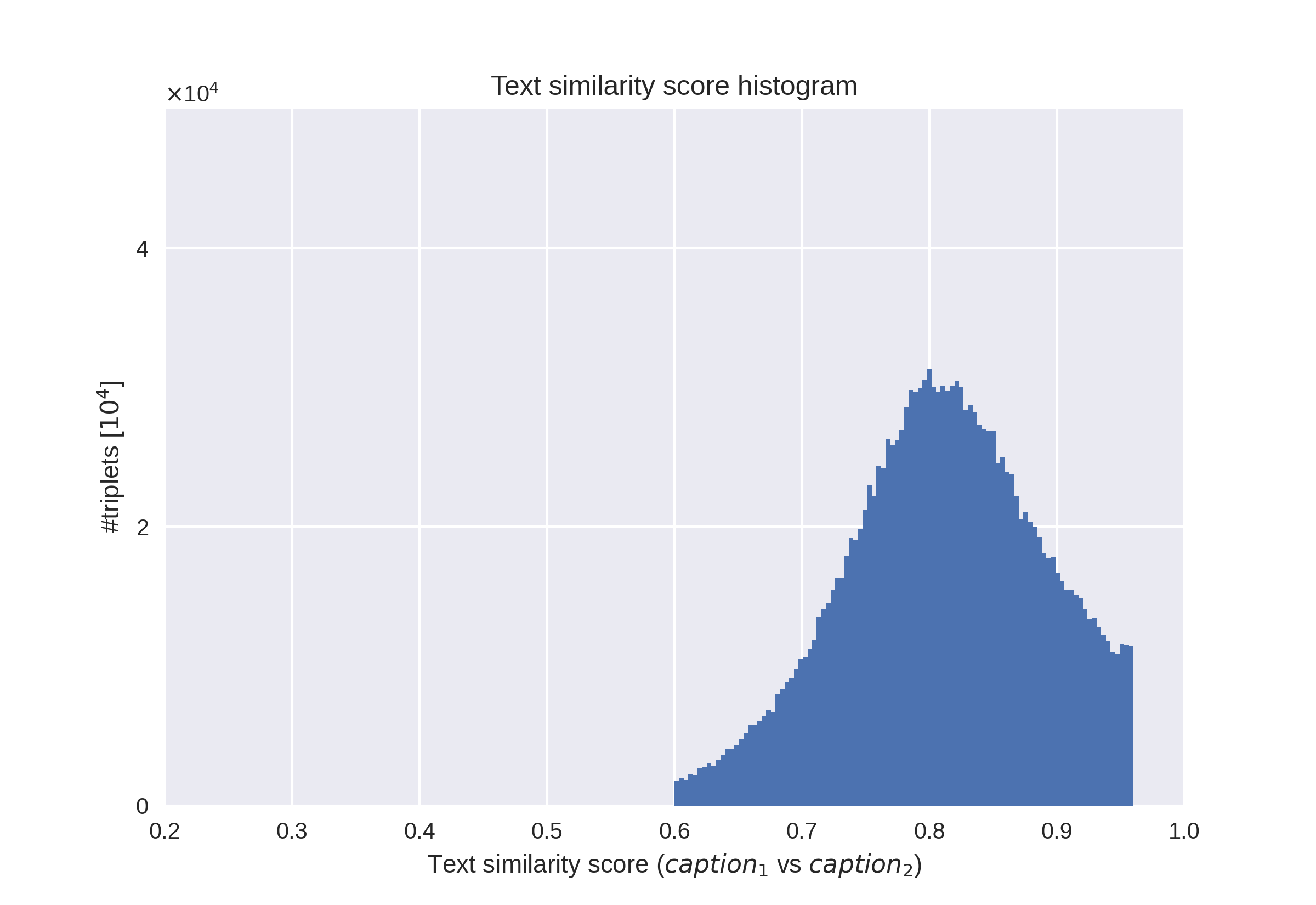}\hfill
  \includegraphics[width=0.48\textwidth]{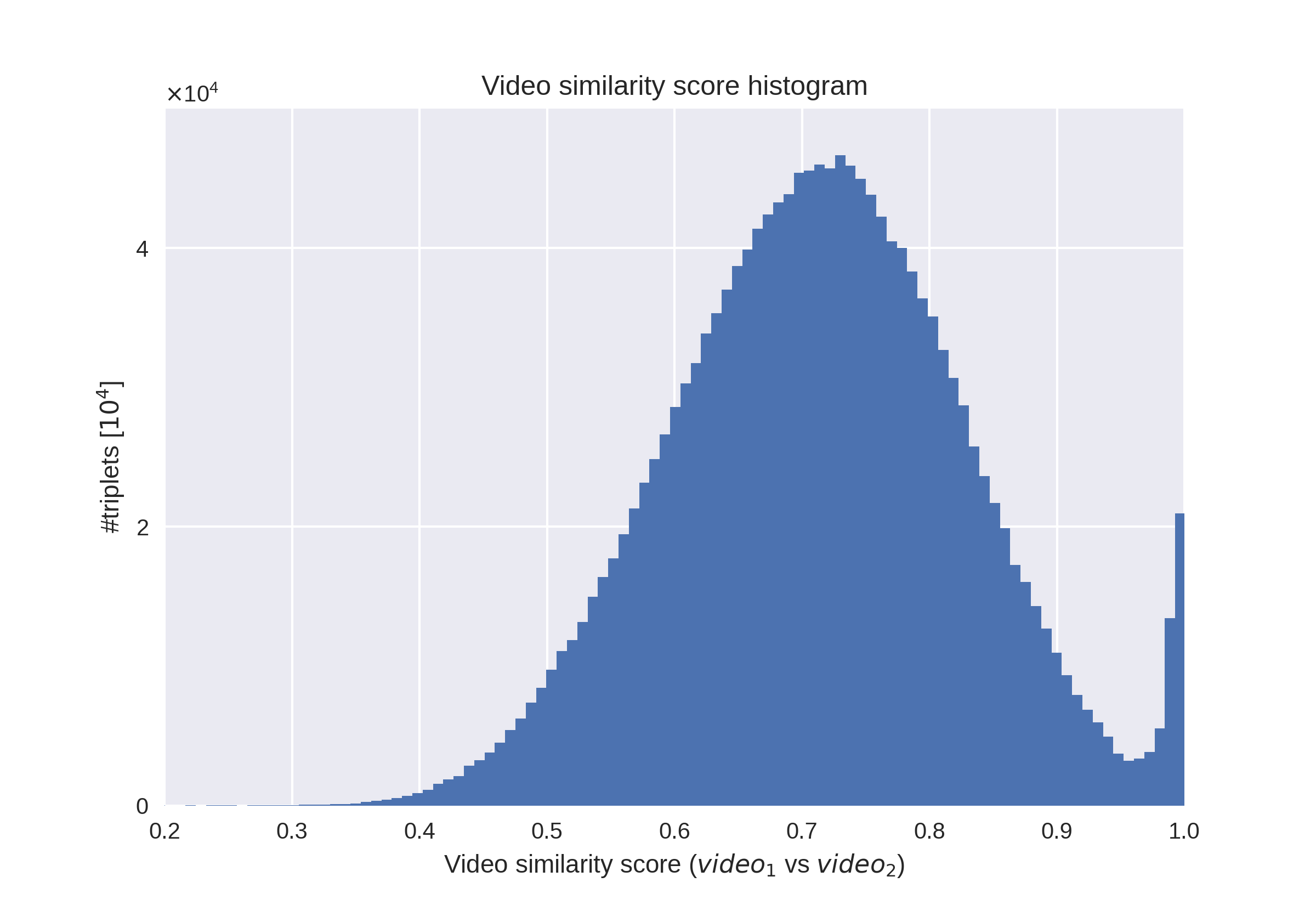}
    \caption{\textbf{Text/video similarity of the caption/video pairs:} 
    Distribution of text similarity scores between caption pairs $(caption_1, caption_2)$ (left) 
    and video similarity scores between video pairs $(video_1, video_2)$ (right),
    using CLIP embeddings and cosine similarity.}
  \label{app:fig:plot-similarity}
\end{figure}

\noindent\textbf{Distribution of caption and video embedding similarities.}
As explained in 
\if\sepappendix1{Section~3.1} \else{Section~\ref{subsec:gen} }\fi
of the main paper,
we filter caption pairs with CLIP text embedding similarity $\ge$ 0.96 
and caption pairs with CLIP text embedding similarity $\le$ 0.6, and for each caption pair, we choose the 10 video pairs with the highest CLIP visual similarity computed at the middle frame of the videos.
We also note that our cosine similarities are normalized between [0, 1].
Here, we further show the distribution of text embedding similarity in caption pairs and visual embedding similarity in video pairs in Figure~\ref{app:fig:plot-similarity}.
The distribution of video similarity scores exhibits two distinct peaks. 
The first peak corresponds to a score of approximately 0.7 and includes video pairs that are significantly dissimilar.
The second peak corresponds to a score close to 1.0 and represents video pairs 
with highly similar visual content.\\

\begin{figure}
  \centering
  \includegraphics[width=.99\linewidth]{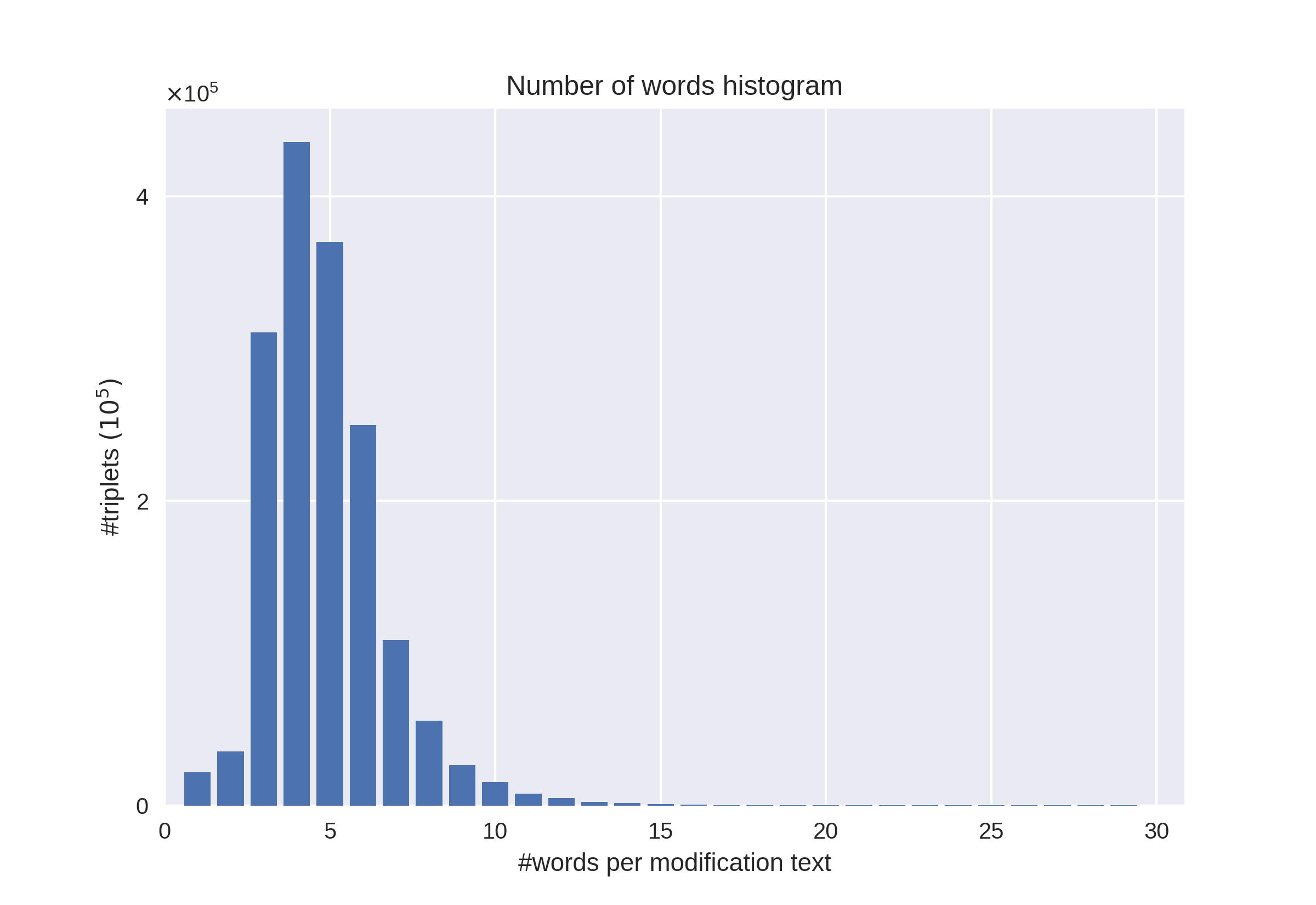}
  \caption{\textbf{Histogram of the number of words in the generated modification text:} Most modification texts have between 3 and 8 words.
  }
  \label{app:fig:nwords-histogram}
\end{figure}

\noindent\textbf{Number of words in modification texts.}
Figure~\ref{app:fig:nwords-histogram} further provides the histogram of the
number of words in the generated modification text. We observe that the majority
of texts contain 3-8 words. \\

\begin{figure}
  \centering
  \includegraphics[width=.99\linewidth]{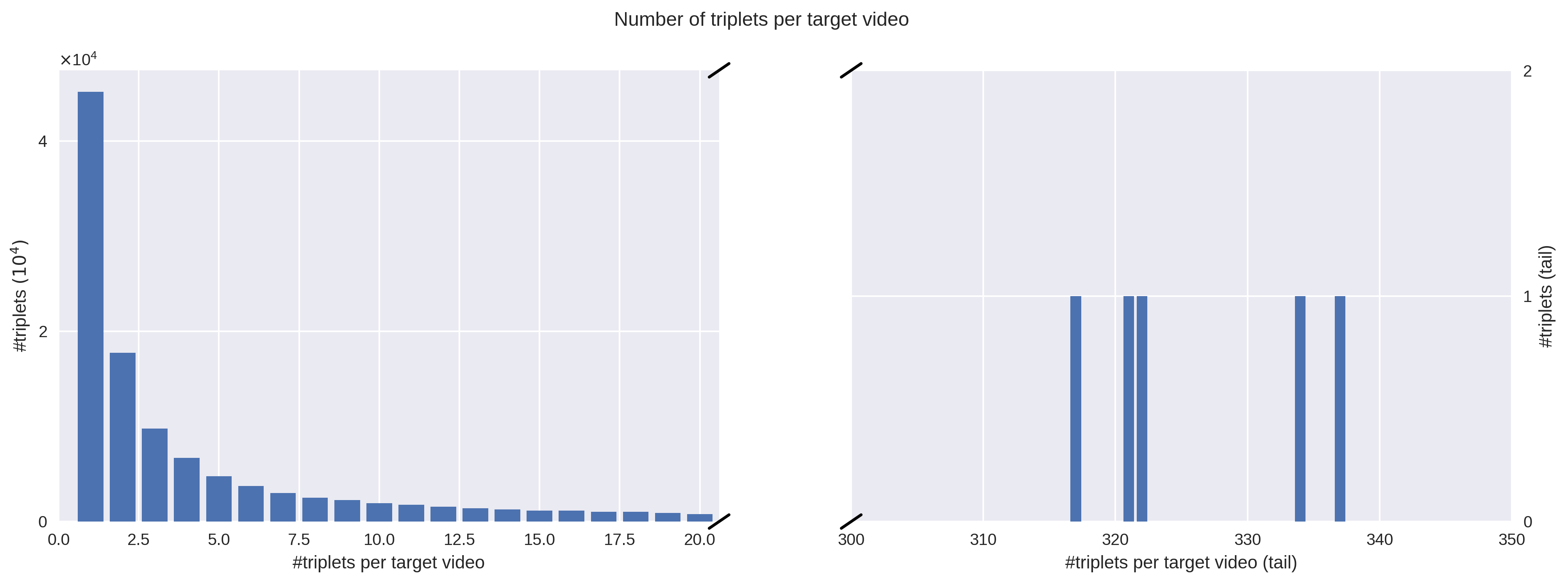}
    \caption{\textbf{Distribution of number of triplets per target video:} 
    We display the histogram depicting the number of triplets associated with each target video in the WebVid-CoVR dataset. 
    Most target videos have 1 or 2 triplets and certain videos exhibit 
    a high number of triplets (zoomed in to the tail on the right plot), %
    e.g., some target videos are present in over 300 triplets, highlighting the variability in modification texts.
    }
  \label{app:fig:ntriplets-pth2}
\end{figure}

\noindent\textbf{Number of triplets per target video.}
In \if\sepappendix1{Section~3.2} \else{Section~\ref{subsec:data} }\fi
of the main paper,
we provided several statistics about our WebVid-CoVR dataset, e.g., 
on average, a target video is associated with 12.7 triplets.
However, in Figure~\ref{app:fig:ntriplets-pth2}, when visualizing 
the distribution of triplets associated with each target video, we see that
the histogram reveals that the majority of target videos are associated to only 1 or 2 triplets. 
The histogram exhibits a long tail, i.e., a small subset of target videos have a considerably larger number of triplets associated. 
These videos have captions such as ``Mountain landscape'', ``Water stream'', and ``Water river'', leading to numerous one-word difference captions associated with them. \\
  
\begin{figure}%
  \centering
  \includegraphics[width=.99\linewidth]{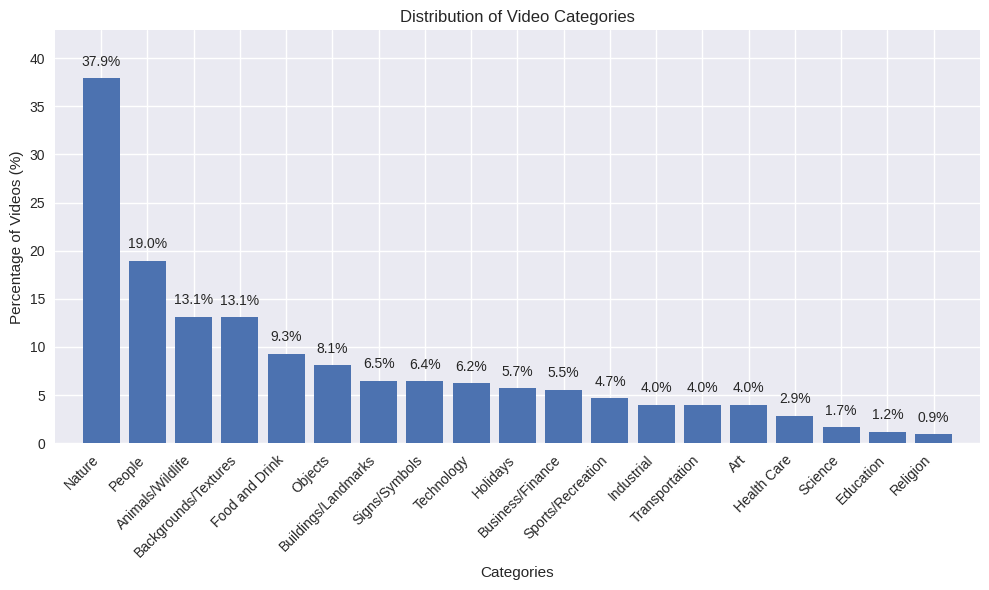}
  \vspace{-0.2cm}
  \caption{\textbf{Distribution of video categories:}  
  {We plot the distribution of categories for videos in WebVid-CoVR, as provided by \cite{cleanvid} as WebVid metadata. Note that 50\% of our WebVid-CoVR videos are present in this metadata collection.
  Looking at the distribution, we observe that around 40\% and 20\% of \ourWV are videos of Nature and People, respectively.}
  }
  \label{app:fig:video-categories}
\end{figure}

\noindent\textbf{Video categories.}
We plot the distribution of video categories in Figure~\ref{app:fig:video-categories}. These categories are found using the WebVid metadata provided by \cite{cleanvid}.
We find 50\% of WebVid-CoVR videos in this metadata collection. Note more than one category can be associated with a single video (e.g., Nature and Animals/Wildlife for a video of a fish in the ocean). \\

\begin{figure} %
  \centering
  \vspace{0.1cm}
  \includegraphics[width=.99\linewidth]{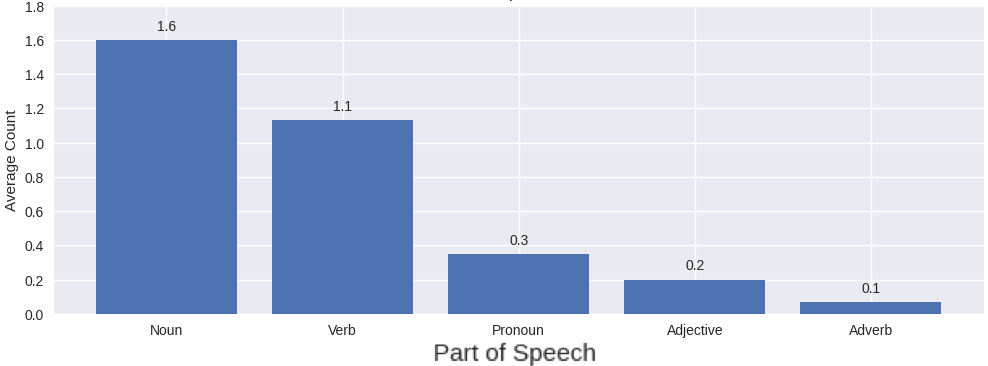}
  \vspace{-0.2cm}
  \caption{\textbf{Distribution of parts of speech in modification texts:}  
  Number of nouns, verbs, pronouns, adjectives, and adverbs in the modification text using part-of-speech (POS) tagging.
  On average, there are more than one noun and one verb per modification text.
  }
  \label{app:fig:pos-distribution}
\end{figure}

\begin{figure} %
  \centering
  \includegraphics[width=.60\linewidth]{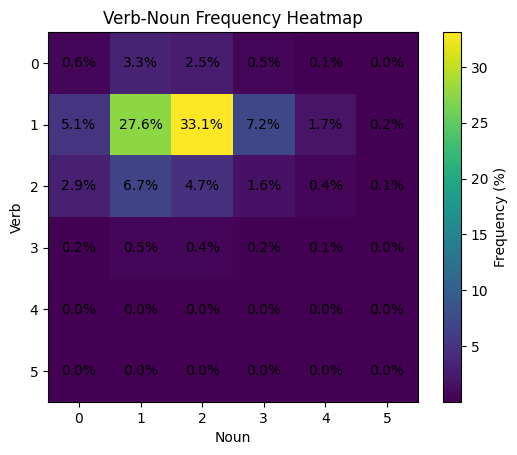}
  \vspace{-0.2cm}
  \caption{\textbf{Verb-noun heatmap:} 
  {This heatmap illustrates the percentage of modification texts containing specific combinations of verbs and nouns. 
    Each cell represents the frequency of a particular verb-noun combination, and the values are presented as percentages. The color intensity indicates the relative frequency of occurrence.
    We observe that over 60\% of the sentences exhibit a pattern of having one verb paired with one or two nouns.}
  }
  \label{app:fig:verb-noun-heatmap}
\end{figure}

\noindent\textbf{Distribution of part-of-speech (POS) tags.}
We conducted POS tagging on the modification texts within the \ourWV dataset to analyze their distribution. 
The resulting analysis reveals the average counts of different parts of speech per modification text, including Nouns, Verbs, Pronouns, Adjectives, and Adverbs. 
We plot the distribution in Figure~\ref{app:fig:pos-distribution},
and see that, on average, a modification text contains 1.6 nouns and 1.1 verbs, 
emphasizing the prevalent use of nouns and verbs in the dataset's modifications. 
The most frequently encountered words within each category's top 3 are as follows:
    Noun: \textit{symbol, water, forest.}
    Verb: \textit{make, turn, change.}
    Pronoun: \textit{it, them, her.}
    Adjective: \textit{green, more, black.}
    Adverb: \textit{instead, more, then.}
We also include a visualization of the verb-noun frequency heatmap in Figure~\ref{app:fig:verb-noun-heatmap}, which provides insights into the distribution of verb-noun count combinations across modification texts in our dataset. From the heatmap, we observe that over 60\% of the sentences exhibit a pattern of having one verb paired with one or two nouns.

\begin{figure} %
  \centering
  \includegraphics[width=.60\linewidth]{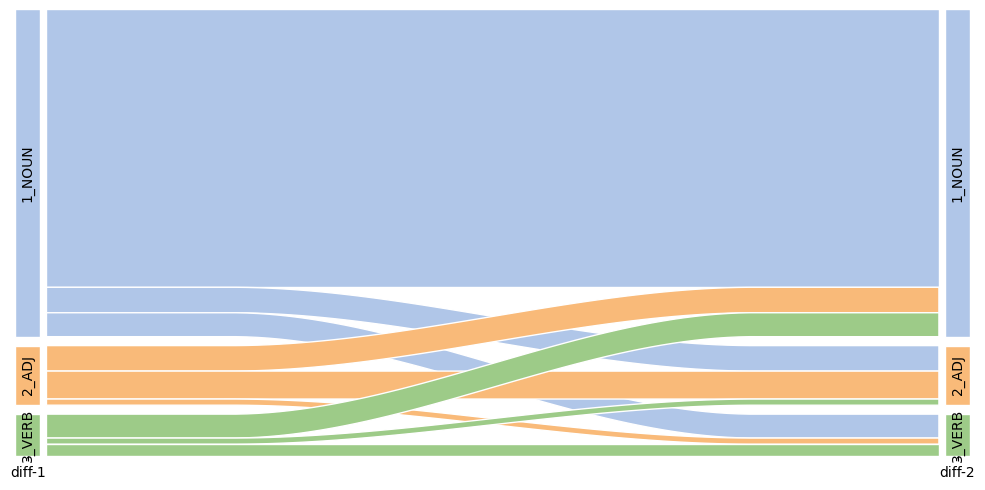}
  \vspace{-0.2cm}
  \caption{\textbf{Transition of POS tags across the difference words between the two captions:} 
  {The visualization primarily focuses on nouns, adjectives, and verbs, which constitute a significant proportion of modifications at 87\% (comprising 65\% nouns, 13\% adjectives, and 9\% verbs). The remaining words fall into categories where the POS tagger was unable to classify the word (12\%) or adverbs (\textless 1\%).
}
  }
  \label{app:fig:pos-transitions}
\end{figure}

We also conducted an analysis using POS tagging on the video {\em captions}.
Figure~\ref{app:fig:pos-transitions} visually illustrates the transition of POS tags across
the difference words in Caption 1 and Caption 2. We observe a predominant pattern of noun-to-noun
changes in our caption pairs. \\

\noindent\textbf{Source of noise.}
As mentioned in \if\sepappendix1{Section~3.2} \else{Section~\ref{subsec:data} }\fi
of the main paper,
about 22\% of the automatic collection can be considered as noisy, 
because this was the percentage of discarded triplets when manually curating the \ourWV test set. 
We expect a similar noise ratio in the training set. To inspect the noise in detail, we manually went over the triplet examples that were marked as unsuitable (therefore discarded) 
when annotating the test set. 
We marked whether the reason for discarding falls within any of the following categories, 
and computed the following percentages (normalized by the number of discarded triplets).
\begin{itemize}
    \item 35\%: The generated modification text does not describe the visual difference. Primarily attributed to either the quality of the video captions or the output generated by the MTG-LLM.
    \item 28\%: Paired videos are visually too similar.
    \item 15\%: Paired videos are visually too different.
    \item 13\%: At least one of the videos is difficult to understand/low quality.
    \item 9\%: Captions are too similar (e.g., one-word difference does not change the meaning: ``On the chairlift'' and ``Ride the chairlift'').
\end{itemize}
While the first category of errors is the largest, it is important to also note that our strict standards for the test set necessitated the discarding of many triplets that could potentially be useful for training. \\

\noindent\textbf{\newt{\ourWV dataset overlap with zero-shot CoIR evaluation datasets.}}
\newt{To contextualize the zero-shot performance, 
we analyze the potential overlap between our \ourWV dataset and the three CoIR datasets. 
We compute the CLIP~\cite{clip2021} embeddings for all target videos in the \ourCC training set and calculate their similarity with the \textit{target} images in the test sets of each CoIR dataset.
We then evaluate the overlap by setting similarity thresholds at 0.7, 0.8, and 0.9.
We define overlap as occurring when at least one sample in the test set has a similarity score above the threshold.
We note that for simplicity, we only consider target images.}

\newt{
The results, summarized in Table~\ref{tab:overlap_cc-webvid}, 
show that no target videos from the \ourWV dataset exhibit a similarity score higher than 0.9 with any of the CoIR datasets, indicating zero overlap at the highest threshold. 
However, as the threshold decreases, the overlap increases, 
specially for CIRCO, which shows a 12.6\% overlap at a 0.8 threshold.}

\input{tables/supmat/overlap_webvic-covr}

\section{\new{\ourCC dataset statistics}}
\label{app:sec:CC-dataset-statistics}
\new{
In this section, we provide analysis on our \ourCC.
We start with 3.3M caption-image pairs, with 2M distinct captions.
As explained in \if\sepappendix1{Section~3.1} \else{Section~\ref{subsec:gen} }\fi
of the main paper, we mine paired images by searching for captions that differ by a single word, excluding punctuation marks.
This process allows us to identify a vast pool of 1.2M distinct caption pairs with 281k distinct captions, resulting in 3.3M triplets after filtering. 
}

\begin{figure}%
  \centering
  \includegraphics[width=.99\linewidth]{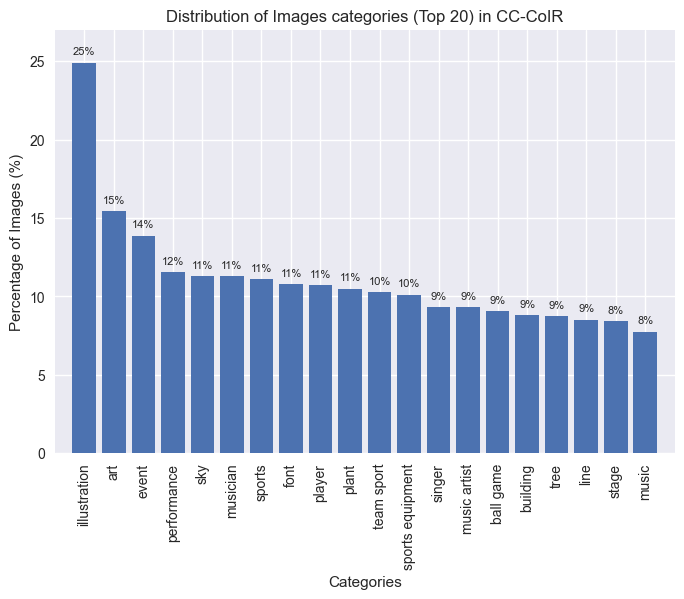}
  \vspace{-0.2cm}
  \caption{\new{\textbf{\new{Distribution of image categories (Top 20) in \ourCC:}} 
  \new{We plot the distribution of categories for images in \ourCC, as provided by \cite{sharma2018CC3M}. Note that 67\% of our \ourCC images have one or more categories in this metadata collection. 
  Looking at the distribution, we observe that around 25\% and 15\% of \ourWV are videos of illustration and art, respectively.}
  }}
  \label{app:fig:cc-image-categories}
\end{figure}

\noindent\textbf{\ourCC image categories.} \new{
Figure~\ref{app:fig:cc-image-categories} illustrates the distribution of the top 20 categories derived from our \ourCC dataset. 
We find 67\% of images within our \ourCC dataset possess one or more associated categories. 
It's worth mentioning that a single video may be associated with multiple categories simultaneously. For instance, a video featuring a singer may be categorized under both "Singer" and "Music Artist". 
It is important to note that while only the top 20 are displayed, the complete dataset encompasses over 10,000 distinct categories. This highlights the wide variety of visual content present within our collection.
}

\begin{figure} %
  \centering
  \includegraphics[width=.99\linewidth]{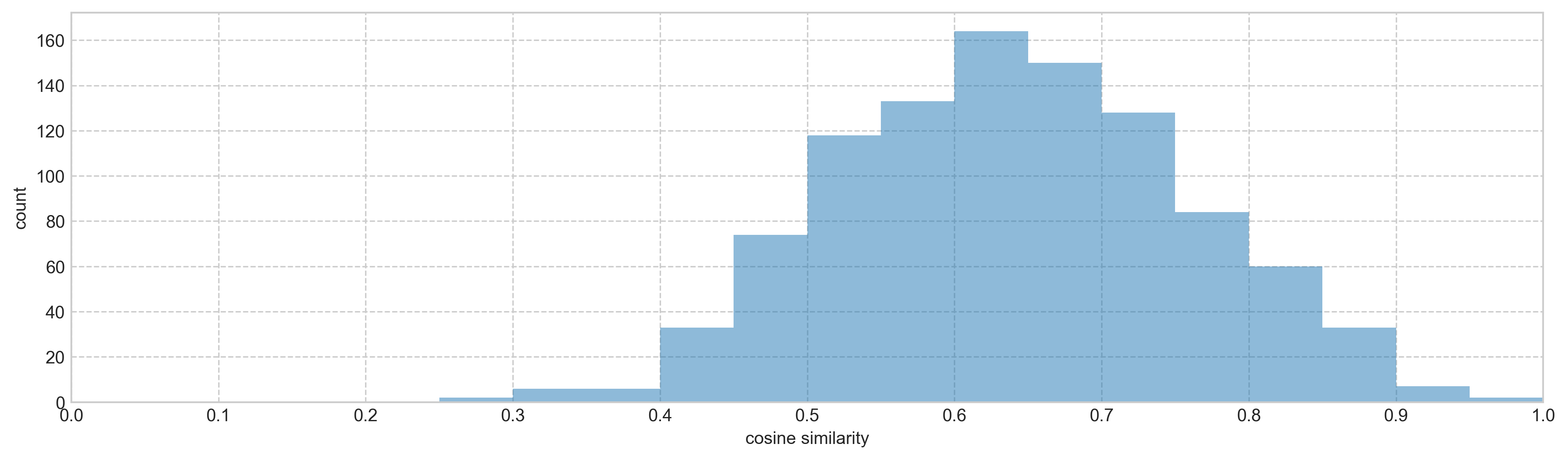}
  \vspace{-0.2cm}
  \caption{\textbf{\newt{Cosine similarity histogram between target image and generated image:}}
  \newt{%
      We plot
  the distribution of cosine similarity scores between target images and images generated using InstructPix2Pix~\cite{brooks2022instructpix2pix} 
  based on the corresponding source image and modification text.
  The results suggest that the majority of target images 
  exhibit a %
  similarity to the generated images.}
  }
  \label{app:fig:noise_cc-coir}
\end{figure}

\noindent\textbf{\newt{Quantifying noise in \ourCC.}}
\newt{
Here, we attempt to quantify how well the modification text describes
the transformation of the query image into the target image.
While this task is challenging, 
one possible approach involves leveraging the InstructPix2Pix~\cite{brooks2022instructpix2pix} model.
We input the source image and corresponding modification text into the model to generate an image that reflects the intended transformation.
The results, summarized in Figure~\ref{app:fig:noise_cc-coir},
reveal that the majority of target images have a cosine similarity ``close'' to that of the generated images. 
This suggests that the modification text generally aligns with the visual changes captured in the target images. 
However, a qualitative examination of the generated images indicates that they cannot always serve as ground truth, suggesting that these results should be interpreted with caution.
}

\noindent\textbf{\newt{\ourCC dataset overlap with zero-shot CoIR evaluation datasets.}}
\newt{
As previously done for \ourWV, we now analyze the overlap between our \ourCC dataset and the CoIR datasets using CLIP~\cite{clip2021} embeddings.
The results, summarized in Table~\ref{tab:overlap_cc-coir}, 
reveal that fewer than 2\% of the target images in the \ourCC training set 
have a similarity score higher than 0.9 in any of the CoIR datasets.
When the threshold is relaxed to 0.8, the overlap increases, with CIRCO showing 20.7\% and CIRR showing 5.2\% of samples with a similarity score above the threshold.}

\input{tables/supmat/overlap_cc-coir}

\section{Implementation details}
\label{app:sec:implementation-details}
We describe the dataset generation computation time (Section~\ref{app:subsec:dataset-computation-time}),
further training details (Section~\ref{app:subsec:training}),
provide the templates we use for our rule-based baseline (Section~\ref{app:subsec:rule-based}),
details about our MTG-LLM finetuning and inference (Section~\ref{app:subsec:mtg-llm}),
and prompt to our prompting experiment (Section~\ref{app:subsec:prompting-llm}.

\subsection{Dataset generation computation time}
\label{app:subsec:dataset-computation-time}
We outline the detailed computation time for each step of the dataset generation. The computation times below are obtained using a \textbf{single} NVIDIA RTX A6000, but it is important to note that most of the processes can be parallelized, which would significantly reduce the wallclock time required. In practice, we used 2 GPUs. 
\begin{itemize}
    \item \textbf{Text embedding extraction:} We extracted text embeddings from 2 million distinct captions out of a total of 2.4 million video-caption pairs. This process completed in less than 2 hours.
    \item \textbf{Caption similarity search:} To identify captions with one-word differences, we employed the \textit{faiss} library~\cite{faiss_johnson2019billion} to select the 100 closest captions, avoiding the need to compare each caption against the entire set of 2 million captions. This optimization significantly reduced the search time, resulting in 2.5 hours.
    \item \textbf{Text similarity filtering:} Thanks to the precomputed text embeddings, the text similarity filtering step incurred no additional time overhead. All the text filtering processes were completed in less than 5 minutes, even on a large pool of 1.2 million captions.
    \item \textbf{Video similarity computation:} To filter by video similarity, we extracted the middle frame from approximately 135,000 videos and computed CLIP embeddings. This step takes approximately 3 hours.
    \item \textbf{MTG-LLM model finetuning:} Finetuning for 715 examples takes less than 10 minutes. Note that the time required to finetune the MTG-LLM model is independent of the number of CoVR triplets we generate.
    \item \textbf{Modification text generation:} This is the most time-consuming stage of the pipeline. It takes around 24 hours to process the 1.6 million caption pairs.
\end{itemize}

\subsection{Training details}
\label{app:subsec:training}
Here, we provide implementation details
in addition to \if\sepappendix1{Section~4.1} \else{Section~\ref{subsec:setup} }\fi
of the main paper. 
In terms of the optimization algorithm, 
we utilize AdamW~\cite{loshchilov2017decoupled}.
For our MTG-LLM, we finetune for one epoch with a batch size of 128 and a learning of $3\mathrm{e}{-5}$ that is warmed up linearly for the first $100$ steps and then kept constant.
For our CoVR model, keeping the visual backbone frozen largely improves the efficiency of the training process:
an epoch on the CIRR dataset takes 4 minutes with a frozen backbone and 25 minutes with a finetuned backbone, while leading to similar performance.
During the training process, 
we employ several image data augmentations. 
These transformations include a random resized crop, 
where the input image is resized to a resolution of $384 \times 384$. 
Additionally, we apply a random horizontal flip and random adjustments to contrast, brightness, sharpness, translation, and rotation.
We use a weight decay of 0.05 and an initial learning rate of $1\mathrm{e}{-5}$ that is decayed to 0 following a cosine schedule over 10 epochs.

\subsection{List of rule-based templates}
\label{app:subsec:rule-based}

In the ablation studies \if\sepappendix1{(Section~4.6} \else{(Section~\ref{subsec:ablations} }\fi
of the main paper),
we introduced a rule-based MTG baseline.
Here, in Table~\ref{tab:rule-based-templates}, we show the templates used for the rules.
We refer to Section~\ref{app:subsec:qualitative_mtg} (Table~\ref{tab:rule-based-comparision})
for qualitative comparison with our finetuned MTG-LLM.

\input{tables/supmat/rule-based-templates}

\subsection{Generating a modification text from paired captions with MTG-LLM}
\label{app:subsec:mtg-llm}
As described in \if\sepappendix1{Section~3.1} \else{Section~\ref{subsec:gen} }\fi
of the main paper,
we use top-k sampling at inference for the MTG-LLM. 
Specifically, we use $k=200$ and $temperature=0.8$.
We further give details about the text input-output format for the MTG-LLM.
At training, we form the input prompt by concatenating captions and target and adding delimiters and stop sequences similar to InstructPix2Pix~\cite{brooks2022instructpix2pix}.
In detail, given a caption pair $(caption_1, caption_2)$ and a corresponding target $Target$, we concatenate them and add a separator in the following way: $caption_1\mathtt{\{separator\}}caption_2\texttt{\textbackslash n\&\&\textbackslash n}Target$,
where $\mathtt{separator}$ is \texttt{\textbackslash n\&\&\textbackslash n}.

For instance, the model takes as input:
\begin{lstlisting}[breaklines, backgroundcolor = \color{backcolour}]
    Clouds in the sky\n&&\nAirplane in the sky \n\n### Response:
\end{lstlisting}
and is trained to generate the response:
\begin{lstlisting}[breaklines, backgroundcolor = \color{backcolour}]
    Clouds in the sky\n&&\nAirplane in the sky \n\n### Response: Add an airplane
\end{lstlisting}
At inference, we simply leave the response empty, and let the model autoregressively generate a modification text.

As mentioned in \if\sepappendix1{Section~3.1} \else{Section~\ref{subsec:gen} }\fi
of the main paper, we add 15 manually prepared text triplets to the existing 
700 text triplets from \cite{brooks2022instructpix2pix} used for training.
The motivation is to address specific
CoVR cases not present in the original set of triplets, such as \textit{``remove clouds and reveal only sky''} 
given input captions \textit{``Clouds timelapse''} and \textit{``Sky timelapse''}.
We show these 15 samples in Table~\ref{tab:added-examplges-mtg-llm}.

\input{tables/supmat/added-examles-mtg-llm}

\new{
The caption pairs from Table~\ref{tab:added-examplges-mtg-llm}
originate from failure cases of the initial iteration of our MTG-LLM on \ourWV. 
We manually corrected some of the failures by typing modification texts without looking at the corresponding videos, 
i.e.,  in a way we would want a text-only model to behave, by focusing on the changed words. 
We note that some cases are ambiguous, especially when the captions are not long or precise enough. For example, `walking swan' and `white swan' may not necessarily result in `change color to white',
but there is no way to know without looking at the visual pair, 
which itself is an interesting area for future work.}

\subsection{\new{Details of the LLaMA prompt}}
\label{app:subsec:prompting-llm}
In the ablation studies \if\sepappendix1{(Section~4.6} \else{(Section~\ref{subsec:ablations} }\fi
of the main paper), 
we justified why we finetuned LLaMA as opposed to simply prompting it
without any training. 
Here, we show how we determine the prompt for the aforementioned experiment.
Specifically, we prepend few-shot examples of pairs of captions and desired generated texts, before 
adding the two captions in question. 
In particular, we use the following sentence:
\begin{lstlisting}[breaklines, backgroundcolor = \color{backcolour}]
    Clouds in the sky&&Airplane in the sky-> Add an airplane\n
    Aerial view of forest&&Aerial view autumn forest-> Change season to autumn\n
    Clouds timelapse&&Sky timelapse-> remove clouds and reveal only sky\n
    Aerial view of a sailboat anchored in the mediterranean sea.&&Aerial view of two sailboat anchored in the mediterranean sea.-> Add one sailboat\n
\end{lstlisting}
Then, we concatenate our two captions for which we wish to generate a modification text.
The previous results in \if\sepappendix1{(Table~8} \else{(Table~\ref{tab:mtg-llm} }\fi
of the main paper, 
are also consistent with our qualitative observations: we found that the LLM struggles to perform the modification text generation without finetuning (see Table~\ref{tab:rule-based-comparision} in the next section).

\section{Additional experiments}
\label{app:sec:experiments}
We provide additional experiments, reporting CoVR 
results when changing the visual query from an image to a video (Section~\ref{app:subsec:video-query}),
\newt{effect of backbones}
(Section~\ref{app:subsec:pretrained-blip-models}),
incorporating visual similarity between videos (Section~\ref{app:sec:visual-similarity}),
results when filtering dynamic or static videos (Section~\ref{app:sec:dynamic-vs-static-content})
effect of the modification text length (Section~\ref{app:sec:modification-text-length}),
and optimal number of frames (Section~\ref{app:sec:optimal-number-of-frames}).

\subsection{Video query for CoVR}
\label{app:subsec:video-query}
As noted in \if\sepappendix1{Section~3} \else{Section~\ref{sec:method} }\fi
of the main paper,
we focus on image queries in this paper. This was because querying with an image
has arguably more applications for realistic search scenarios.
Here, we explore the setup of using a \textit{video} as the visual query instead of an image query.
We can do this since our dataset consists of video-text-video triplets.
To encode a query video, we sample 15 equally-spaced frames and compute visual
embeddings for each frame using the BLIP-2 image encoder.
We then average the per-frame
embeddings and forward it through the BLIP-2 cross-attention layers to obtain a multimodal query embedding $f(q,t)$.
Note that we keep the target video representation fixed to 15 frames with weighted embedding averaging
as described in \if\sepappendix1{Section~3.3} \else{Section~\ref{subsec:training} }\fi
of the main paper.
As seen in Table~\ref{tab:video-query}, using 15 query frames leads to
similar performance to using the middle frame.

\input{tables/supmat/video-query}

\subsection{Effect of backbones}
\label{app:subsec:pretrained-blip-models}

\noindent\textbf{Variants of pretrained BLIP-2 models.}
All experiments in this paper
are performed with the BLIP-2 model~\cite{li2023blip2} finetuned on COCO~\cite{coco}. 
Here, we include experiments when changing this backbone with ViT-L without COCO finetuning (BLIP-2 base).
For this experiment
(as in the last row of \if\sepappendix1{Table~2} \else{Table~\ref{tab:video} }\fi
of the main paper),
we use pretrained cross-attention layers of BLIP-2 as our multimodal combined representation,
and finetune them on \ourWV.
\new{In Table~\ref{tab:pretrained-blip-models}, we observe that the BLIP-2 model fine-tuned with COCO has a similar performance to BLIP-2 Base,
with a slight improvement on R@1.}

\input{tables/supmat/pretrained-blip-models}

\noindent\textbf{\newt{Effect on CIR benchmarks.}}
\newt{
As shown in \if\sepappendix1{Table~2} \else{Table~\ref{tab:video} }\fi 
of the main paper, the BLIP-2 backbone with 
cross-attention layer finetuning achieves the highest performance on the \ourWVt dataset. 
In Table~\ref{tab:clip}, we extend the comparison to additional CIR benchmarks,
evaluating frozen CLIP with average fusion (no training),
CLIP with MLP fusion, 
and BLIP and BLIP-2 with cross-attention layer finetuning. 
The results show that MLP fusion fails to generalize effectively across these datasets when trained with \ourWV,
while the BLIP-2 backbone continues to deliver the best results across all benchmarks.
}

\input{tables/supmat/clip}

\subsection{Incorporating visual similarity between videos}
\label{app:sec:visual-similarity}
When constructing the \ourWV dataset, we rely solely on caption similarity 
to identify similar video pairs for generating triplets, as discussed in \if\sepappendix1{Section~3.1.} \else{Section~\ref{subsec:gen}. }\fi 
Here we provide an additional analysis on the effect of incorporating visual similarity 
in a similar fashion as we did with the caption similarity.

Specifically, we train variants of our model where we filter the \ourWV training triplets 
to only keep those whose video pair similarity is above a certain threshold. 
We measure the similarity using CLIP features on the middle frames, 
as in \if\sepappendix1{Section~3.1.} \else{Section~\ref{subsec:gen}. }\fi 
The results in Table~\ref{tab:visual-similarity} demonstrate that increasing the visual similarity threshold 
consistently decreases the downstream performance, 
while also discarding a large portion of the training data.

\input{tables/supmat/visual-similarity}

This suggests that relying solely on caption similarity is reasonable, 
and that incorporating visual similarity more strictly can be detrimental. 
A potential reason is that enforcing visual similarity constraints could bias the model 
to ignore the input text modification. 
Overall, our results indicate that the automatically mined video pairs with caption similarity already exhibit sufficient visual consistency for training the CoVR task.

\subsection{Dynamic vs static content}
\label{app:sec:dynamic-vs-static-content}
The videos in our \ourWV dataset contain both dynamic sequences with motion, as well as static content.
To analyze the prevalence, we compute the optical flow using the Gunnar Farneback's algorithm~\cite{Gunnar2003OpticalFlow} and 
empirically choose a magnitude threshold of 1 to distinguish between videos with static and dynamic elements. 
The magnitude value is obtained by averaging the Euclidean norms of motion vectors
in both horizontal and vertical directions across the computed video frames. 
We identify that around 25\% of the triplets contain static target videos, 
which represents approximately 21\% of the overall target videos. 

\newt{As an additional method to distinguish between static and dynamic triplets, 
we analyze the part of speech of the word that changes in the target video caption.
Specifically, we examine whether the modified word is a noun, verb, or another part of speech. We find that 65\% of the changes involved nouns, 9\% involve verbs, and the remaining 26\% correspond to other parts of speech (such as adjectives).
}

We train models omitting either only the static portion or only the dynamic portions. 
The results in Table~\ref{tab:static-dynamic} show that training on both dynamic and static triplets is beneficial, 
with a minor decrease in performance if we omit static videos during training (while maintaining the same iteration count). 
This may be because image training data can still be complementary to video training~\cite{bain21_frozen}.

Overall, we demonstrate that \ourWV contains both static and dynamic videos, 
hence posing an advantage over image datasets by providing more diversity. 

\newt{Ideally, leveraging dynamic videos would imply capturing not just static states but also the transitions between them, 
which is a key distinction between composed video retrieval and composed image retrieval.
For instance, when given an image of 'wheat flour' and a text query like 'change this to dough,' 
an image search will likely retrieve the end result, such as 'dough.' 
In contrast, composed video retrieval could potentially
capture the entire process, showing 'how' the transformation from 'flour' to 'dough' occurs, 
providing a richer and more informative result.
We leave this potential application to future research, 
which will benefit from better datasets and improved modeling techniques.
}

\input{tables/supmat/static-dynamic}

\subsection{Effect of modification text length}
\label{app:sec:modification-text-length}
We analyze the impact of the modification text length, 
by experimenting with generating multiple candidates per caption pair and selecting the longest modification text, 
increasing the average length from 23.36 to 33.35 characters.
However, as shown in Table~\ref{tab:text-length}, this decreases performance on 
all three datasets (\ourWVt, CIRR, and FashionIQ). 
We hypothesize that since our triplets represent a single modification, longer texts tend to be more verbose without improving quality.

\input{tables/supmat/text-length}

This shows that while the generated modification texts are not as long on average as the ones from CIRR, 
our generated texts still provide useful training signal,
as evidenced by the state-of-the-art performance of models trained on our dataset %
when transferred to standard CoIR tasks (CIRR and FashionIQ in \if\sepappendix1{Table~4} \else{Table~\ref{tab:sota_cirr+fiq} }\fi
and
CIRCO in \if\sepappendix1{Table~5).} \else{Table~\ref{tab:circo_test}). }\fi 
Moreover, the average number of characters of the modification text
in \ourWV and the widely used FashionIQ CoIR benchmark are comparable 
(see \if\sepappendix1{Table~1} \else{Table~\ref{tab:datasets} }\fi
of the main paper: 23.36 vs 27.13, respectively).

\subsection{\new{Optimal number of frames}}
\label{app:sec:optimal-number-of-frames}
\begin{figure} %
  \centering
  \includegraphics[width=.99\linewidth]{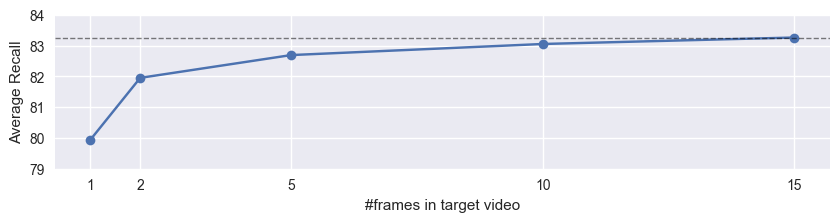}
   
  \caption{\new{\textbf{Training with more target video frames} 
  {consistently improves performance on \ourWVt.}}
  }
  \label{app:fig:n_frames}
\end{figure}

\new{In the main paper, we sample 15 frames from \ourWV videos during training.
Here, we experiment with %
incrementally increasing the number of frames for training and testing.
We report the average recall on \ourWVt and observe
a steady increase in performance with more frames (see Figure~\ref{app:fig:n_frames}).}
\newt{However, it's worth noting that the performance is already quite high with just one frame, 
likely due to the static image bias inherent in the WebVid dataset, where videos are typically short in duration.
}

\section{Qualitative analysis}
\label{app:sec:qualitative-examples}
In this section, we provide
examples of caption filtering (Section~\ref{app:subsec:qualitative_filtering}),
qualitative comparison between different MTG approaches (Section~\ref{app:subsec:qualitative_mtg}),
qualitative examples of our \ourWV triplets (Section~\ref{app:subsec:qualitative_triplets}),
samples from our manual test set annotation process (Section~\ref{app:subsec:manual}),
qualitative CoVR results on \ourWVt (Section~\ref{app:subsec:recall-webvid})
and CoIR results on CIRR (Section~\ref{app:subsec:recall-cirr}).

\subsection{Examples of filtered captions}
\label{app:subsec:qualitative_filtering}
As described in \if\sepappendix1{Section~3.1} \else{Section~\ref{subsec:gen} }\fi
of the main paper,
we employ a filtering process to select paired captions 
that facilitate the generation of meaningful training data.
In this section, we provide examples of the filtered captions.

\noindent\textbf{Filtering template captions.} 
Upon analyzing the paired captions, 
we observed that a significant portion of the pairs originated 
from a small set of template captions. 
Out of 1.2M distinct caption pairs, 
approximately 719k ($60\%$) were generated from these template captions. 
The following examples showcase some of these template captions:

\begin{itemize}
    \item \textbf{Abstract:} \textit{Abstract color movement tunnel, Abstract color nature background, Abstract color smoke flowing on white background, Abstract colorful paint ink spread explode, Abstract colorful pattern background, Abstract colorful red cement wall background or texture. the camera moves up, Abstract colorful satin background animation, Abstract colorful shiny bokeh background., Abstract colorful smoke on black background,} etc
    \item \textbf{Background:} \textit{Abstract background, Animated backgrounds, Animation, background., Aquarium background, Artistic background, Aurora background, Balloons background, Basketballs background, Beach background, Bluebell background, Bright background, Brush background, Bubbles background, Bubbly background, Celebrate background, Celebratory background, Cg background, Christmas background, Christmas background, Circles background, Color background, Colored background, Colorful background, Colorfull background,}, etc.
    \item \textbf{Concept:} \textit{Brazil high resolution default concept, Brazil high resolution dollars concept, Businessman with advertising hologram concept, Businessman with algorithm hologram concept, Businessman with automation hologram concept, Businessman with bitcoin hologram concept, Businessman with branding hologram concept, Businessman with public relations hologram concept, Close up of an eye focusing on a freelance concept on a futuristic screen., Coins fall into piggy bank painted with flag of ghana. national banking system or savings related conceptual 3d animation, Communication concept, Communication network concept., Communication team concept, Concept of connection, Concept of dancing at disco party. having fun with friends., Concept of education, Concept of geography, Cyber monday concept}, etc 
    \item \textbf{Flag:} \textit{Flag of america, Flag of andorra, Flag of aruba, Flag of austria, Flag of azerbaijan, Flag of bahrain, Flag of belarus, Flag of belize, Flag of black, Flag of bolivia, Flag of brazil, Flag of bulgaria, Flag of cameroon, Flag of canada,} etc.
\end{itemize}

\noindent\textbf{Filtering caption pairs with high or low similarity.} 
To ensure the generation of meaningful modifications, 
we further refine the selection of caption pairs by filtering out 
those with excessively high or low similarity. 
Caption pairs with highly similar meanings may result 
in trivial or unnoticeable modifications. 
Conversely, pairs with significant dissimilarity can lead to large visual 
differences that are difficult to describe accurately.
We show below some of the filtered captions based on the CLIP text embedding cosine similarity.
\begin{itemize}
    \item \textbf{High similarity:} $10\%$ of the pairs have CLIP text similarity above 0.96.
    \begin{itemize}
        \item Close-up of a tree with green leaves and \underline{sunlight}
        \item Close-up of a tree with green leaves and \underline{sunshine}
    \end{itemize}
    \begin{itemize}
        \item Businessman \underline{speaking} on the phone
        \item Businessman \underline{talking} on the phone
    \end{itemize}
    \begin{itemize}
        \item Boat on \underline{a} sea
        \item Boat on \underline{the} sea
    \end{itemize}
    \item \textbf{Low similarity:} $2\%$ of the pairs have CLIP text similarity below 0.60.
    \begin{itemize}
        \item \underline{Leaves} close-up
        \item \underline{Peacock}, close-up
    \end{itemize}
    \begin{itemize}
        \item Moon \underline{jellyfish}
        \item Moon \underline{night}
    \end{itemize}
    \begin{itemize}
        \item Close up of a \underline{lynx}
        \item Close up of a \underline{milkshake}
    \end{itemize}
\end{itemize}

\noindent\textbf{Exclusion of digit differences and out-of-vocabulary words.}
In order to maintain the high quality and coherence of the generated modification text, 
we apply additional filtering criteria. 
Specifically, we exclude caption pairs where the differences between captions 
are numerical digits (often representing dates) 
or involve out-of-vocabulary words (using the python libraries wordfreq and enchant) that may hinder the generation process.

\begin{itemize}
    \item \textbf{Difference between the captions is a digit:} Approximately $2\%$ of the pairs. 
    \begin{itemize}
        \item \underline{23.09.2015} navigation on the moscow river
        \item \underline{07.08.2015} navigation on the moscow river.	
    \end{itemize}
    \begin{itemize}
        \item Light leaks element \underline{190}
        \item Light leaks element \underline{215}
    \end{itemize}
    \begin{itemize}
        \item Pure silver, shape of granules of pure silver each one is unique \underline{44} (2)
        \item Pure silver, shape of granules of pure silver each one is unique \underline{95} (2)
    \end{itemize}
   \item \textbf{Difference in one of the captions has an out-of-vocabulary word:} Approximately $7\%$ of the pairs. 
    \begin{itemize}
        \item Businessman writing on hologram desk tech word- bitcoin
        \item Businessman writing on hologram desk tech word- crm
    \end{itemize}
    \begin{itemize}
        \item Mitomycin-c - male doctor with mobile phone opens and touches hologram active ingrident of medicine
        \item Oxazepam - male doctor with mobile phone opens and touches hologram active ingrident of medicine
    \end{itemize}
    \begin{itemize}
        \item Blue forget-me-nots
        \item Blue galaxy
    \end{itemize}
\end{itemize}

\input{tables/supmat/rule-based-comparison}

\subsection{Qualitative comparison of MTG approaches}
\label{app:subsec:qualitative_mtg}
In \if\sepappendix1{Section~4.6} \else{Section~\ref{subsec:ablations} }\fi
of the main paper and Section~\ref{app:subsec:prompting-llm}, we show that finetuning our MTG-LLM works better
than a rule-based approach and than few-shot prompting of the LLM.
In this section, we provide a qualitative comparison of three different methods 
for generating modification text: (i) rule-based, (ii) prompting-based, and (iii) our MTG-LLM finetuning. 
We present examples of paired captions and the corresponding modification texts generated by each method in Table~\ref{tab:rule-based-comparision}.

\textbf{Rule-based method.} 
The rule-based method relies on predefined rules to generate modification text. 
We illustrate an example limitation in the last row of Table~\ref{tab:rule-based-comparision},
where the difference text is simply a preposition (i.e., `of' vs `above'), and the modification
text becomes `Remove of'.
The rule-based method performs well when the modifications follow a specific pattern, 
but it may struggle with more complex modifications (e.g., `tree' vs `trees' should generate `add more trees' for plurality).

\textbf{Prompting LLM.} 
The prompting-based method involves using a pretrained language model without finetuning.
However, this method is prone to hallucinations and may generate modification 
text that does not accurately represent the intended difference. 
For example, in the second example, 
the prompting LLM suggests removing the term `animal' instead of replacing `bird' with `bear'.

\textbf{MTG-LLM (Our approach).} 
Our MTG-LLM approach utilizes a large language model finetuned on a manually annotated dataset 
specifically for modification text generation. 
It tends to be the most robust across different cases.

\subsection{Training triplet examples}
\label{app:subsec:qualitative_triplets}
\begin{figure*}%
  \centering
  \includegraphics[width=.8\linewidth]{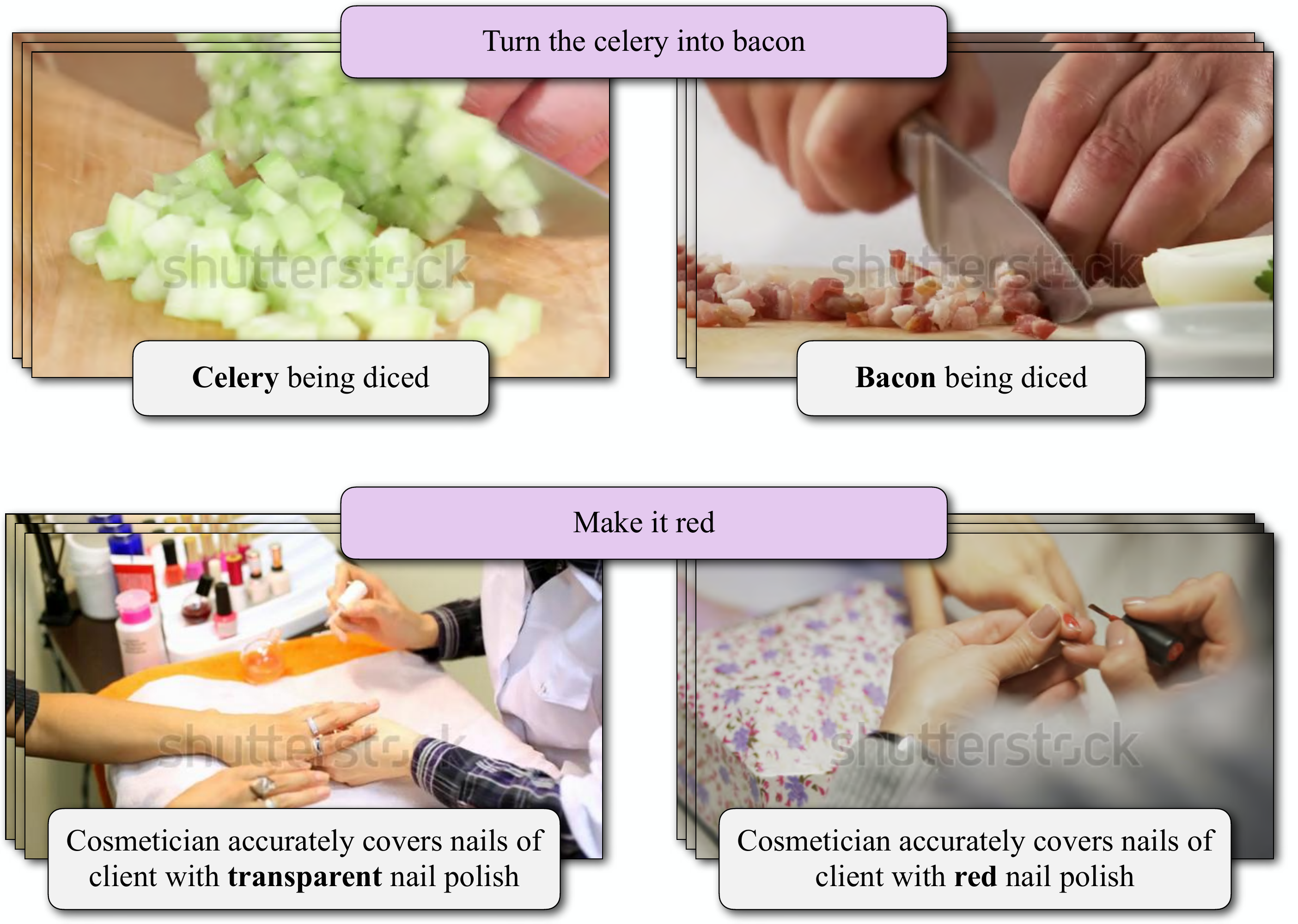}
  \includegraphics[width=.8\linewidth]{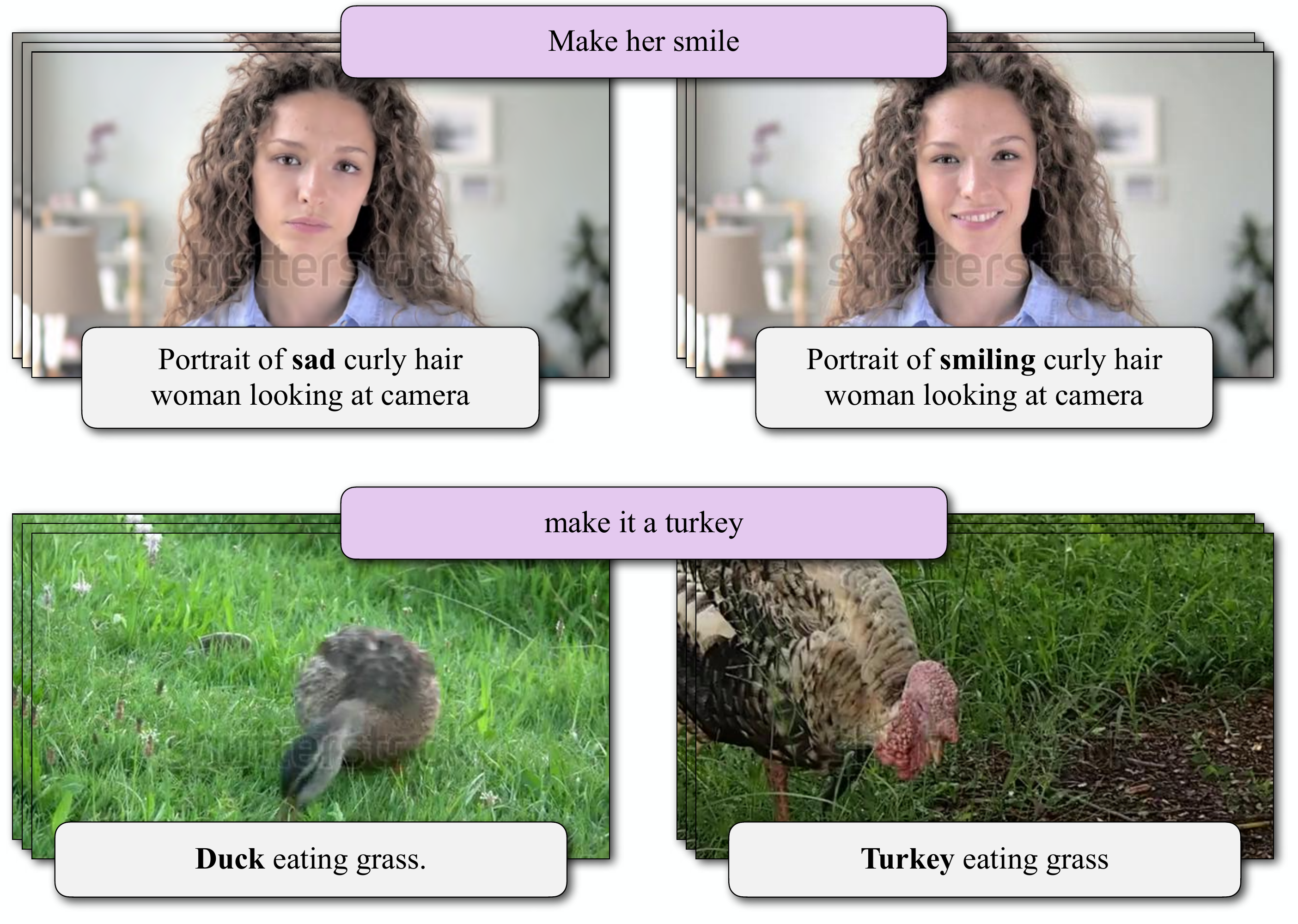}
  \caption{\textbf{Examples of generated triplets:} 
  {We illustrate triplet samples (one per row) generated using our automatic dataset creation methodology. 
  Each sample consists of two videos with their corresponding captions (at the bottom of each video)
  and the generated modification text using our MTG-LLM (in purple).}
  }
  \label{app:fig:triplet-12}
\end{figure*}

\begin{figure*}%
  \centering
  \includegraphics[width=.8\linewidth]{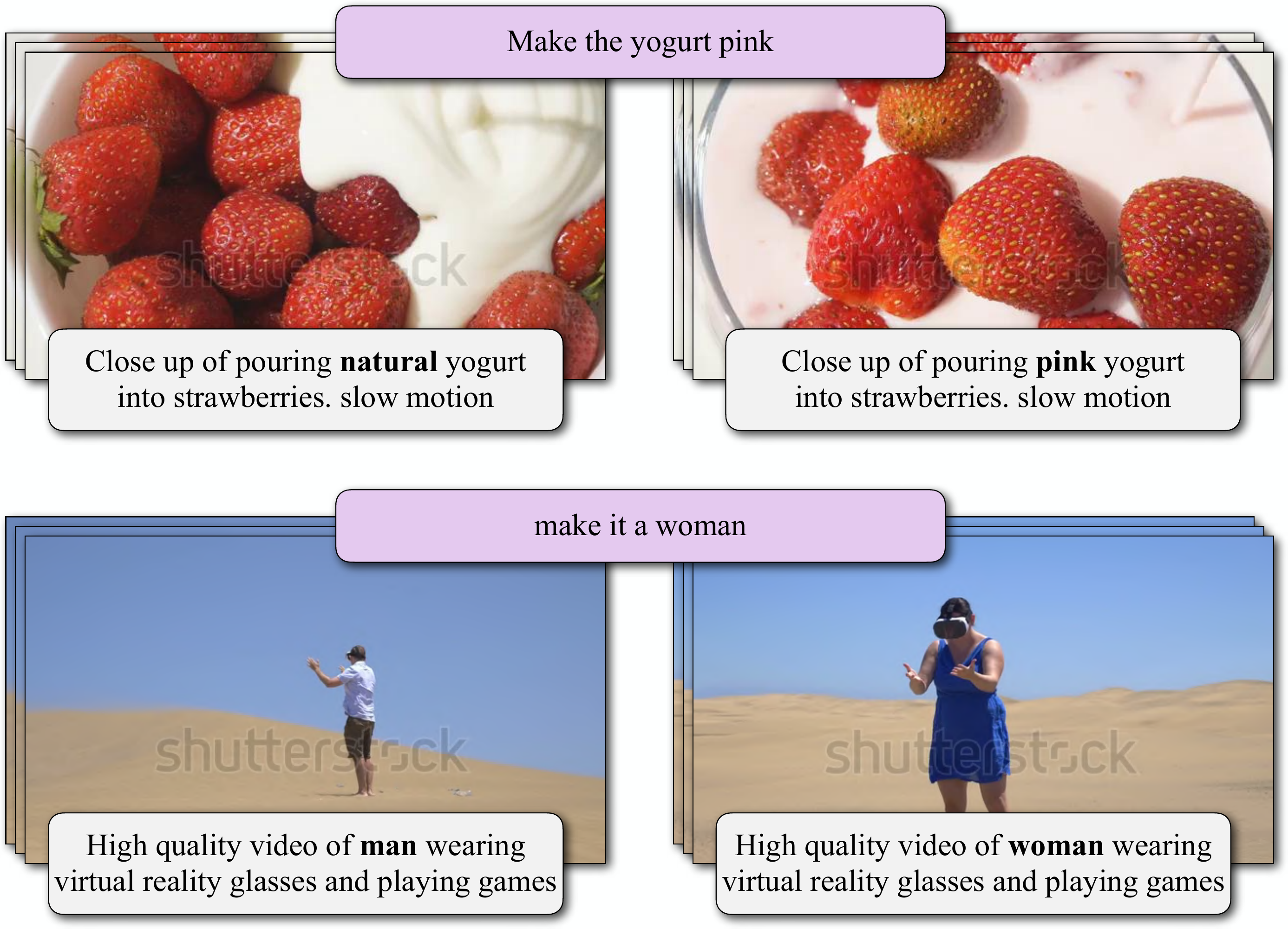}
  \includegraphics[width=.8\linewidth]{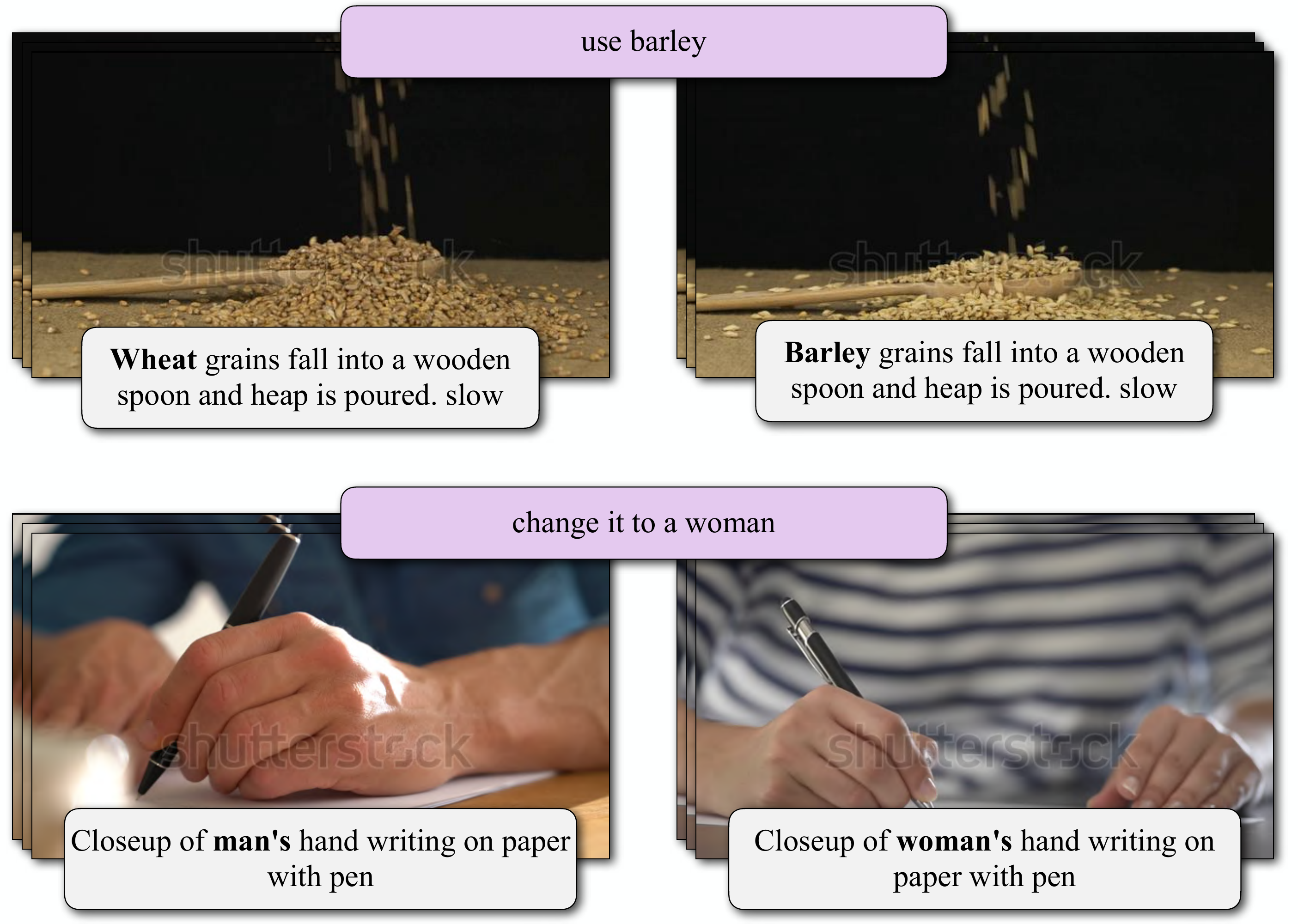}
  \caption{\textbf{Examples of generated triplets (ctd)} 
  }
  \label{app:fig:triplet-34}
\end{figure*}

\begin{figure*}%
  \centering
  \includegraphics[width=.8\linewidth]{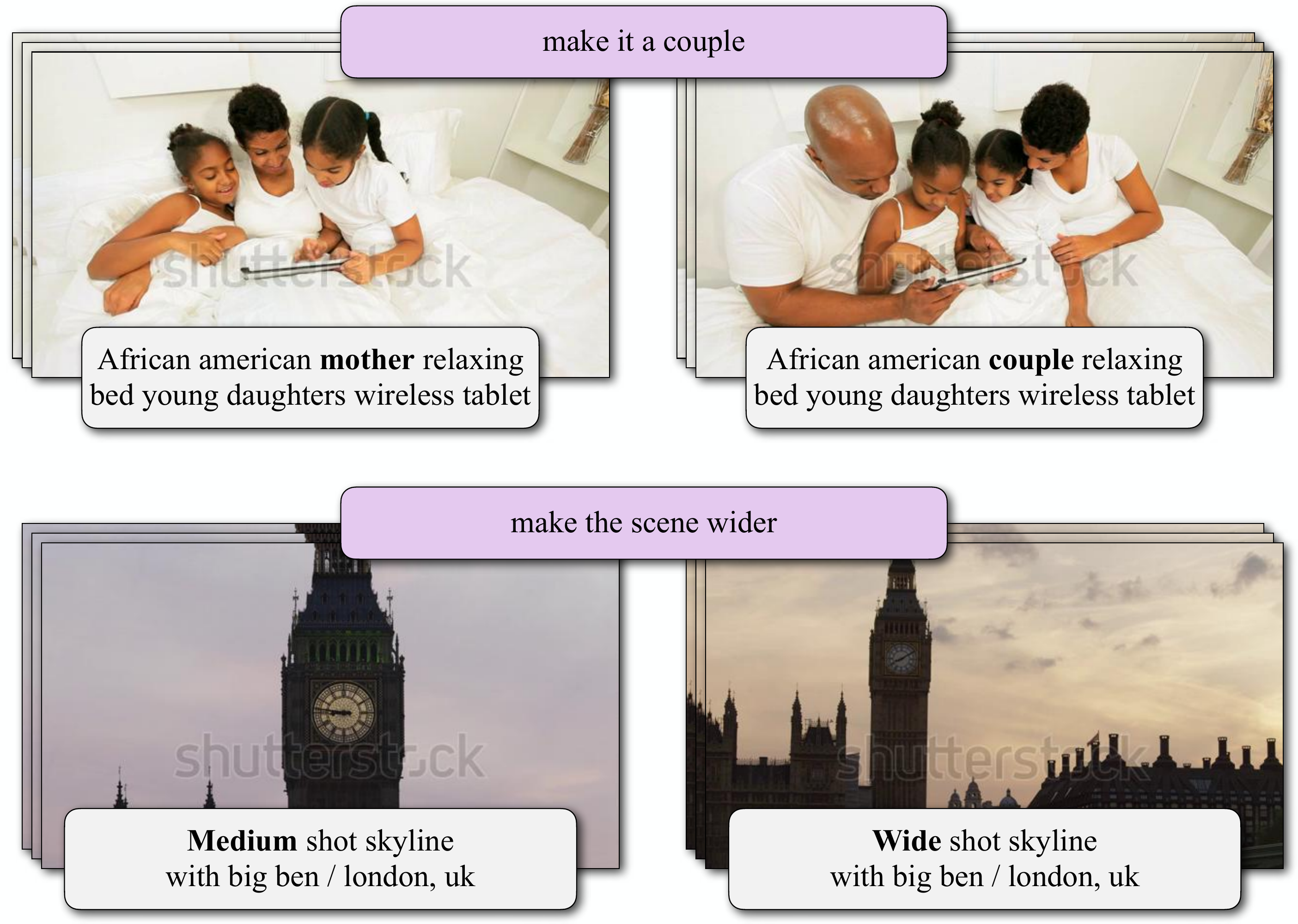}
  \caption{\textbf{Examples of generated triplets (ctd)} 
  }
  \label{app:fig:triplet-5}
\end{figure*}

\begin{figure*}%
  \centering
  \includegraphics[width=.98\linewidth]{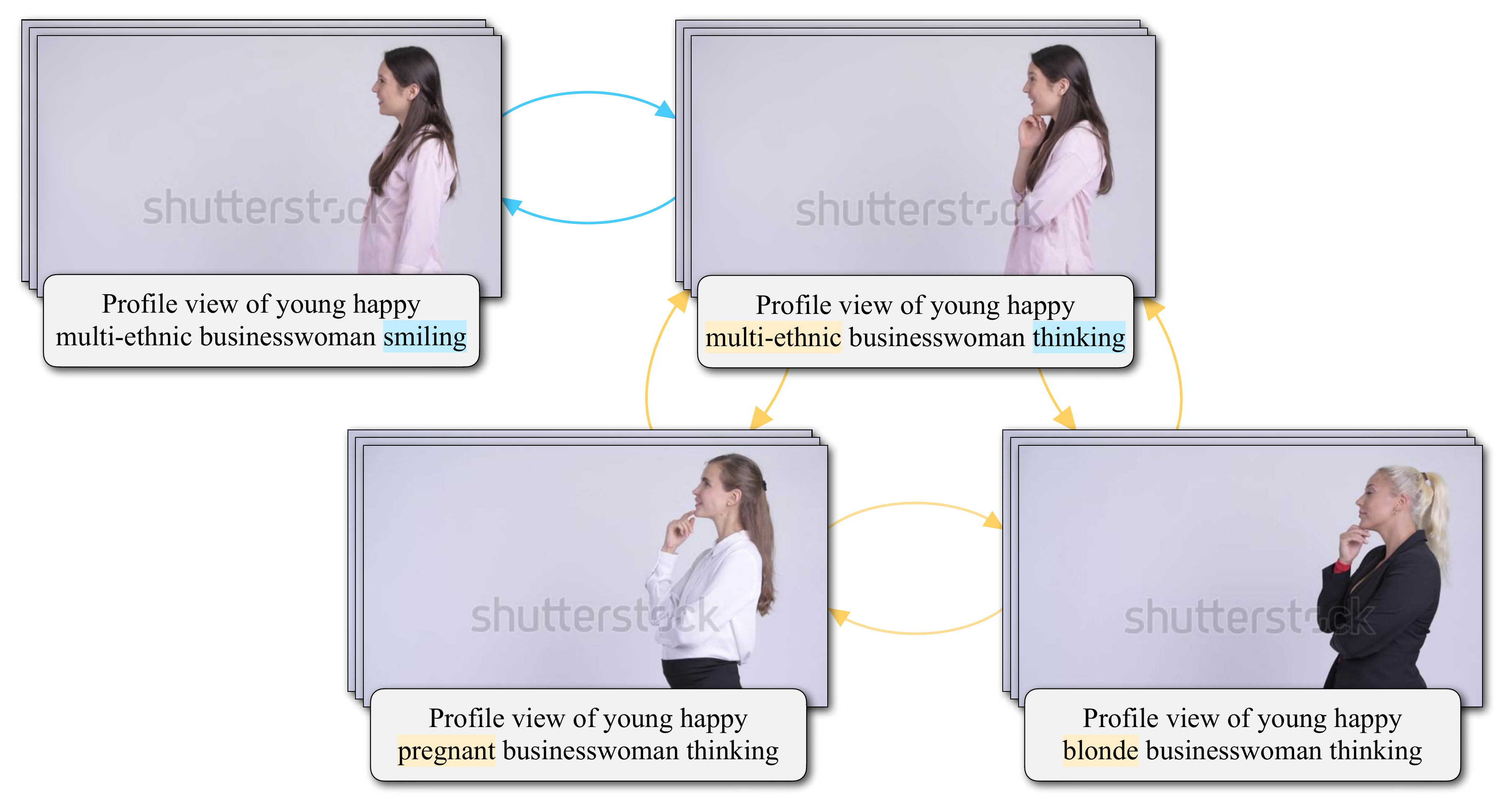}
  \caption{\textbf{Generated triplets from multiple similar captions:} We can train with as many triplets 
  as pairs of captions with one word difference by generating modification texts using our
  trained MTG-LLM:
  \colorbox{bluecolor}{\textit{she is thinking}}, 
  \colorbox{bluecolor}{\textit{Have her look happy}}, 
  \colorbox{yellowcolor}{\textit{Make the businesswoman pregnant}}, 
  \colorbox{yellowcolor}{\textit{make her blonde}}, 
  \colorbox{yellowcolor}{\textit{make her multi-ethnic}}, 
  \colorbox{yellowcolor}{\textit{Make the woman pregnant}}, etc.
  }
  \label{app:fig:triplet-multi}
\end{figure*}

\begin{figure*}%
  \centering
  \includegraphics[width=.98\linewidth]{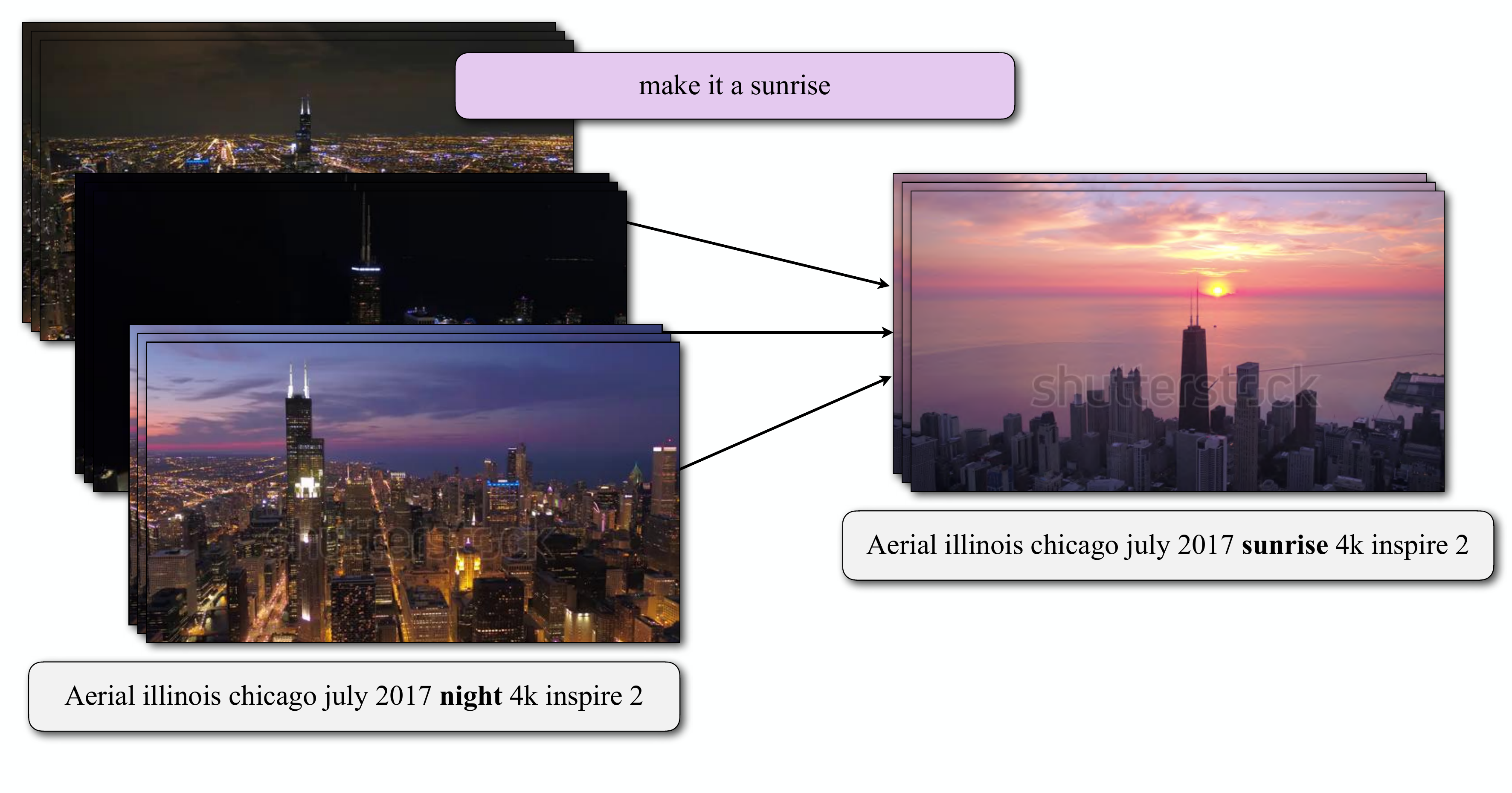}
  \includegraphics[width=.98\linewidth]{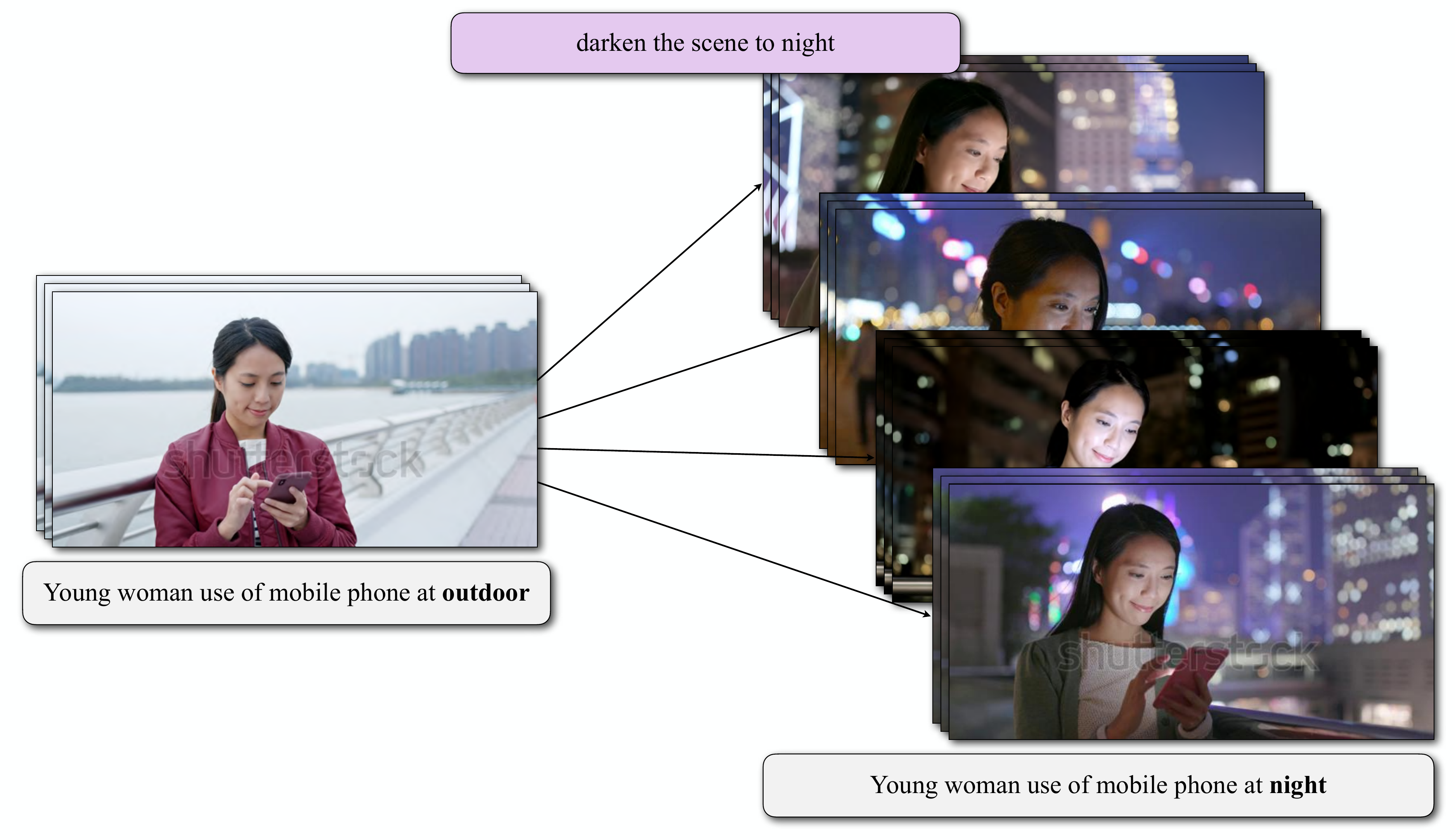}
  \caption{\textbf{Generated triplets with multiple videos:} 
  In cases where there are several videos with the same caption,
  we can generate multiple triplets by leveraging the multiple videos. 
  It can be seen as a way of data augmentation.}
  \label{app:fig:triplet-multi-12}
\end{figure*}

Figures~\ref{app:fig:triplet-12},~\ref{app:fig:triplet-34}, and~\ref{app:fig:triplet-5} 
all show examples of triplets generated using our automatic dataset creation.
These examples demonstrate the effectiveness of our approach 
in generating coherent modification texts for paired videos.
This capability serves as a form of data augmentation and increasing the diversity in the training set.
In Figure~\ref{app:fig:triplet-multi}, we show that the dataset is not composed by pairs only,
as there are many captions that have many relations between them.
Furthermore, in Figure~\ref{app:fig:triplet-multi-12} we show cases where 
a single caption is associated with multiple videos. 
This scenario allows us to generate multiple triplets by leveraging the 
diverse visual content captured in different videos.
The triplets shown in the aforemention figures exhibit a wide range of variations, 
encompassing different themes such as
emotions, food, actions, camera edits, gender changes, and time of the day. 

\subsection{Manual test set annotation}
\label{app:subsec:manual}
In this section, we further describe the process of manually annotating the test set
for our \ourWVt CoVR benchmark, previously discussed in \if\sepappendix1{Section~3.2} \else{Section~\ref{subsec:data} }\fi
of the main paper.
The annotation process involves presenting the annotator 
with generated modification texts from three different runs of MTG-LLM, 
along with three frames each from the query and target videos. The annotator's task is to evaluate the quality of the modification texts and the suitability of the videos for the CoVR task.

\begin{figure*}
    \centering
    {\includegraphics[width=0.38\textwidth]{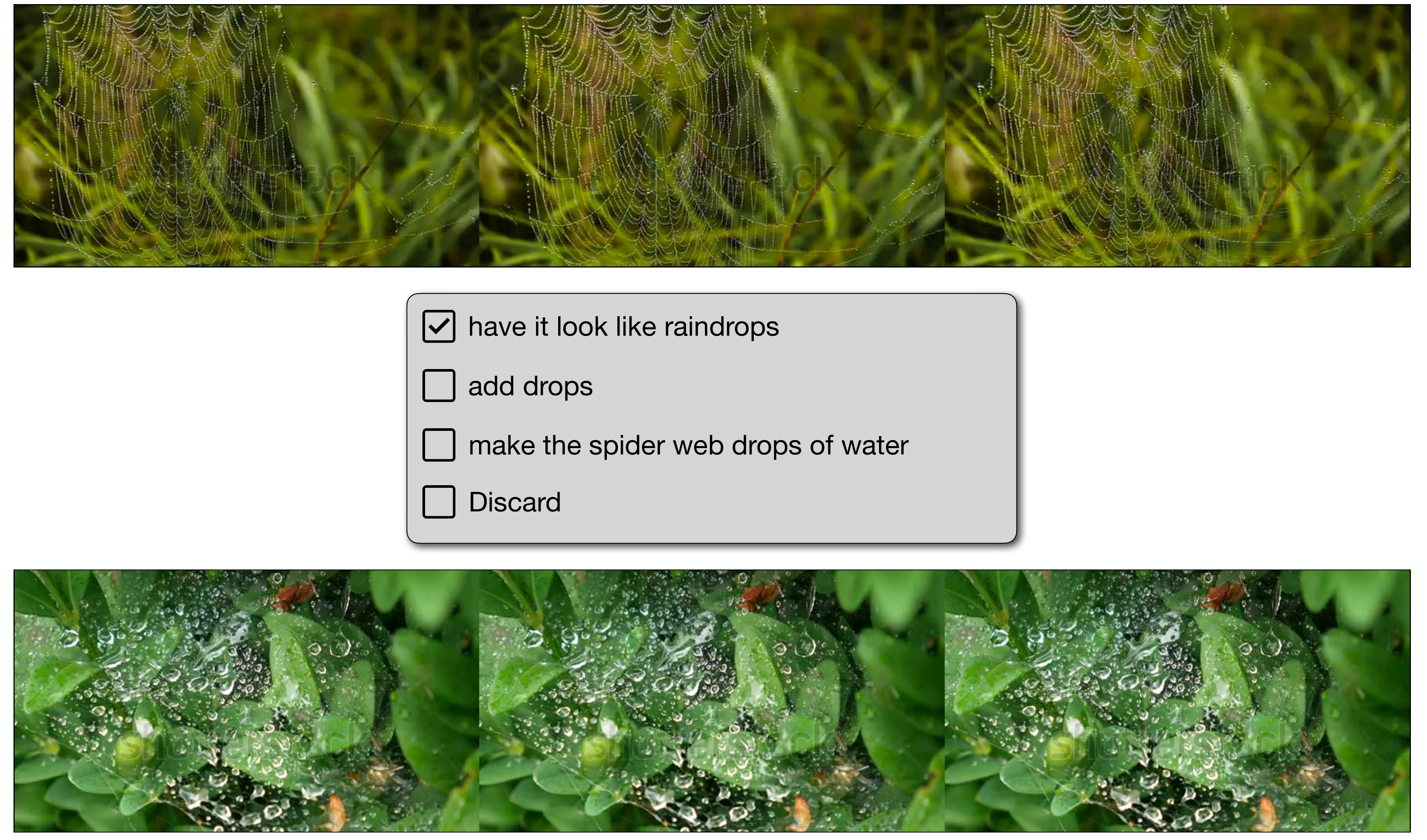}}\hfill 
    {\includegraphics[width=0.38\textwidth]{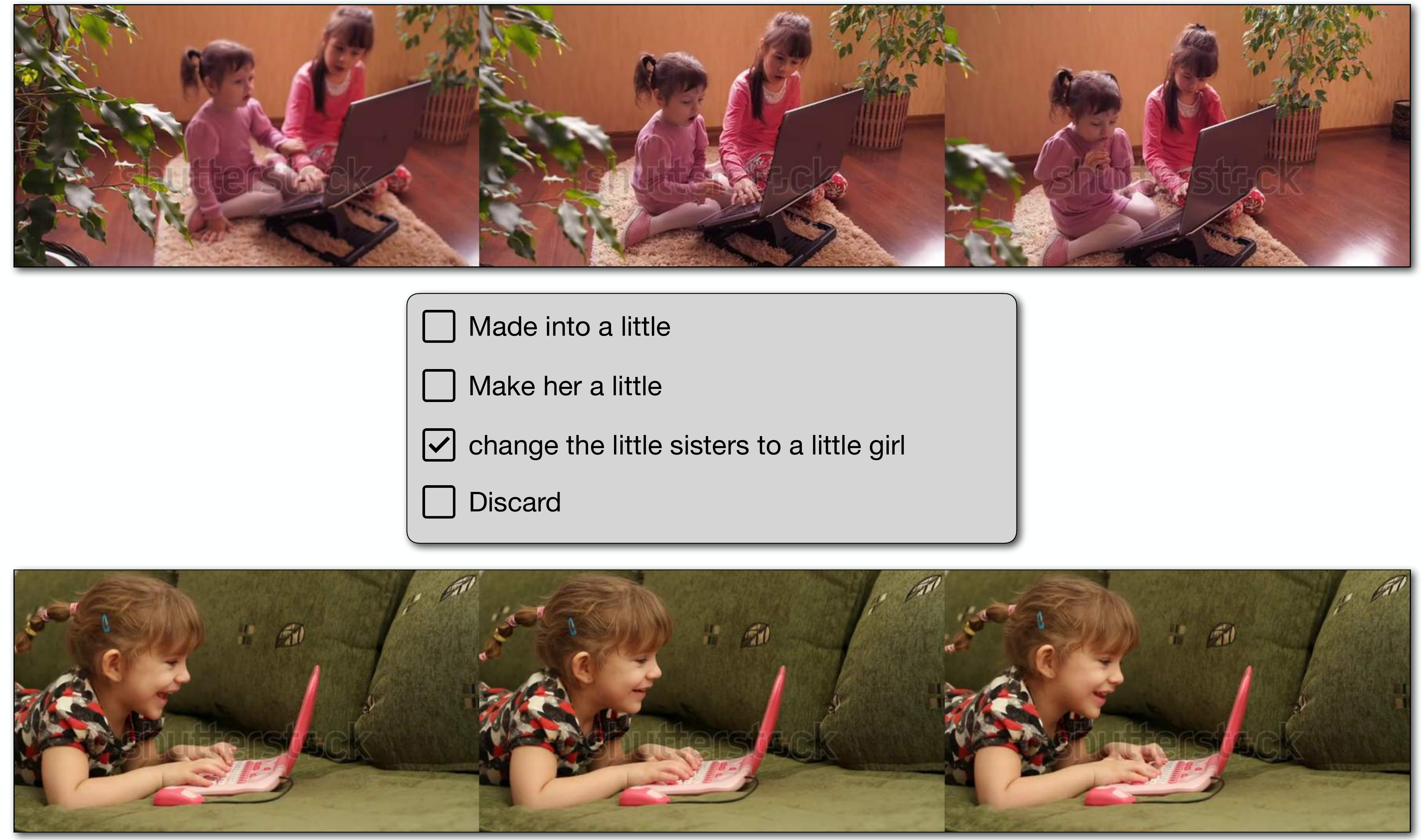}}\\\vspace{0.2cm}
    {\includegraphics[width=0.38\textwidth]{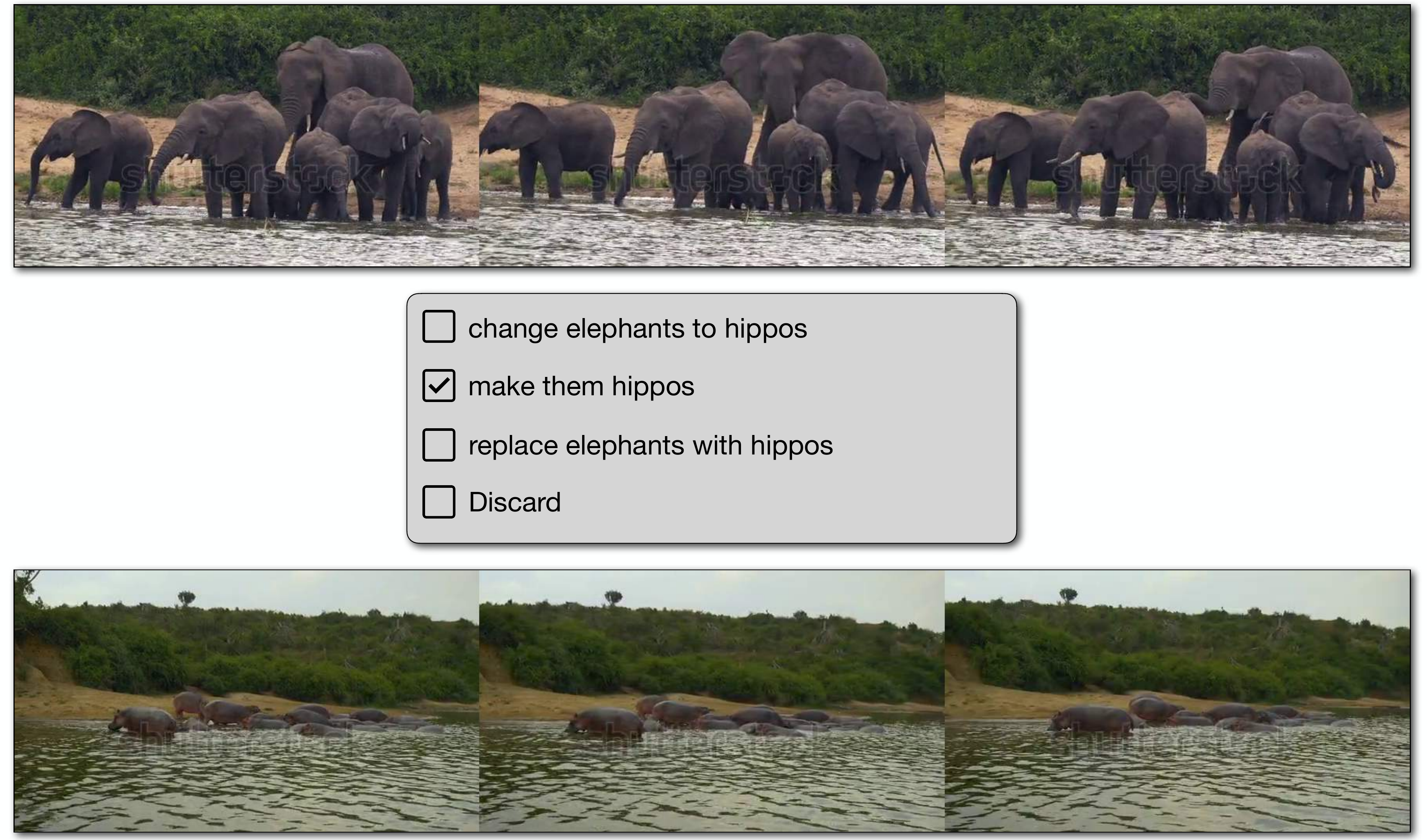}}\hfill
    {\includegraphics[width=0.38\textwidth]{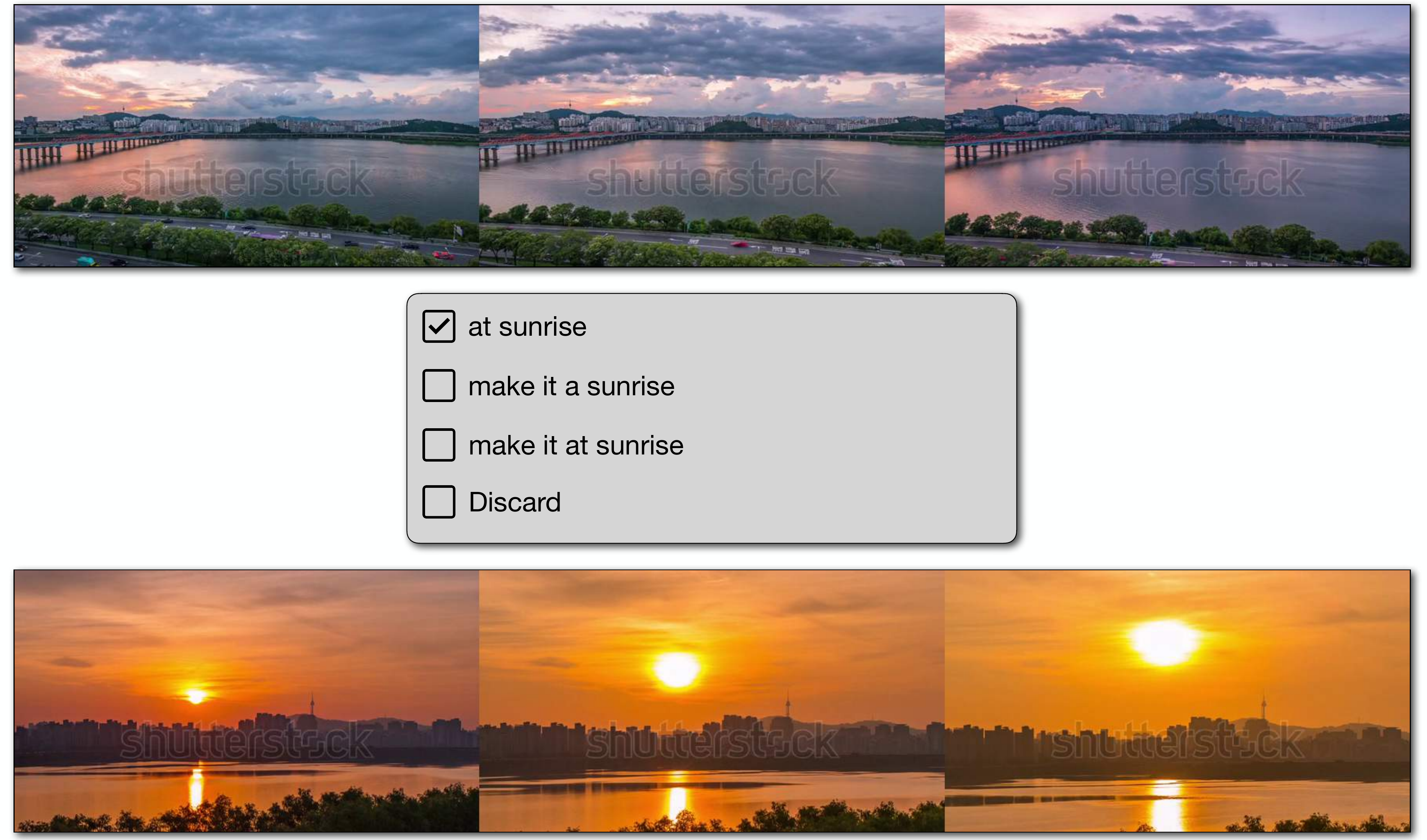}}\\\vspace{0.2cm}
    {\includegraphics[width=0.38\textwidth]{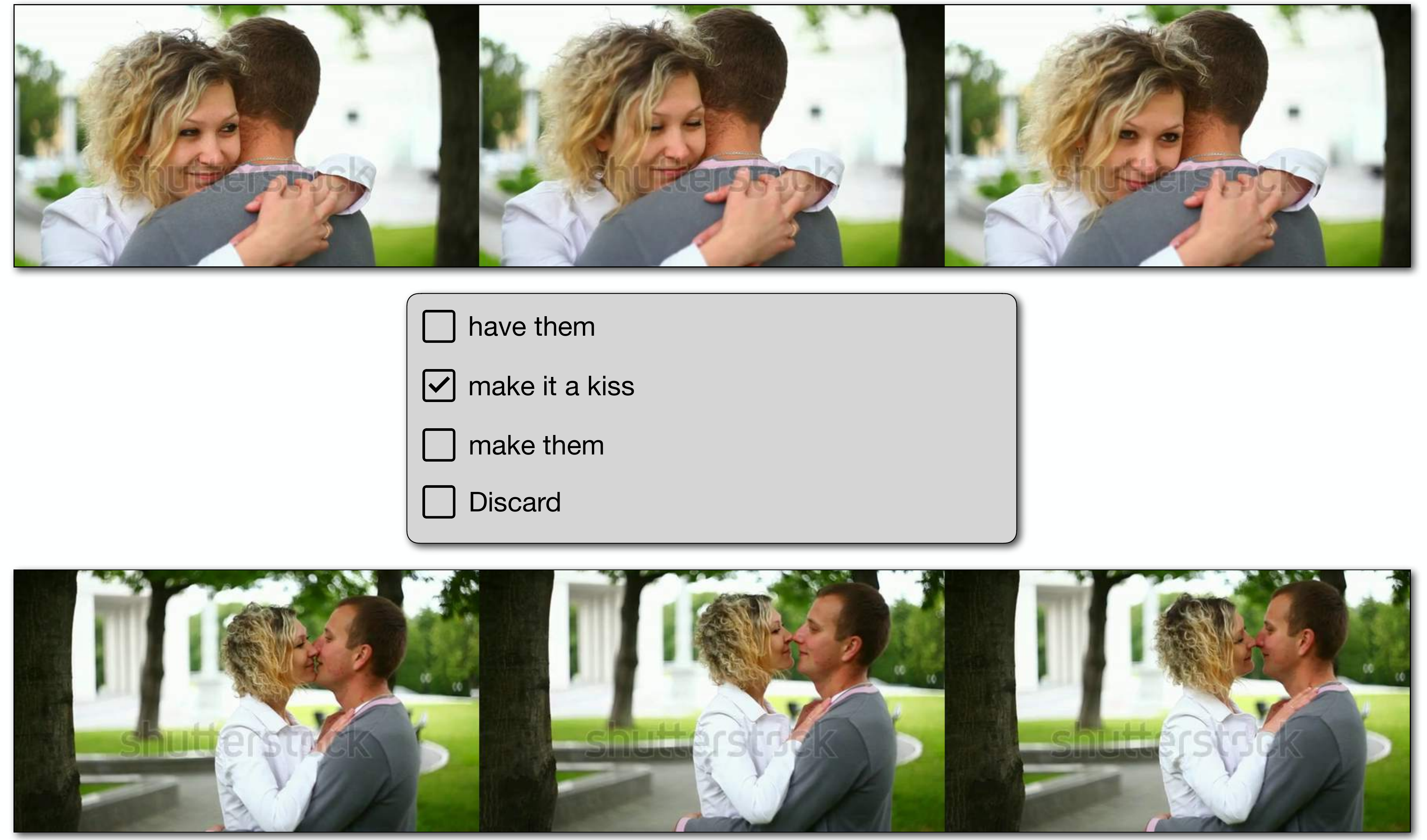}}\hfill
    {\includegraphics[width=0.38\textwidth]{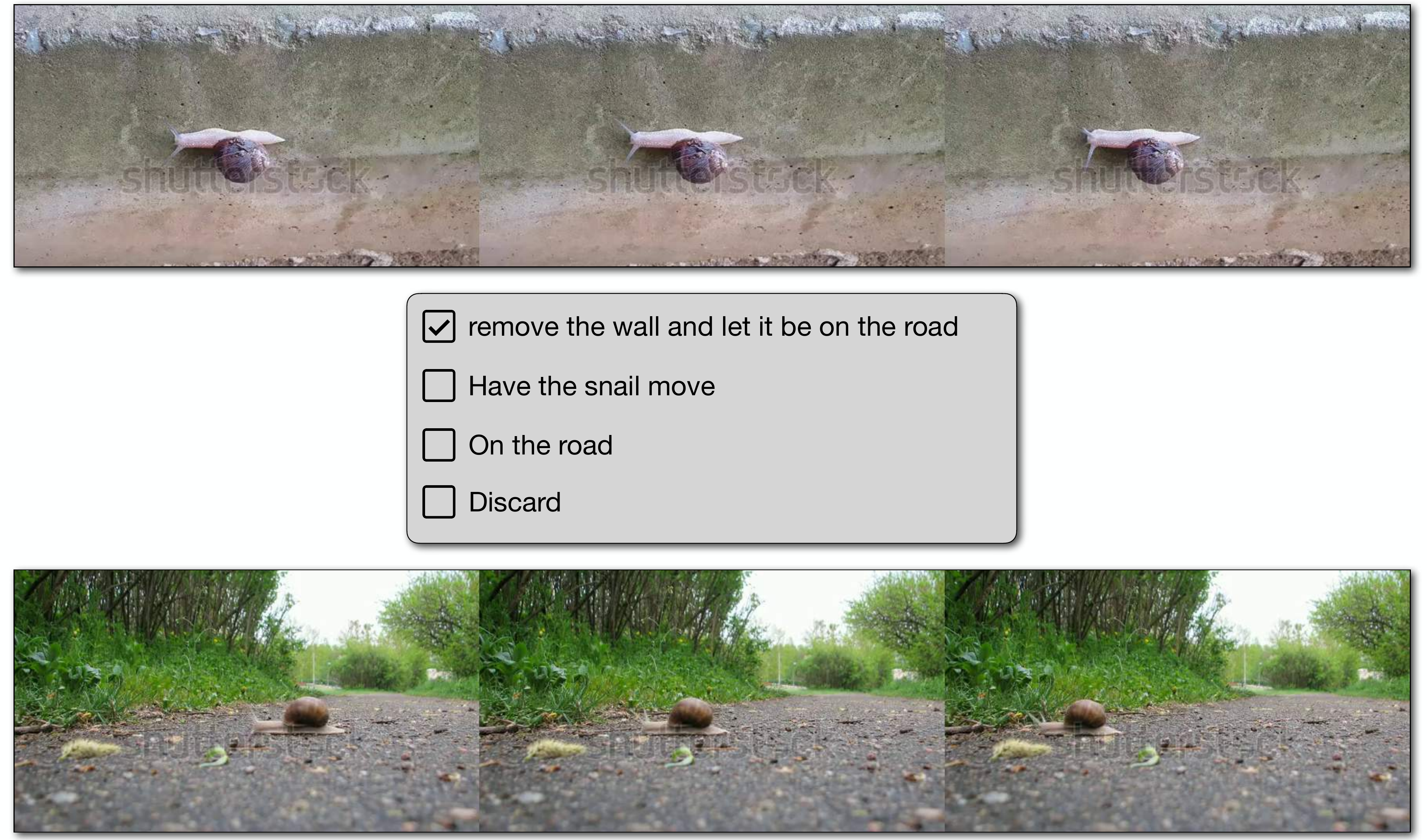}}\\\vspace{0.2cm}
    {\includegraphics[width=0.38\textwidth]{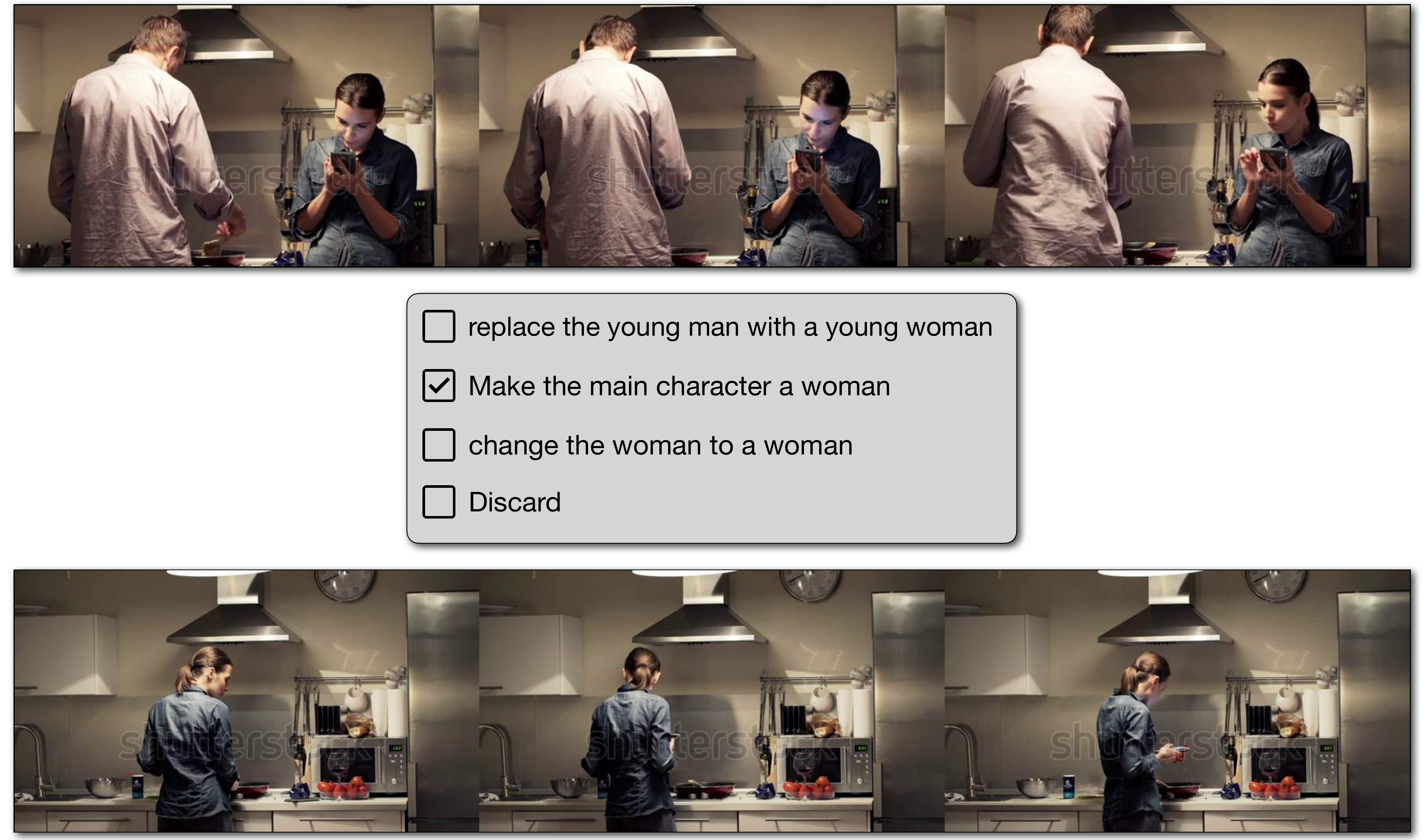}}\hfill
    {\includegraphics[width=0.38\textwidth]{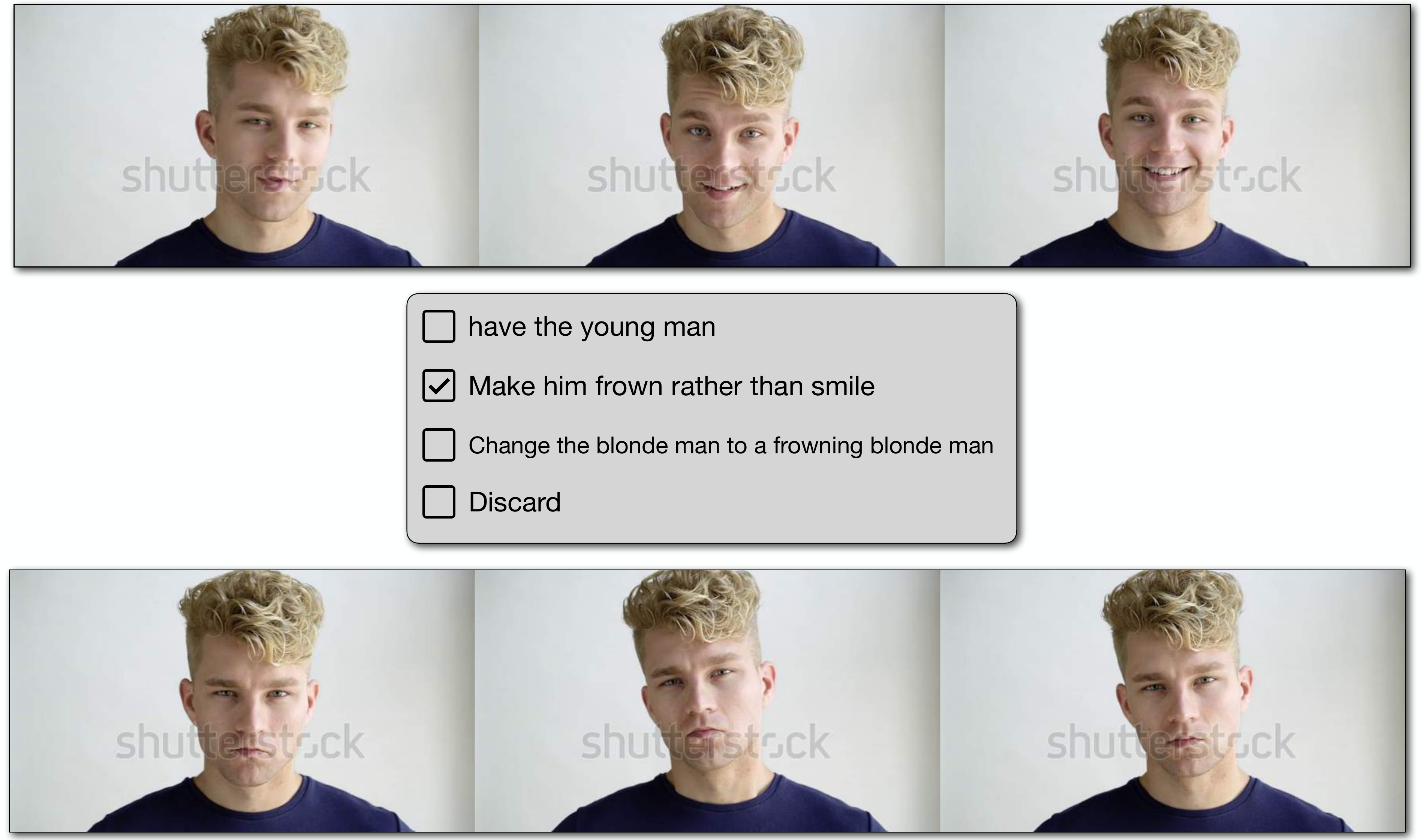}}\\\vspace{0.2cm}
    {\includegraphics[width=0.38\textwidth]{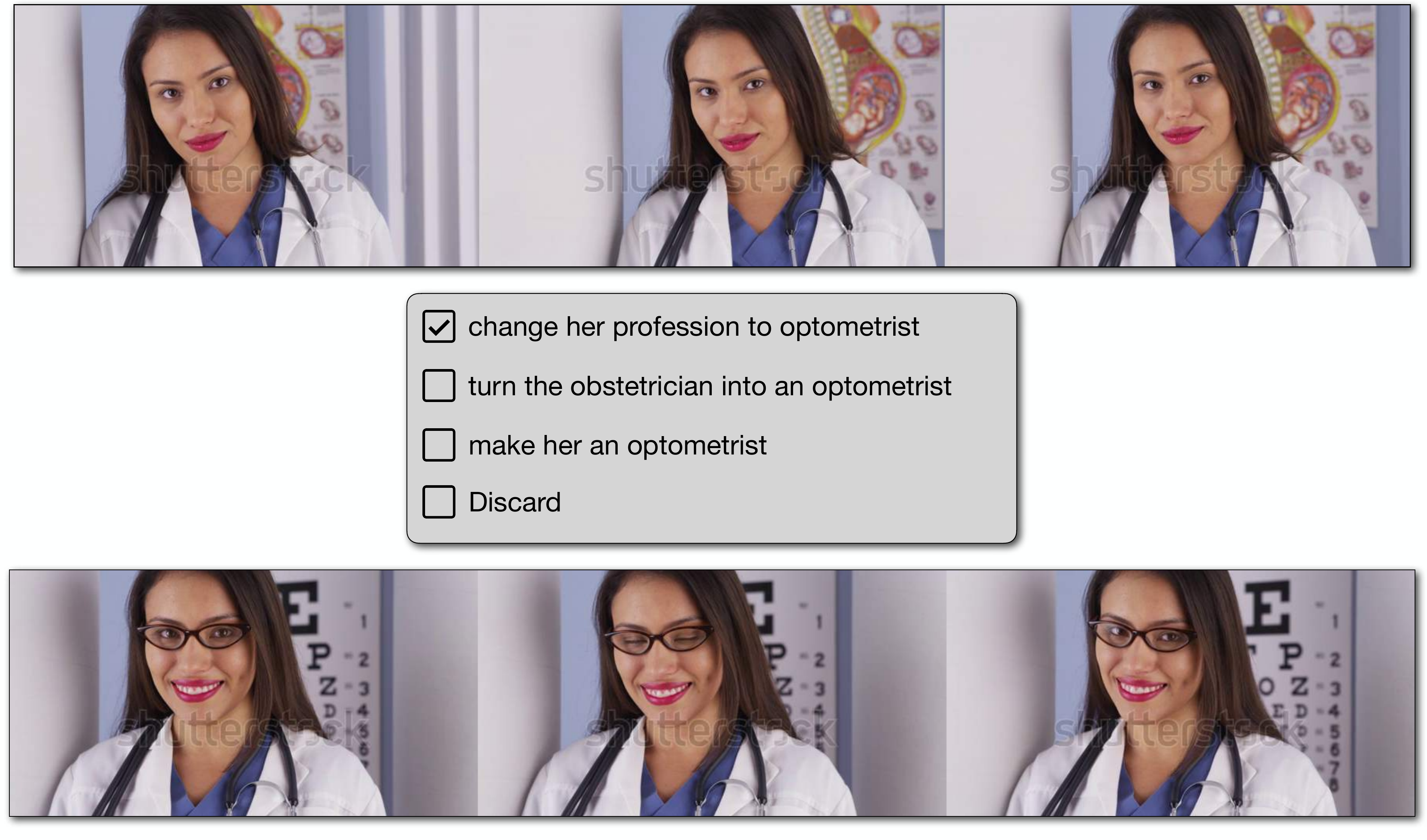}}\hfill
    {\includegraphics[width=0.38\textwidth]{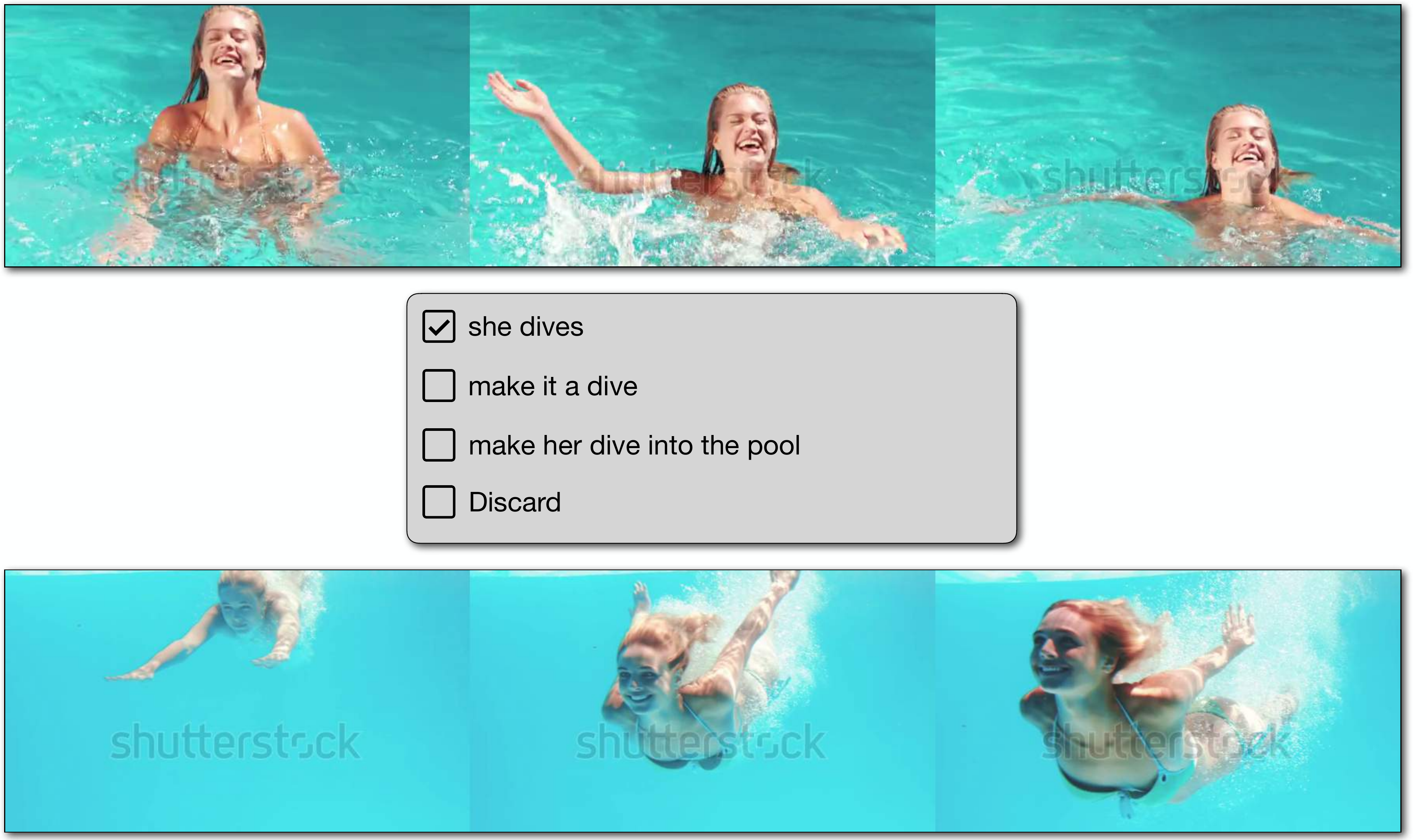}}
    \caption{\textbf{Manual annotation examples (kept):} We show samples from \ourWVt which are automatically mined triplets that are marked as correct during the annotation process. Each sample consists of two videos and a set of modification text options (in between each video pair). 
    The chosen modification text is indicated by a checkmark.}
    \label{app:fig:manual-test-correct}
\end{figure*}

\begin{figure*}
    \centering
    {\includegraphics[width=0.48\textwidth]{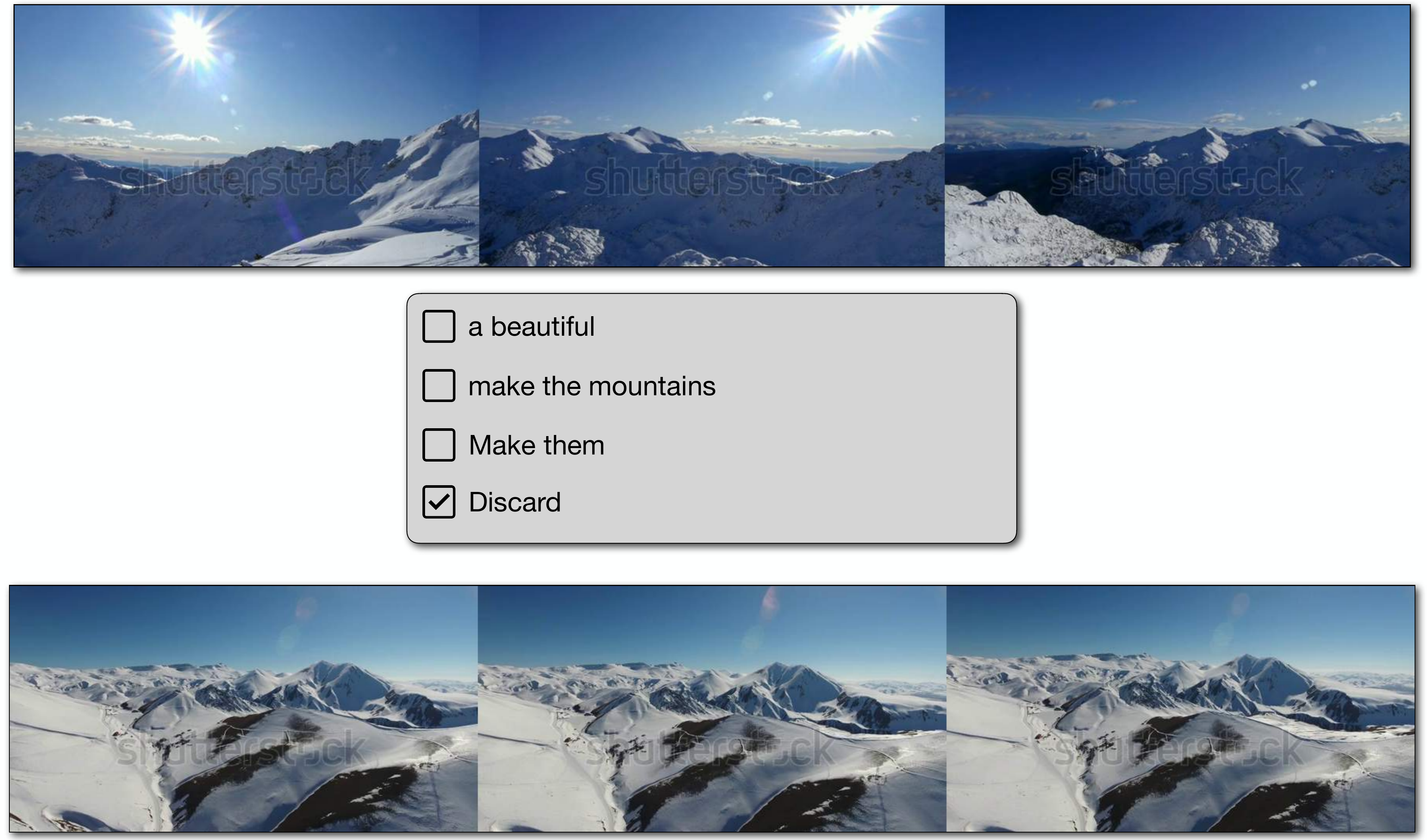}}\hfill
    {\includegraphics[width=0.48\textwidth]{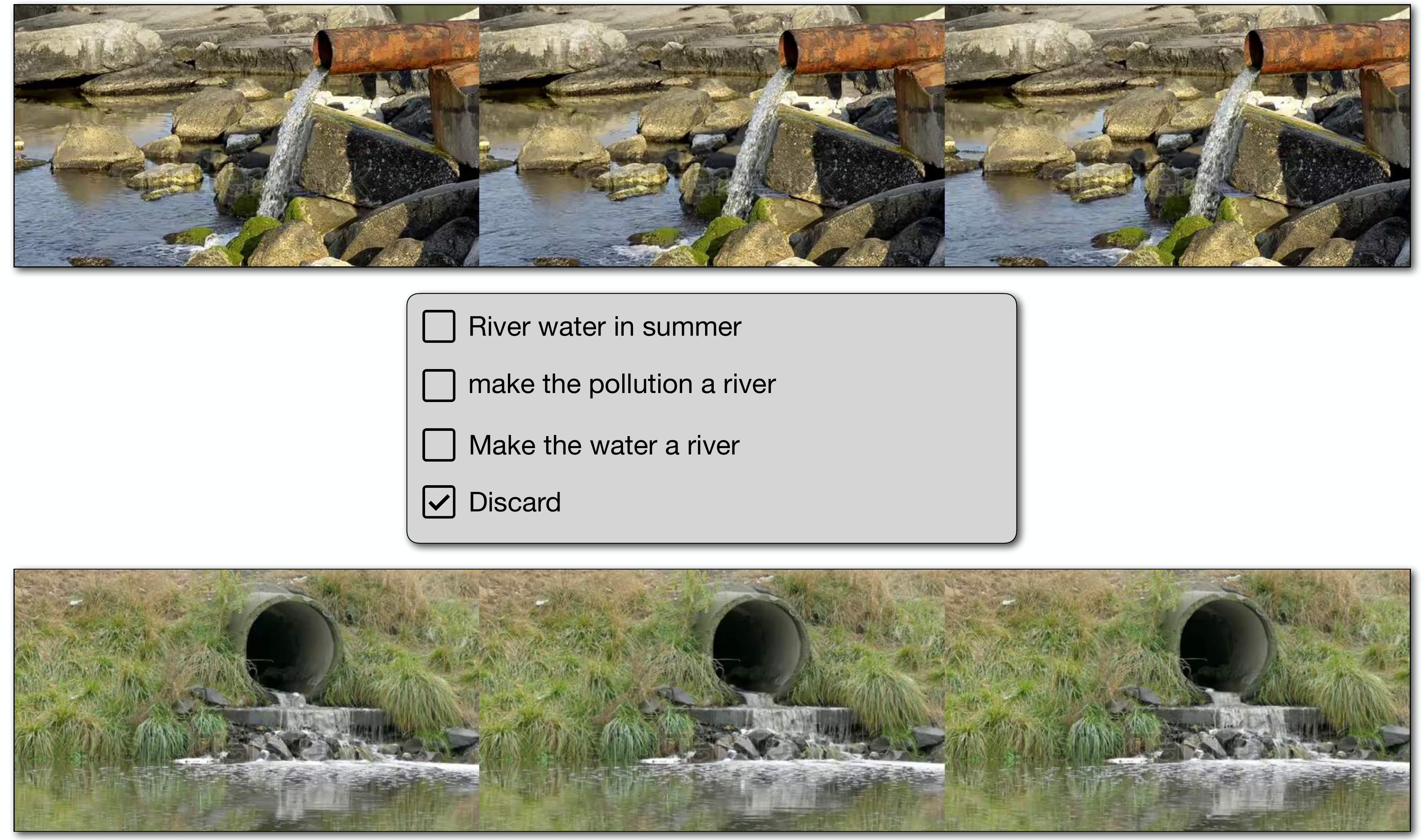}}\\\vspace{0.2cm}
    {\includegraphics[width=0.48\textwidth]{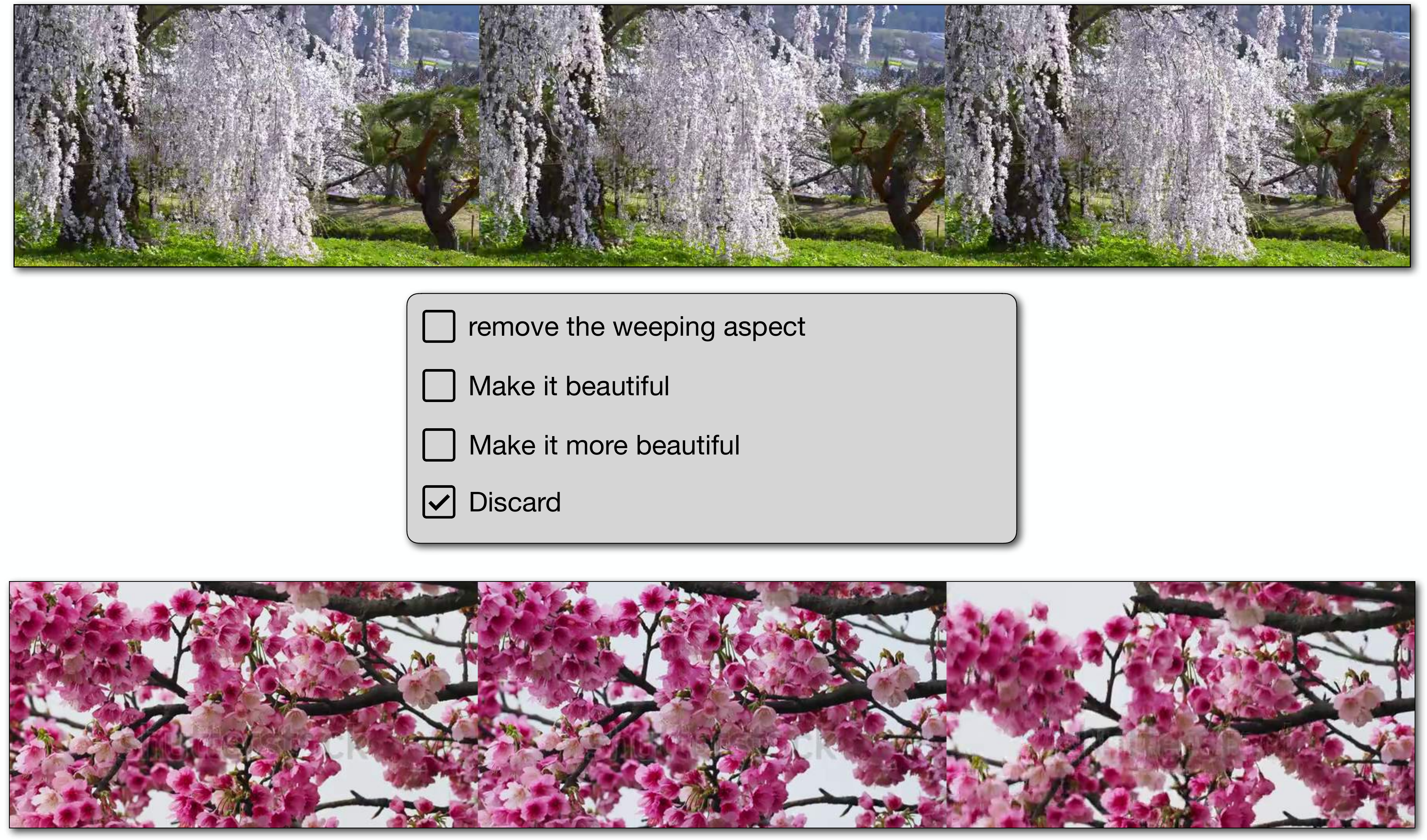}}\hfill
    {\includegraphics[width=0.48\textwidth]{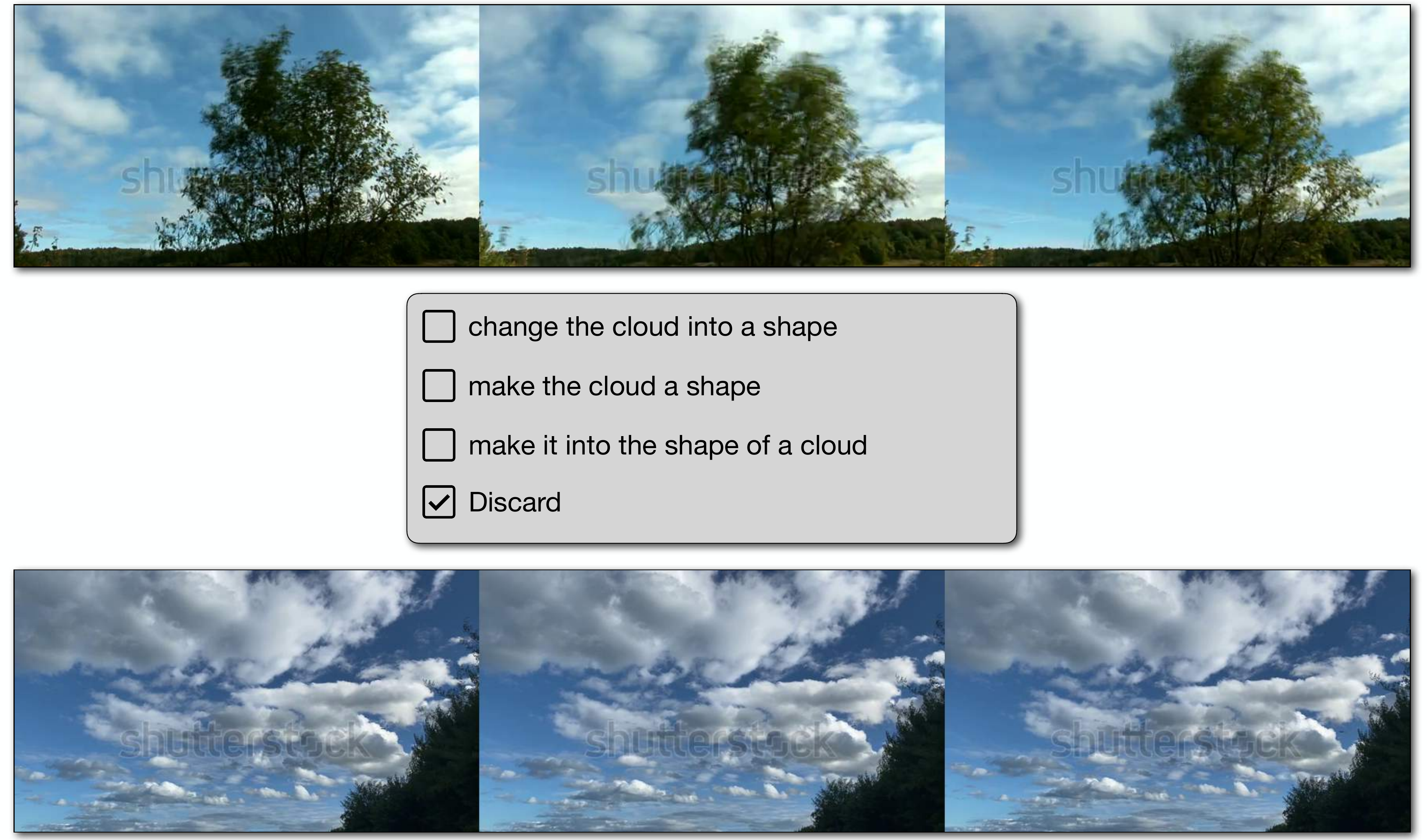}}\\\vspace{0.2cm}
    {\includegraphics[width=0.48\textwidth]{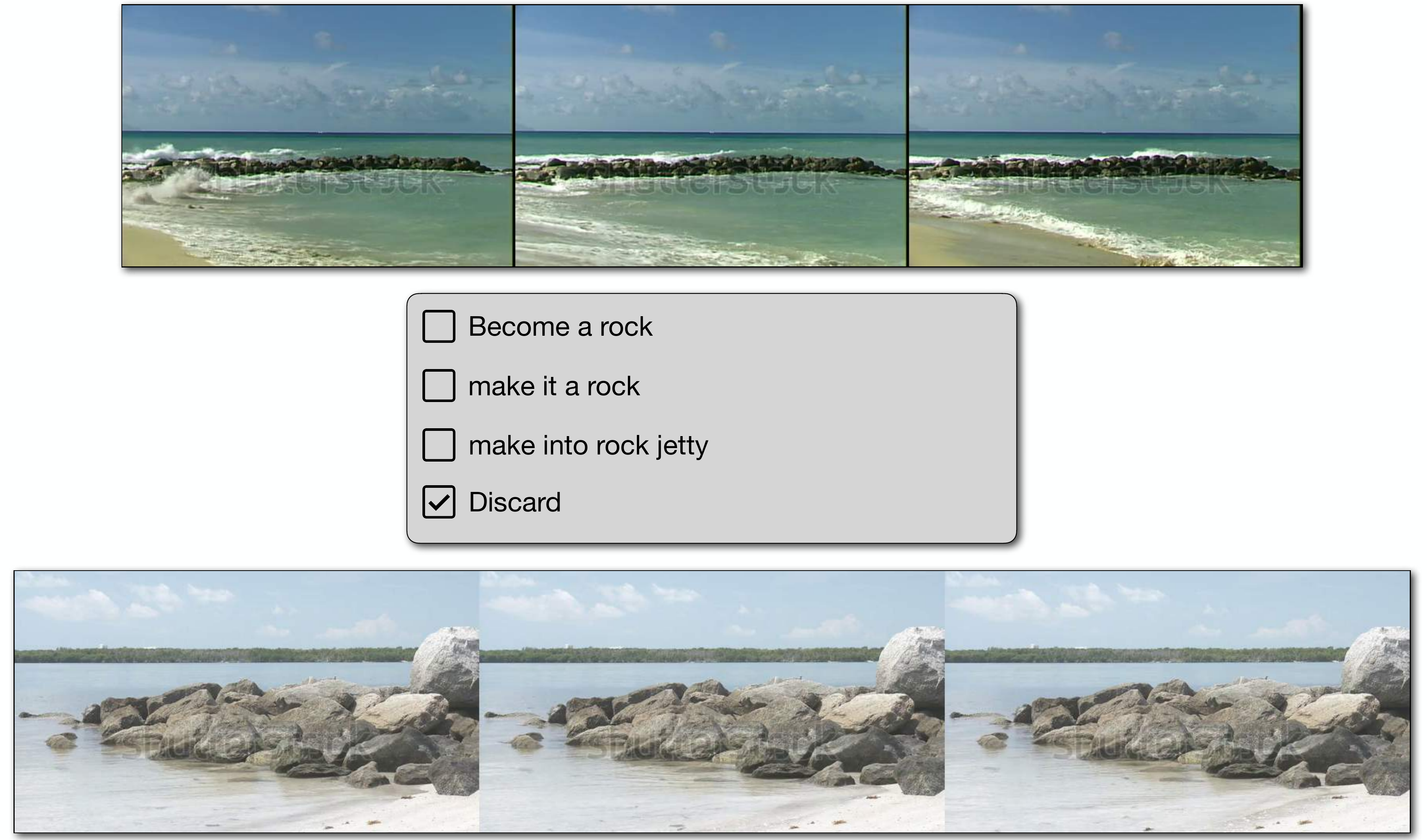}}\hfill
    {\includegraphics[width=0.48\textwidth]{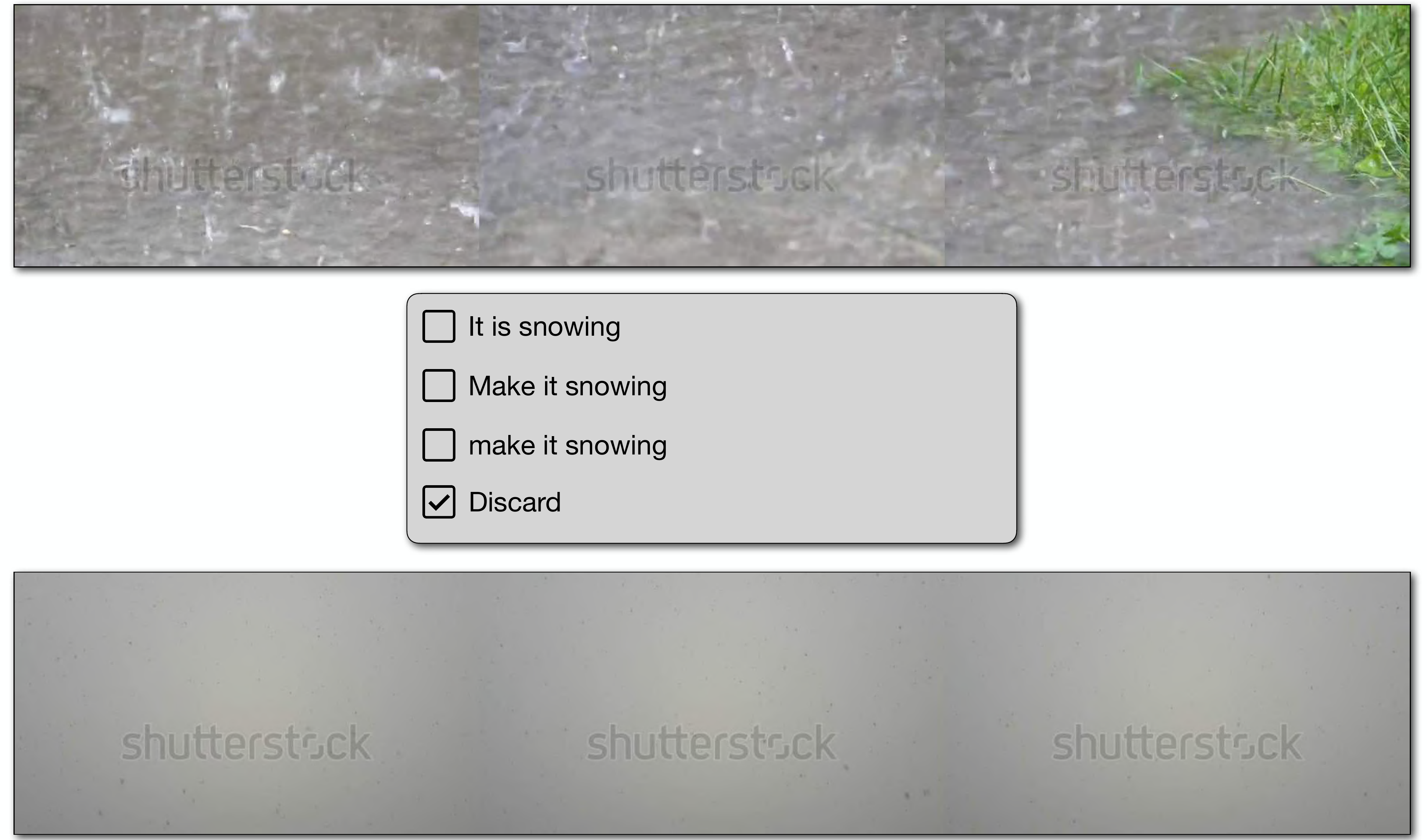}}\\\vspace{0.2cm}
    {\includegraphics[width=0.48\textwidth]{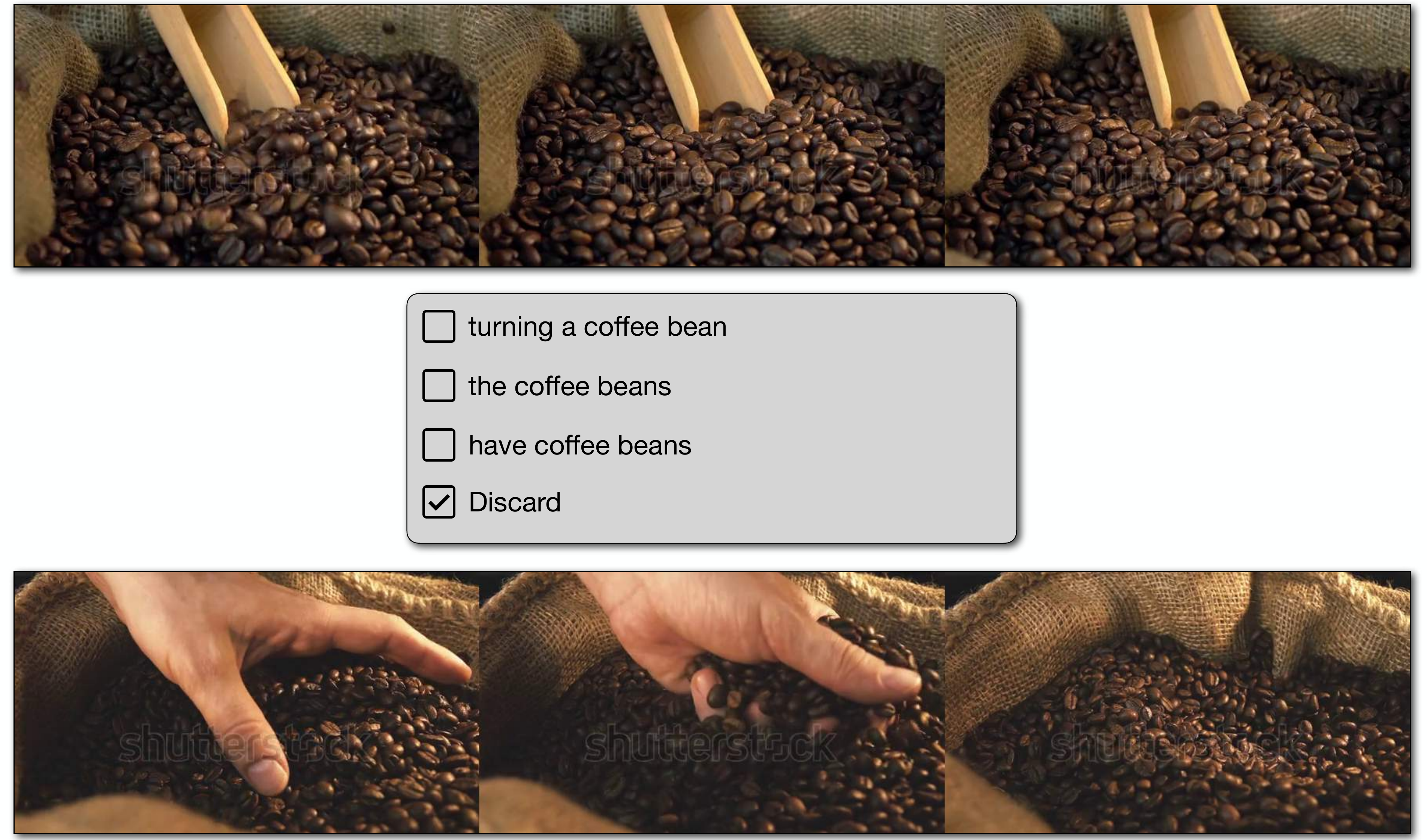}}\hfill
    {\includegraphics[width=0.48\textwidth]{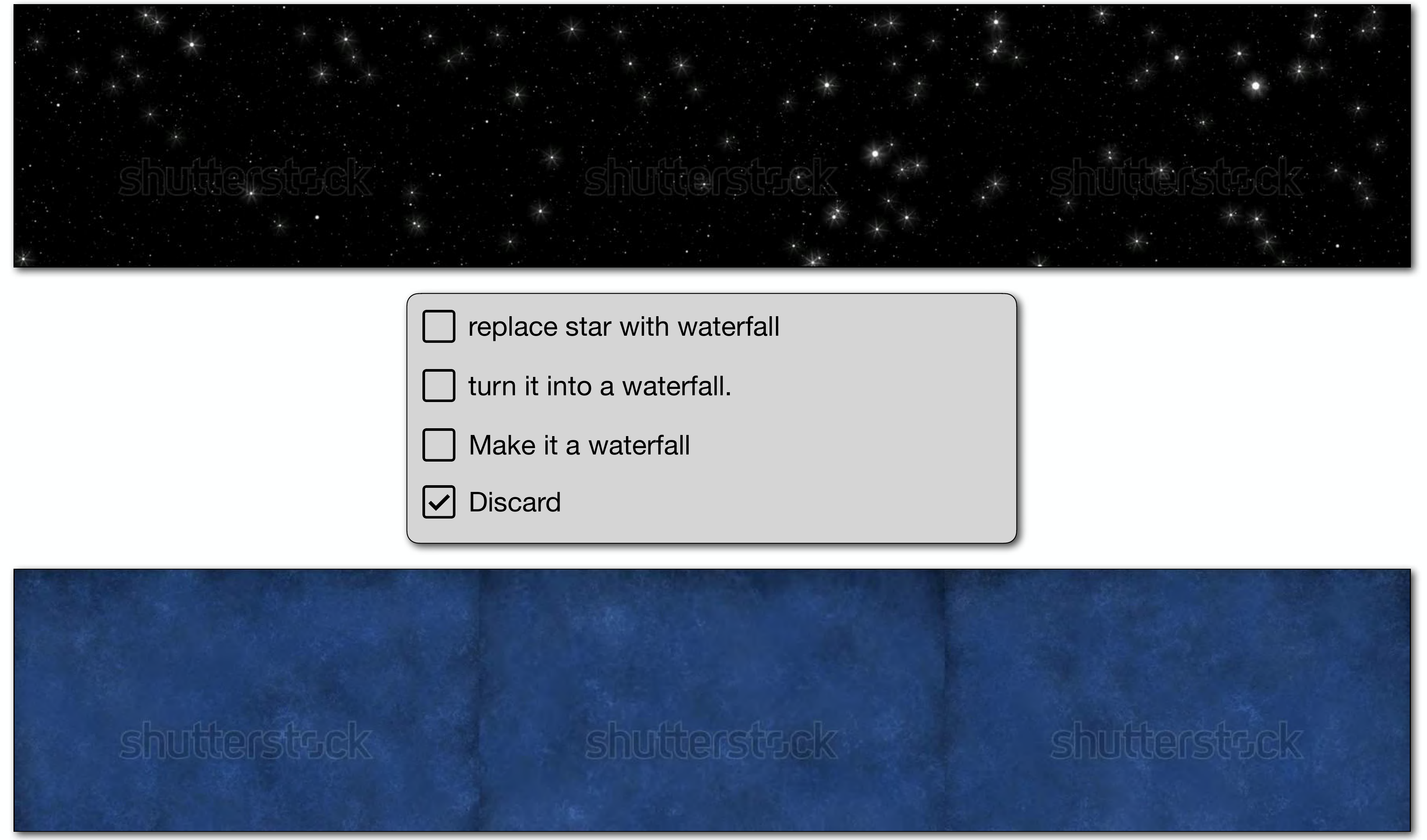}}
    \caption{\textbf{Manual annotation examples (discarded):} We show automatically mined triplets that are discarded during the annotation process. Discarded texts include 
    videos that are too similar (bottom left), too dissimilar (bottom right), or have bad modification texts (top left).
    }
    \label{app:fig:manual-test-incorrect}
\end{figure*}

A total of 3.1k triplets were shown for annotation.
In Figure~\ref{app:fig:manual-test-correct} and Figure~\ref{app:fig:manual-test-incorrect}, 
we present 10 examples that were considered correct during the annotation, along with the chosen modification texts (marked with a checkmark). 
These examples demonstrate successful modification texts and appropriate video content for the CoVR task.

On the other hand, in Figure~\ref{app:fig:manual-test-incorrect}, 
we show 8 examples that were discarded during the annotation. 
These examples were rejected either because the modification texts were incorrect 
or because the videos were deemed unsuitable for the CoVR task due to being either 
too similar ({e.g., bottom left, both videos are showing the same coffe with almost no modification}) or too incoherent ({e.g., top right example ``Make the water a river''}).

\subsection{Qualitative CoVR results on \ourWVt}
\label{app:subsec:recall-webvid}
\begin{figure*}%
  \centering
  \includegraphics[width=.8\linewidth]{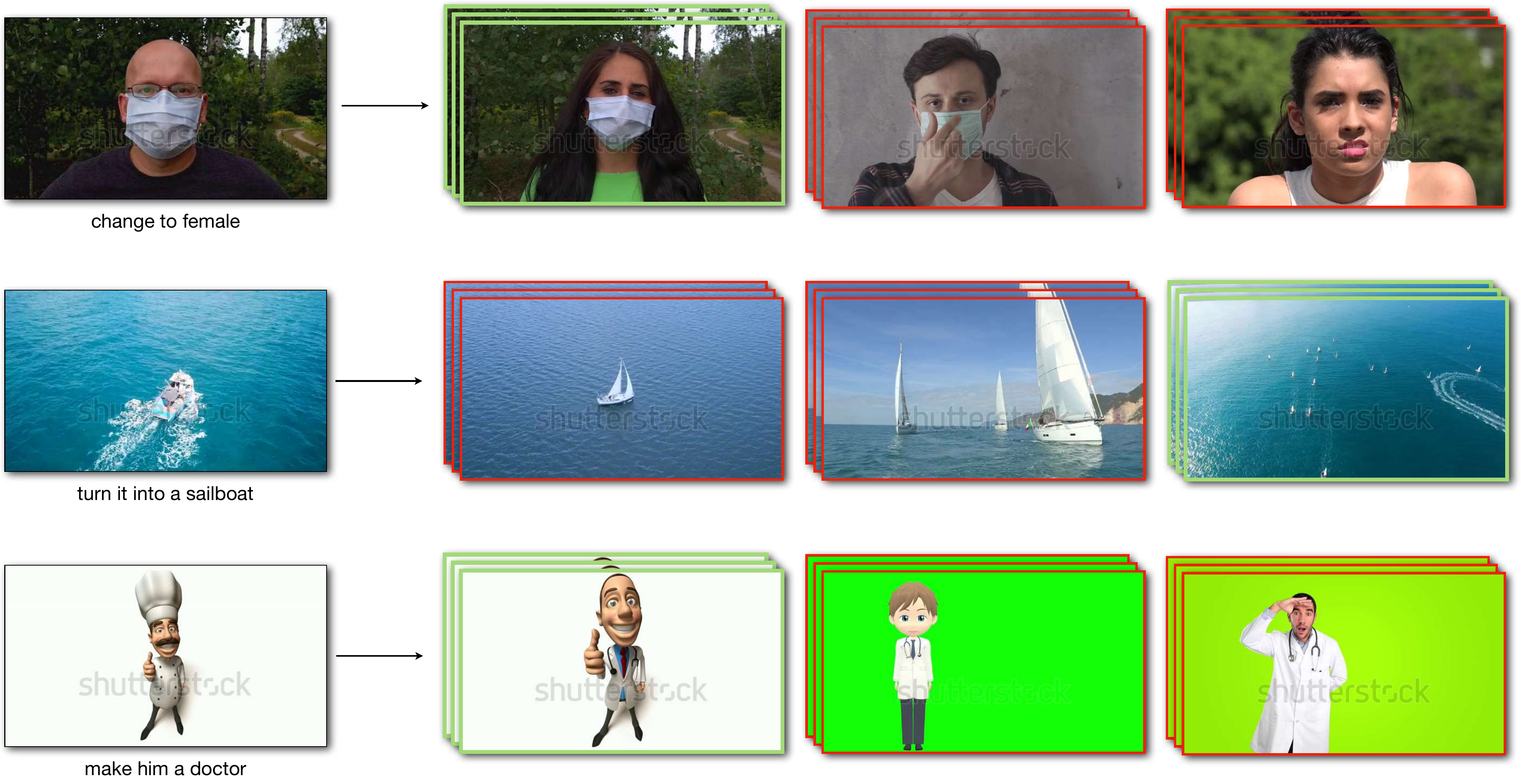}
  \includegraphics[width=.8\linewidth]{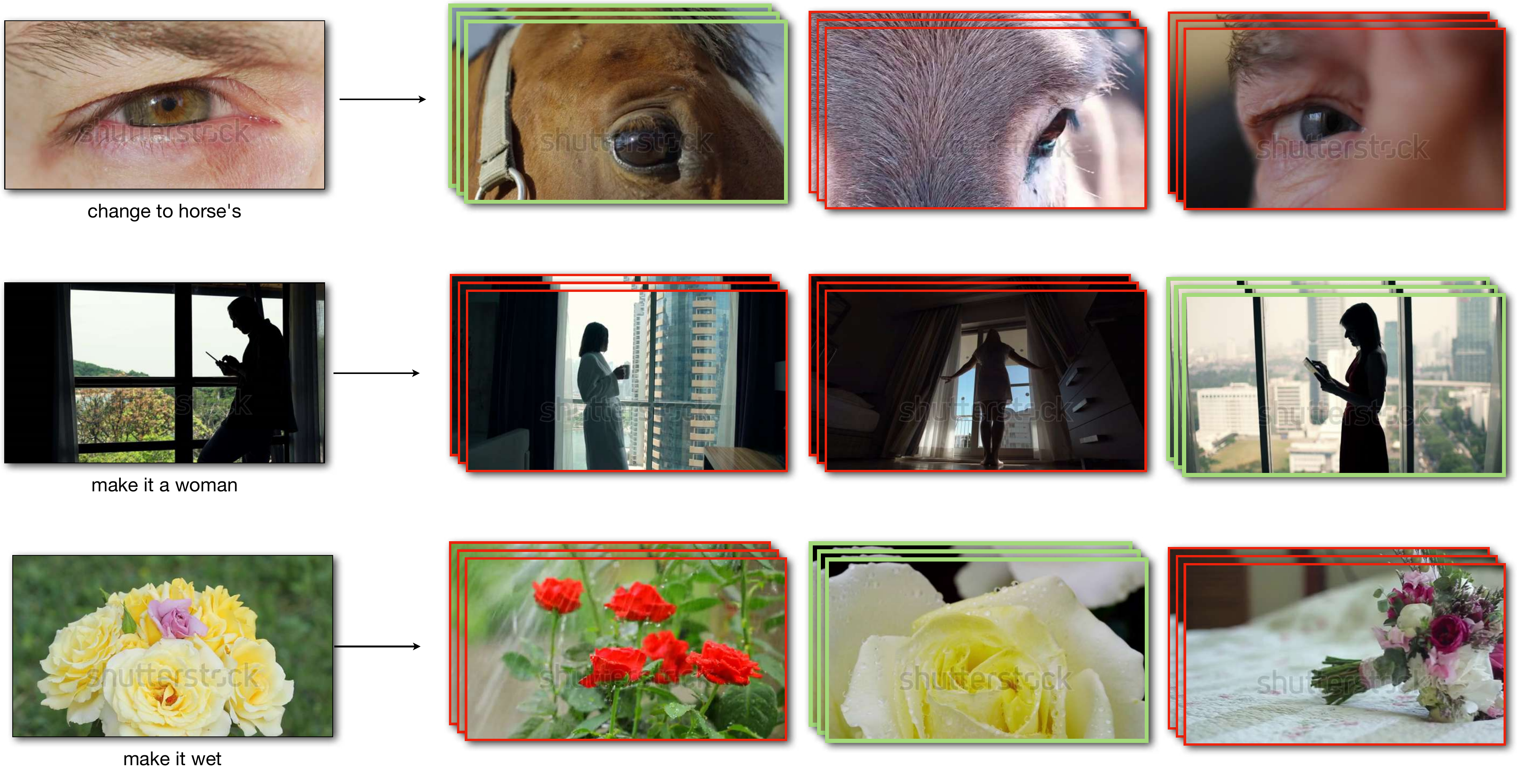}
  \includegraphics[width=.8\linewidth]{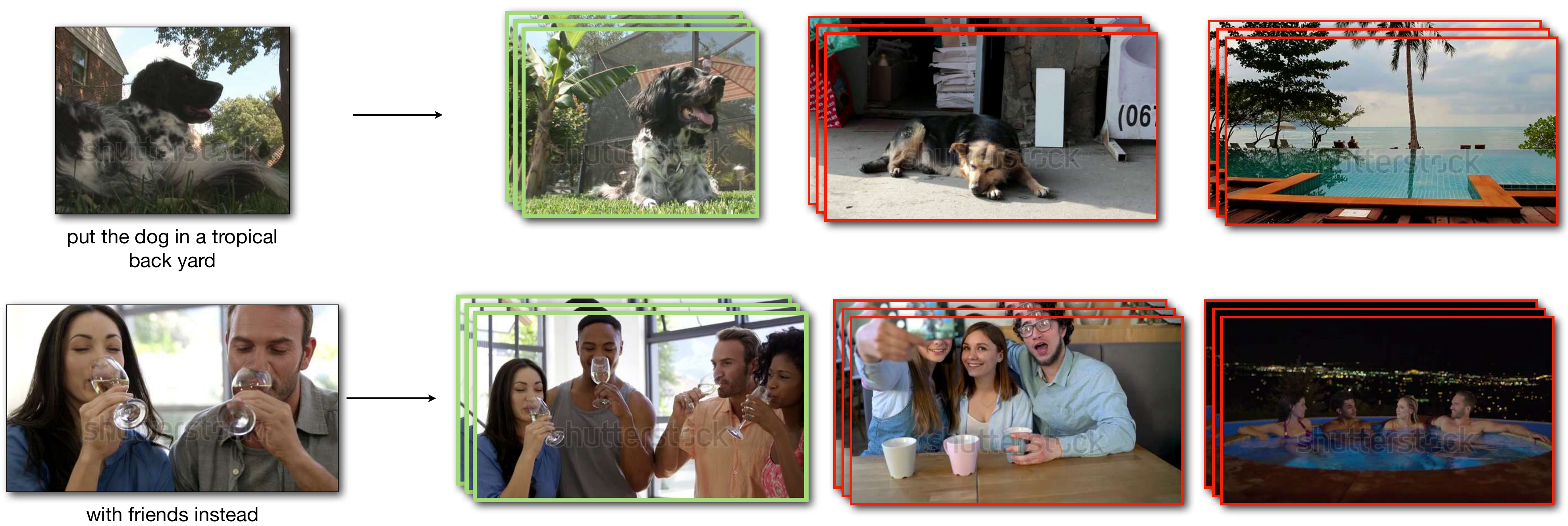}
  \caption{\textbf{Qualitative CoVR results on \ourWVt:} We display the input image and modification text queries on the left, along with the top 3 retrieved videos by our model on the right. 
  Ground-truth is denoted with a green border.  
  }
  \label{app:fig:recall-webvid-covr}
\end{figure*}

In Figure~\ref{app:fig:recall-webvid-covr}, we show qualitative CoVR results
on our manually verified \ourWVt set. We observe that top ranked
video frames have high visual and semantic similarity with the queries even
when not corresponding to the ground truth (marked with a green border).

\subsection{Qualitative CoIR results on the CIRR benchmark}
\label{app:subsec:recall-cirr}
\begin{figure*}%
  \centering
  \includegraphics[width=.8\linewidth, trim=0 18cm 0 0, clip]{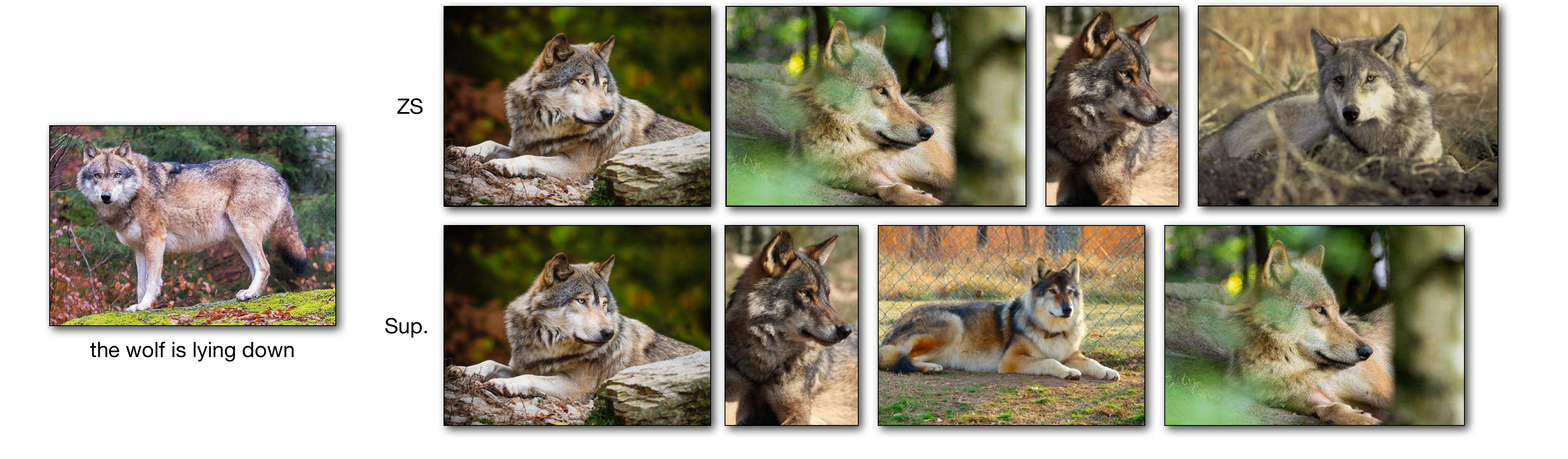}
  \includegraphics[width=.8\linewidth]{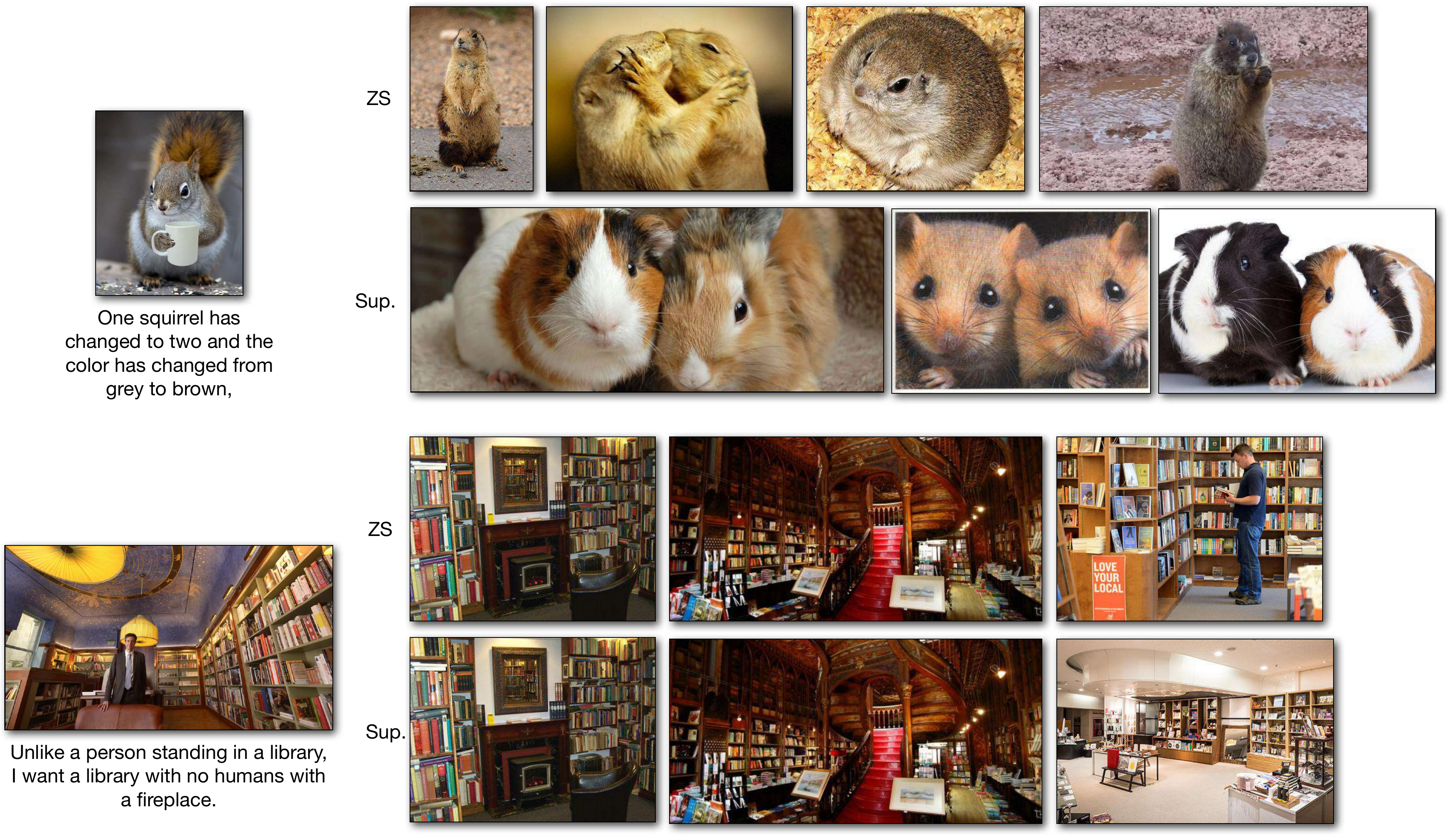}
  \includegraphics[width=.8\linewidth]{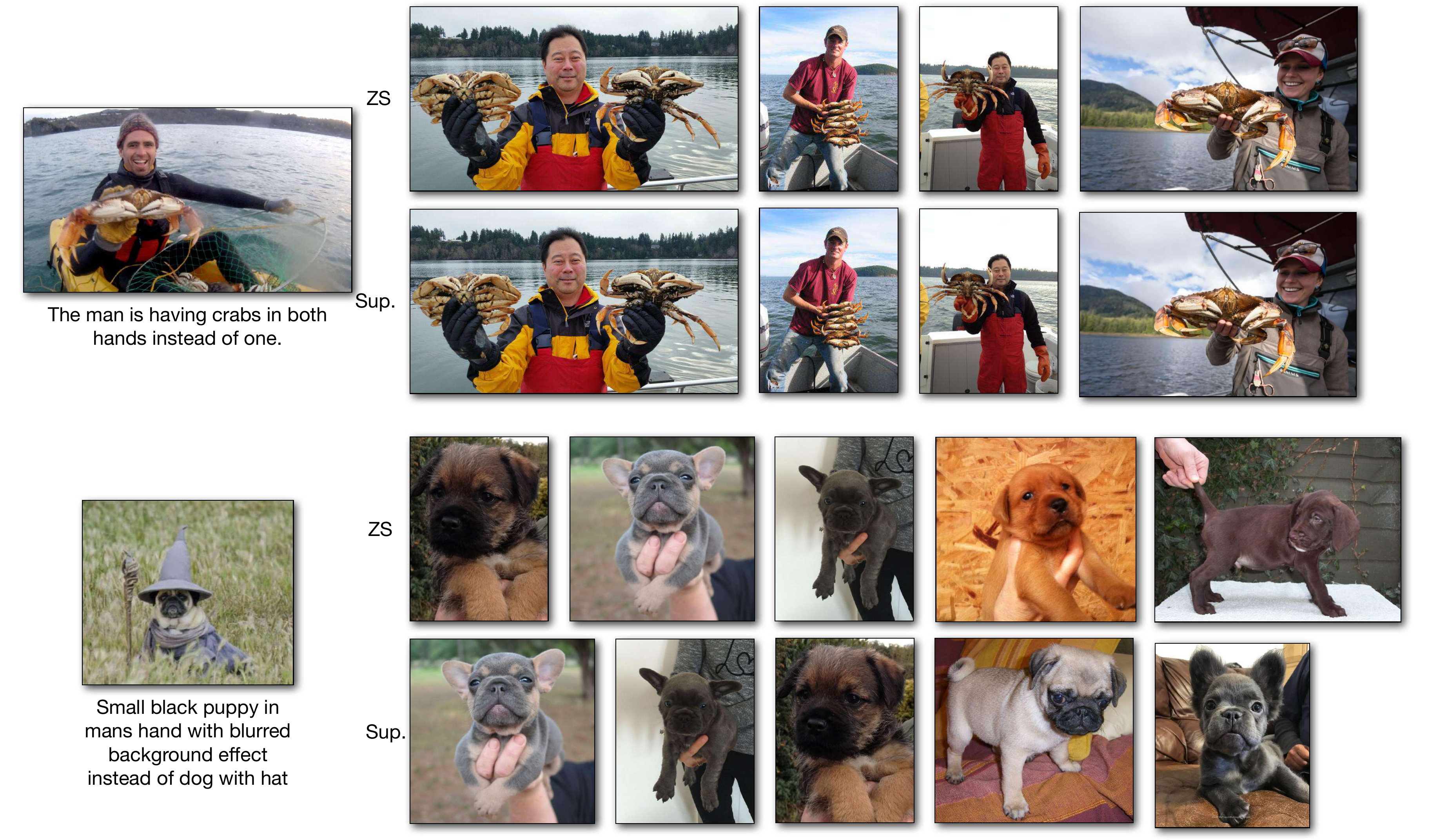}
  \caption{\textbf{Qualitative CoIR results on CIRR test set:} Given a query image and a modification text, we show our top retrieved videos of our zero-shot (ZS) model trained with \ourWV and the model finetuned on CIRR ground-truth supervision (Sup.).
  }
  \label{app:fig:recall-cirrr}
\end{figure*}

In Figure~\ref{app:fig:recall-cirrr}, we demonstrate qualitative CoIR results
of our models trained only on \ourWV (ZS) and the one further finetuned on CIRR training set (Sup.),
tested on the CIRR {test set. We observe promising retrieval quality for both models.}

%% file: tables/supmat/overlap_webvic-covr.tex
\begin{table}
    \caption{\textbf{\newt{Overlap between \ourWV training data and zero-shot CoIR benchmarks:}}
    \newt{
    We present the percentage overlap between target videos in our \ourWV training dataset and the target images in the test sets of three CoIR datasets at different CLIP cosine similarity thresholds (0.7, 0.8, and 0.9).
    Overlap is defined as the presence of at least one target image in the test set with a similarity score above the specified threshold. 
    The results indicate no overlap at the highest threshold (0.9).
    Note that this analysis focuses on the overlap of target images,
    not triplets. 
    }}
    \label{tab:overlap_cc-webvid} 
    \centering
    \resizebox{.99\linewidth}{!}{
    \begin{tabular}{l|c|ccc|c} 
    \toprule
         & &  & FashionIQ &  \\ 
        Threshold & CIRR & Dress & Shirt & Toptee & CIRCO \\ 
        \midrule
        0.7 & 45.4\% & 13.6\% & 13.8\% & 10.9\% & 79.0\%  \\
        0.8 & \textcolor{white}{0}1.6\% & \textcolor{white}{0}0.3\% & \textcolor{white}{0}0.0\% & \textcolor{white}{0}0.0\% & 12.6\%  \\
        0.9 & \textcolor{white}{0}0.0\% & \textcolor{white}{0}0.0\% & \textcolor{white}{0}0.0\% & \textcolor{white}{0}0.0\% & \textcolor{white}{0}0.0\%  \\
        \bottomrule
    \end{tabular}
    }
\end{table}

%% file: tables/supmat/overlap_cc-coir.tex
\begin{table}
    \caption{\textbf{\newt{Overlap between \ourCC training data and zero-shot CoIR benchmarks:}}
    \newt{
        We repeat the overlap analysis with \ourCC (as is similarly done for \ourWV in Table~\ref{tab:overlap_cc-webvid}). 
    The results indicate minimal overlap at the highest threshold (0.9), with fewer than 2\% similarity across all datasets, suggesting limited direct overlap.
    }}
    \label{tab:overlap_cc-coir} 
    \centering
    \resizebox{.99\linewidth}{!}{
    \begin{tabular}{l|c|ccc|c} 
    \toprule
         & &  & FashionIQ &  \\ 
        Threshold & CIRR & Dress & Shirt & Toptee & CIRCO \\ 
        \midrule
        0.7 & 34.0\% & \textcolor{white}{0}6.9\% & 10.8\% & \textcolor{white}{0}9.6\% & 59.1\%  \\
        0.8 & \textcolor{white}{0}5.2\% & \textcolor{white}{0}0.2\% & \textcolor{white}{0}0.3\% & \textcolor{white}{0}0.2\% & 20.7\%  \\
        0.9 & \textcolor{white}{0}0.2\% & \textcolor{white}{0}0.0\% & \textcolor{white}{0}0.0\% & \textcolor{white}{0}0.0\% & \textcolor{white}{0}1.6\%  \\
        \bottomrule
    \end{tabular}
    }
\end{table}

%% file: tables/supmat/rule-based-templates.tex
\begin{table}%
\centering
\caption{\textbf{Rule-based templates:}
For our rule-based MTG baseline, we randomly
choose one of the below templates during training.
}
    \begin{tabular}{l}
    \toprule
    \textit{Remove $\mathtt{txt\_diff_1}$ }\\
    \textit{Take out $\mathtt{txt\_diff_1}$ and add $\mathtt{txt\_diff_2}$ } \\
    \textit{Change $\mathtt{txt\_diff_1}$ for $\mathtt{txt\_diff_2}$ } \\
    \textit{Replace $\mathtt{txt\_diff_1}$ with $\mathtt{txt\_diff_2}$ } \\
    \textit{Replace $\mathtt{txt\_diff_1}$ by $\mathtt{txt\_diff_2}$ } \\
    \textit{Replace $\mathtt{txt\_diff_1}$ with $\mathtt{txt\_diff_2}$ } \\
    \textit{Make the $\mathtt{txt\_diff_1}$ into $\mathtt{txt\_diff_2}$ } \\
    \textit{Add $\mathtt{txt\_diff_2}$ } \\
    \textit{Change it to $\mathtt{txt\_diff_2}$ } \\
    \bottomrule
\end{tabular}
\label{tab:rule-based-templates}
\end{table}

%% file: tables/supmat/added-examles-mtg-llm.tex
\begin{table*}%
\centering
\caption{\textbf{Added examples to the MTG-LLM training:}
We add the below 15 examples to the set of 700 text triplets from \cite{brooks2022instructpix2pix}.
}
    \begin{tabular}{ll}
    \toprule
    Caption$_1$ & Clouds in the sky \\
Caption$_2$ & Airplane in the sky \\
Target output & Add an airplane \\
\midrule
Caption$_1$ & Woman with the tablet computer sitting in the city. \\
Caption$_2$ & Woman with tablet computer sitting in the park. \\
Target output & In the park \\
\midrule
Caption$_1$ & Walking swan \\
Caption$_2$ & White swan \\
Target output & Change color to white \\
\midrule
Caption$_1$ & Child playing on beach, sea waves view, girl spinning on coastline in summer 4k \\
Caption$_2$ & Child playing on beach, sea waves view, girl running on coastline in summer 4k \\
Target output & Make her spin \\
\midrule
Caption$_1$ & Aerial view of forest \\
Caption$_2$ & Aerial view autumn forest \\
Target output & Change season to autumn \\
\midrule
Caption$_1$ & Palm tree in the wind \\
Caption$_2$ & Palm trees in the wind \\
Target output & Add more palm trees \\
\midrule
Caption$_1$ & Schoolgirl talking on the phone \\
Caption$_2$ & Girl talking on the phone \\
Target output & Make her older \\
\midrule
Caption$_1$ & Clouds timelapse \\
Caption$_2$ & Sky timelapse \\
Target output & remove clouds and reveal only sky \\
\midrule
Caption$_1$ & Aerial view of a sailboat anchored in the mediterranean sea, vathi, greece. \\
Caption$_2$ & Aerial view of two sailboat anchored in the mediterranean sea, vathi, greece. \\
Target output & Add one sailboat \\
\midrule
Caption$_1$ & France flag waving in the wind. realistic flag background. looped animation background. \\
Caption$_2$ & Italian flag waving in the wind. realistic flag background. looped animation background. \\
Target output & Swap the flag for an italian one \\
\midrule
Caption$_1$ & Woman jogging with her dog in the park \\
Caption$_2$ & Woman playing with her dog in the park. \\
Target output & Stop jogging and make them play \\
\midrule
Caption$_1$ & Oil Painting Reproductions of by humans william-glackens \\
Caption$_2$ & Oil Painting Reproductions of zombies by william-glackens \\
Target output & Replace the humans with zombies \\
\midrule
Caption$_1$ & The girl who loved the sea by banafria \\
Caption$_2$ & The girl, wearing a hat, who loved the sea by banafria \\
Target output & Put a hat on her \\
\midrule
Caption$_1$ & famous painting Paris, a Rainy Day of Gustave Caillebotte \\
Caption$_2$ & famous painting Paris, a Sunny Day of Gustave Caillebotte \\
Target output & Change it to more pleasant weather \\
\midrule
Caption$_1$ & Bee on purple flower \\
Caption$_2$ & Bee on a flower \\
Target output & Change color of the flower \\
    \bottomrule
\end{tabular}
\label{tab:added-examplges-mtg-llm}
\end{table*}

%% file: tables/supmat/video-query.tex
\begin{table}
    \caption{\textbf{Querying with a video:}
        We report results on \ourWVt by using multiple frames from the query {\em video}.
        Recall that the rest of the paper investigates the setup where the middle video frame is used as an {\em image} query.
        We use 5 query video frames (uniformly sampled throughout the video). 
        The number of target video frames remains unchanged as 15.
        The performance is similar to the image query setup, with marginal increase.
    }
    \label{tab:video-query} %
    \centering
    \resizebox{.99\linewidth}{!}{
    \begin{tabular}{l|cccc}
        \toprule
        Visual query & R@1 & R@5 & R@10 & R@50 \\ 
        \midrule
        Image (middle frame)     & \textbf{59.82} & {83.84} & \textbf{91.28} & {98.24} \\
        Video  & 59.55 & \textbf{84.19} & 90.85 & \textbf{98.32} \\ %
    \bottomrule
    \end{tabular}
    }
\end{table}

%% file: tables/supmat/pretrained-blip-models.tex
\begin{table} %
\setlength\tabcolsep{5pt}
    \caption{\textbf{Variants of pretrained BLIP-2 backbones:} 
    We compare the BLIP-2 model without finetuning (base) and BLIP-2 finetuned on COCO (the one used in the rest of the paper)~\cite{BLIP}.
    For this experiment, we finetune the models on \ourWV using the cross-attention layers of
    BLIP-2 as the fusion method.
    }
    \centering
    \begin{tabular}{l|cccc}
        \toprule
        Backbone  & R@1 & R@5 & R@10 & R@50 \\ 
        \midrule
        BLIP-2 Base & 59.66 & \textbf{84.04} & 90.92 & \textbf{98.32} \\
        \rowcolor{ourcolor!64}
        BLIP-2 COCO  & \textbf{59.82} & {83.84} & \textbf{91.28} & {98.24} \\
    \bottomrule
    \end{tabular}
\label{tab:pretrained-blip-models}
\end{table}

%% file: tables/supmat/clip.tex
\begin{table} %
\centering
 \caption{\newt{\textbf{CLIP vs BLIP performance on CIR benchmarks:} 
    The table compares frozen CLIP with average fusion,
    CLIP with MLP fusion, 
    and BLIP and BLIP-2 with cross-attention layer finetuning.
    Results demonstrate that while MLP fusion struggles to generalize across datasets when trained with \ourWV, 
    the BLIP-2 backbone consistently outperforms CLIP across all benchmarks.
        }
	}
	\setlength\tabcolsep{5pt}
	\resizebox{0.99\linewidth}{!}{
		\begin{tabular}{ll|c|c|c}
        \toprule 
         Backbone & Pretraining & CIRR & FashionIQ & CIRCO \\
         & Data & R@1 & R@10 & mAP@5 \\
            \toprule 
            CLIP & - & 10.65 & 17.63 & 3.63 \\
            CLIP + MLP & \ourWVandCC & 11.39 & 19.10 & 5.67 \\
            \midrule
            BLIP & \ourWVandCC & 38.48 & 27.70 & 21.43 \\
            BLIP-2 & \ourWVandCC & \textbf{43.74} & \textbf{38.15} & \textbf{28.29} \\
            \bottomrule
		\end{tabular}
	}
	\label{tab:clip}
\end{table}

%% file: tables/supmat/visual-similarity.tex
\begin{table}
    \caption{\textbf{Effect of visual similarity:} We observe worse performance on CoIR zero-shot benchmarks as we increase the visual similarity threshold in our training data. We train each model for the same number of iterations.}
    \label{tab:visual-similarity} 
    \centering
    \resizebox{.99\linewidth}{!}{
    \begin{tabular}{lccc} 
    \toprule
        Threshold& Data (\%) & CIRR R@1 &FashionIQ R@10 mean\\ 
        \midrule
        0.00 (None)& 100\%& \textbf{41.30} & \textbf{37.01} \\ 
        0.55& \textcolor{white}{0}92\%& 40.72 & 36.13 \\ 
        0.65& \textcolor{white}{0}71\%& 38.07 & 35.71 \\ 
        0.70& \textcolor{white}{0}55\%& 35.78 & 35.03  \\ 
        \bottomrule
    \end{tabular}
    }
\end{table}

%% file: tables/supmat/static-dynamic.tex
\begin{table}
    \caption{\textbf{Performance on \ourWVt when training on dynamic vs static video triplets:} 
    \newt{We employ two methods to classify videos as static or dynamic: 
    one based on the optical flow of the target video, 
    and another based on the type of word (noun or verb) modified in the target caption. 
    The results indicate that training on the full dataset, 
    which includes both videos with temporal information and not, 
    yields the highest performance.}}

    \label{tab:static-dynamic} 
    \centering
    \resizebox{.99\linewidth}{!}{
    \begin{tabular}{lccccc} 
    \toprule
         & Data (\%) & R@1 & R@5	& R@10 & R@50 \\ 
        \midrule
        Static target videos & 25\%& 54.77 & 79.85 & 87.75 & 97.50 \\ 
        Dynamic target videos & 75\%&  58.80 & 83.10 & 90.61 & \textbf{98.24} \\ 
        \midrule
        \newt{Nouns change} & \newt{65\%} & \newt{58.06} & \newt{82.43} & \newt{89.98} & \newt{97.97} \\
        \newt{Verbs change} & \textcolor{white}{0}\newt{9\%} & \newt{53.44} & \newt{77.82} & \newt{85.64} & \newt{96.99} \\
        \midrule
        All & 100\%& \textbf{59.82} & \textbf{83.84} & \textbf{91.28} & \textbf{98.24} \\ 
        \bottomrule
    \end{tabular}
    }
\end{table}

%% file: tables/supmat/text-length.tex
\begin{table} %
	\caption{\textbf{Increasing the average modification text length} 
		in \ourWV 
		by selecting the longest of multiple generated candidates per caption pair 
		degrades downstream performances
		on \ourWVt, CIRR, and FashionIQ.
	}
    \centering
    \resizebox{0.99\linewidth}{!}{
    \begin{tabular}{r|cc|cc|cc}
        \toprule
         Modification text & \multicolumn{2}{c|}{\ourWVt}  & 
         \multicolumn{2}{c|}{CIRR} & 
         \multicolumn{2}{c}{FashionIQ} 
          \\
        avg \#chars & R@1 & MeanR & R@1 & MeanR  &  R@10 & R@50  \\
        \midrule
        \rowcolor{ourcolor!64}
        (ours) 23.36 &   \textbf{59.82} & \textbf{83.30} & \textbf{41.42} & \textbf{73.22} & \textbf{36.81} & \textbf{56.70} \\
        (longest) 33.35 & 58.84 & 82.72 & 35.64 & 67.60 & 33.20 & 52.13 \\
        \bottomrule
    \end{tabular}
    }
    \label{tab:text-length}
\end{table}

%% file: tables/supmat/rule-based-comparison.tex
\begin{table*} %
\centering
\caption{\textbf{Comparison between modification text generation approaches:}
    We provide qualitative examples for a pair of captions, and three methods to generate
    modification text: (i) rule-based, (ii) prompting-based, (iii) our MTG-LLM finetuning.
    Rule-based method is limited, for example in the case where the difference text is a preposition (last row),
    whereas the prompting-based method is prone to hallucinating (e.g., `remove iceberg', `change the pose of the runner').
    Our approach tends to be the most robust across cases.}
    \label{tab:rule-based-comparision}
    \begin{tabular}{ll}
    \toprule
    Caption$_1$ & \textit{Happy} girl dancing \\
    Caption$_2$ & \textit{Beautiful} girl dancing \\
    Rule-based & Make the \textit{Happy} into \textit{Beautiful} \\
    Prompting LLM & Change girl \\
    MTG-LLM & Make her more \textit{beautiful} \\
    \midrule
    Caption$_1$ & Black \textit{bird} \\
    Caption$_2$ &  black \textit{bear} \\
    Rule-based & Add \textit{bear} \\
    Prompting LLM & Remove animal \\
    MTG-LLM & Turn it into a \textit{bear} \\
    \midrule
    Caption$_1$ & \textit{Autumn} landscape in the mountains. \\
    Caption$_2$ & \textit{Winter} landscape in the mountains \\
    Rule-based & Remove \textit{Autumn} \\
    Prompting LLM & Change scene to winter. \\
    MTG-LLM & Change to \textit{winter} \\
    \midrule
    Caption$_1$ & Flying over the \textit{ice} \\
    Caption$_2$ & Flying over the \textit{mountains} \\
    Rule-based & Take out \textit{ice} and add \textit{mountains} \\
    Prompting LLM & Remove iceberg \\
    MTG-LLM & Move the \textit{ice} to the \textit{mountains} \\
    \midrule
    Caption$_1$ & \textit{Dandelion} field \\
    Caption$_2$ & \textit{Rice} field  \\
    Rule-based & Change \textit{dandelion} for \textit{rice} \\
    Prompting LLM & Change the season \\
    MTG-LLM & Turn it into a \textit{rice} field. \\
    \midrule
    Caption$_1$ & \textit{Happy} woman \\
    Caption$_2$ & \textit{Running} woman \\
    Rule-based & Change it to \textit{running} \\
    Prompting LLM & Remove the girl and change the pose of the runner \\
    MTG-LLM & Make her a runner \\
    \midrule
    Caption$_1$ & Woman working on cellphone at \textit{night}  \\
    Caption$_2$ & Woman working on cellphone at \textit{outdoor} \\
    Rule-based & Replace \textit{night} with \textit{outdoor} \\
    Prompting LLM & Change the woman and add one cellphone \\
    MTG-LLM & Make it an \textit{outdoor} \\
    \midrule
    Caption$_1$ & Barber cuts the hair of the client with \textit{clipper}. \\
    Caption$_2$ & Barber cuts the hair of the client with \textit{trimmer} \\
    Rule-based & Take out \textit{clipper} and add \textit{trimmer} \\
    Prompting LLM & Add or change the hairdresser's tool \\
    MTG-LLM & Change to a \textit{trimmer} \\
    \midrule
    Caption$_1$ & Aerial shot \textit{of} a lake. \\
    Caption$_2$ & Aerial shot \textit{above} a lake  \\
    Rule-based & Remove \textit{of} \\
    Prompting LLM & Add a lake to the sequence \\
    MTG-LLM & make it an aerial shot \textit{of} a lake \\
    \bottomrule
\end{tabular}
\end{table*}